\pdfoutput=1
\documentclass[11pt]{article}
\usepackage[printwatermark]{xwatermark}
\usepackage[]{EMNLP2022}
\usepackage{times}
\usepackage{latexsym}
\usepackage[T1]{fontenc}
\usepackage[utf8]{inputenc}
\usepackage{microtype}
\usepackage{inconsolata}
\usepackage{graphicx}
\usepackage{xcolor}
\usepackage{xspace}
\usepackage{amsmath}
\usepackage{caption}
\usepackage{subcaption}
\usepackage{booktabs}
\usepackage{multirow}
\usepackage[normalem]{ulem}
\usepackage{stfloats}

\DeclareMathOperator*{\argmax}{arg\,max}

\newcommand{\newText}[1]{#1}
\newcommand{\oldText}[1]{}

\usepackage[printwatermark]{xwatermark}
\newwatermark*[pages=21-23,color=green!40!black!30!white,angle=0,scale=.5,xpos=0,ypos=13.5cm]{\textit{New Content Added 2023-03-30}}

\title{Neural Theory-of-Mind?\\On the Limits of Social Intelligence in Large LMs}

\newcommand\uwcse{$^\heartsuit$}
\newcommand\cmu{$^\diamondsuit$}

\newcommand\aitwo{$^\spadesuit$}

\newcommand{\aspace}{\hspace{.8em}}

\author{Maarten Sap\aitwo\cmu \aspace
Ronan Le Bras\aitwo \aspace
Daniel Fried\cmu \aspace
Yejin Choi\aitwo\uwcse \\
\normalsize{\aitwo Allen Institute for AI, Seattle, WA, USA}\\
\normalsize{\cmu Language Technologies Institute, Carnegie Mellon University, Pittsburgh, USA}\\
\normalsize{\uwcse Paul G. Allen School of Computer Science, University of Washington, Seattle, WA, USA}\\
\texttt{\href{mailto:maartensap@cmu.edu}{maartensap@cmu.edu}}
}

\newcommand{\socialIQa}{\textsc{SocialIQa}\xspace}
\newcommand{\tomi}{\textsc{ToMi}\xspace}
\newcommand{\TheoryOfMind}{Theory of Mind\xspace}

\newcommand{\ToM}{\textsc{ToM}\xspace}
\newcommand{\gptT}{\textsc{GPT-3}\xspace}
\newcommand{\gptDavinci}{\textsc{GPT-3-DaVinci}\xspace}
\newcommand{\gptCurie}{\textsc{GPT-3-Curie}\xspace}
\newcommand{\gptAda}{\textsc{GPT-3-Ada}\xspace}

\newcommand{\ptlm}{LLM\xspace}
\newcommand{\ptlms}{LLMs\xspace}

\newcommand{\personEmoji}{\includegraphics[height=.8em,trim=0em .5em 0em .5em]{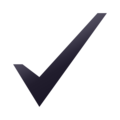}}
\newcommand{\robotEmoji}{\includegraphics[height=.8em,trim=0em .5em 0em .5em]{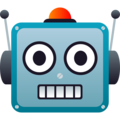}}

\begin{document}

\maketitle

\begin{abstract}

Social intelligence and \TheoryOfMind (\ToM), i.e., the ability to reason about the different mental states, intents, and reactions of all people involved, allow humans to effectively navigate and understand everyday social interactions.
As NLP systems are used in increasingly complex social situations, their ability to grasp social dynamics becomes crucial.
% AI and NLP systems are increasingly deployed in settings that require complex social reasoning abilities.
% AI and NLP systems require increasingly complex social reasoning abilities, as they are deployed in increasingly multi-user and user-adaptive settings.
% AI and NLP systems are increasingly used in complex social settings, such as to enhance communication between multiple people or to interact and adapt to users.
% However, doing so effectively requires social intelligence and \TheoryOfMind (\ToM), i.e., the ability to reason about the different mental states, realities, intents, and reactions of all people involved.

In this work, we examine the open question of social intelligence and \TheoryOfMind in modern NLP systems from an empirical and theory-based perspective. % through empirical exploration and critical discussion.
We show that one of today's largest language models \cite[GPT-3;][]{brown2020language} lacks this kind of social intelligence out-of-the box, using two tasks:
% \maarten{Mention something about how people seemingly expect AGI to emerge from reading text? Or something about learning from predicting words?}
% In this work, we demonstrate that even today's largest language model \cite[GPT-3;][]{brown2020language} lacks this kind of social intelligence out-of-the box.
% To shed light onto this limitation, we analyze model performance on two tasks: %\ronan{Wondering if we should pitch it slightly differently, as this may sound as if we simply report the performance on two well-known benchmarks. Maybe more along the lines of: we analyze the performance and shed light on the limitations of current LMs?}\maarten{love it, added!}
\socialIQa \cite{sap2019socialIQa}, which measures models' ability to understand intents and reactions of participants of social interactions, and \tomi \cite{le-etal-2019-revisiting}, which measures whether models can infer mental states and realities of participants of situations.
% We show that models struggle on these social intelligence tasks, through comprehensive probing and question-answering experiments.

Our results show that models struggle substantially at these \TheoryOfMind tasks, with well-below-human accuracies of 55\% and 60\% on \socialIQa and \tomi, respectively. %, that lie well below human performance.
% Indeed, the best performance on \socialIQa reaches 55\% (>30\% below human performance), and accuracy on \tomi questions about mental states peaks at 60\% (30--40\% below accuracy on factual questions).
To conclude, we draw on theories from pragmatics to contextualize this shortcoming of large language models, by examining the limitations stemming from their data, neural architecture, and training paradigms.
Challenging the prevalent narrative that only scale is needed, we posit that person-centric NLP approaches might be more effective towards neural \TheoryOfMind.
% We conclude that increasing the scale of language models is likely not the most effective approach ,  that

% We discuss the implications and limitations of the current language modelling paradigm, \maarten{to update:} and posit that more interaction-oriented or participant-centric approaches to learning language are perhaps necessary.
% Our results suggest that learning language from static text alone will not endow models with social intelligence, and that perhaps more interactive or entity-aware approaches are necessary.    
\end{abstract}

% The argument I want to make: LM objective is not enough, we need to incorporate explicit social modeling if we want to get models to do social intelligence. Because LM-ing does not typically require reasoning about many people's mental states

% Something about how Theory of Mind is the ultimate type of reasoning, because it requires mental simulations from all kinds of perspectives.

\begin{figure}[t!]
    \centering
    \includegraphics[width=.95\columnwidth]{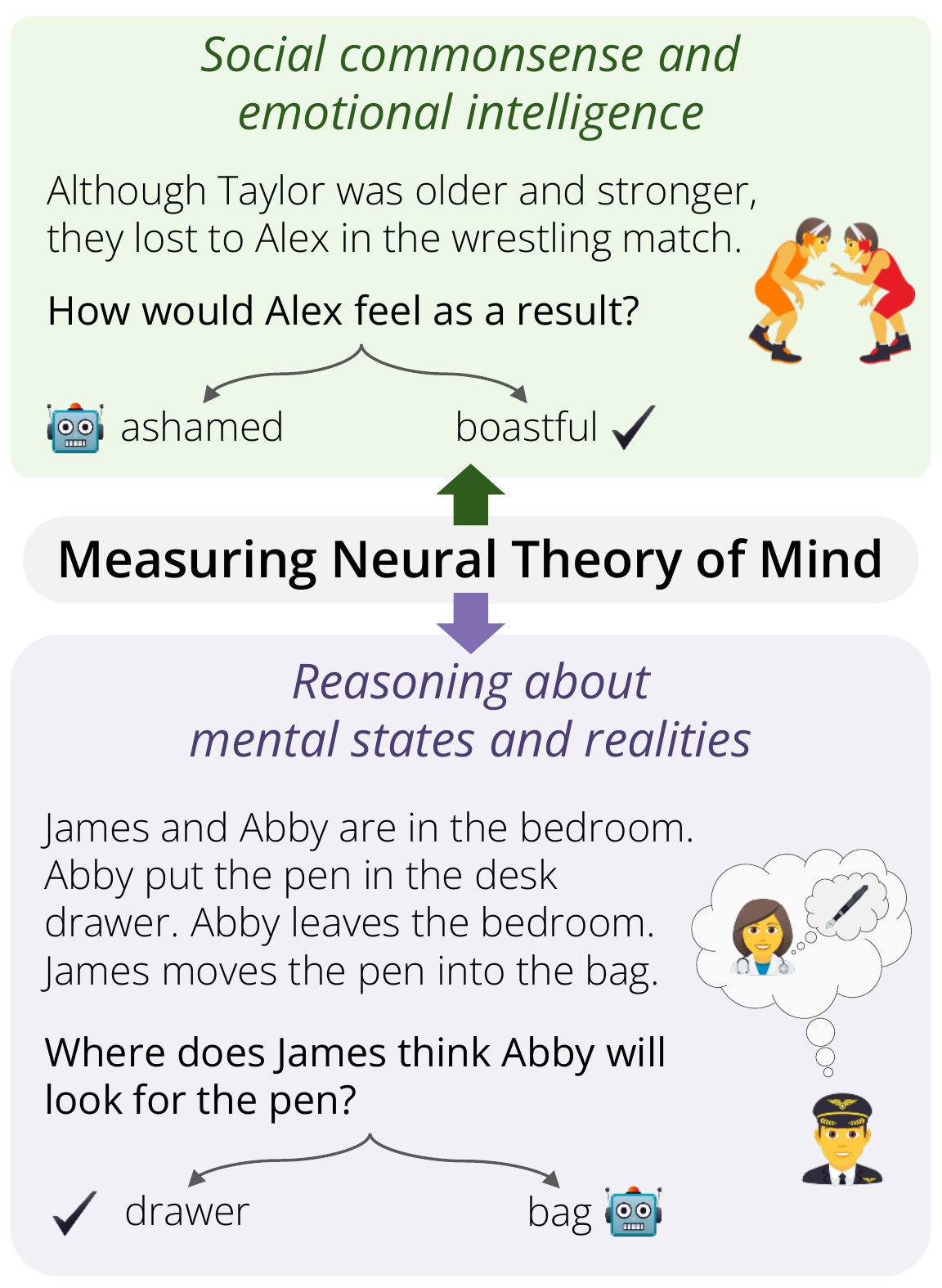}
    \caption{\TheoryOfMind is the ability for humans to reason about the intents, reactions, and mental states of others. 
    We asses these abilities in \ptlms
    % We show that \ptlms struggle with \ToM 
    through two question-answering tasks that measure social commonsense and emotional intelligence (\socialIQa; top) and reasoning about people's mental states and realities (\tomi; bottom); finding that \gptT (\robotEmoji) struggles on both tasks.
    We discuss why that may be, drawing from \oldText{pragmatics theories}\newText{theories of the pragmatics} of language.
    % \oldText{Added the sentence ``James and Abby are in the bedroom'' to make it clearer.}
    }
    \label{fig:intro-fig}
    %\vspace{-1.6em}
\end{figure}

\section{Introduction}

% \begin{figure*}[t]
%     \centering
%     \includegraphics[width=.95\textwidth]{graphics/IntroFig_v3.pdf}
%     \caption{\TheoryOfMind is the ability for humans to reason about the intents, reactions, and mental states of others. 
%     We assess the \ToM abilities of \ptlms through two question answering tasks that measure social commonsense and emotional intelligence (\socialIQa) and reasoning about people's mental states and realities (\tomi).
%     }
%     \label{fig:intro-fig}
% \end{figure*}

% As the prevalence of AI and NLP systems in everyday social interactions grows, so does their need for \textit{social intelligence} and 
% \textit{\TheoryOfMind} (\ToM), i.e., the ability to infer and reason about the intents, feelings, and mental states of others.
With the growing prevalence of AI and NLP systems in everyday social interactions, the need for AI systems with \textit{social intelligence} and 
\textit{\TheoryOfMind} (\ToM), i.e., the ability to infer and reason about the intents, feelings, and mental states of others, becomes increasingly evident \cite{Pereira2016socialpower,Langley2022-fl}.
For humans, \TheoryOfMind 
is a crucial component that enables us to interact and communicate effectively with each other \cite{Premack1978-sh,apperly2010mindreaders}.
% This type of social intelligence
It allows us, for example, to infer that someone likely feels boastful instead of ashamed after winning a wrestling match (Fig.~\ref{fig:intro-fig}; top).
In addition, \ToM also 
% Additionally, this ability 
enables us to reason about people's mental realities, e.g., if someone was out of the room while a pen was moved, she will likely search for the pen where she last saw it instead of where it was moved to (Fig.~\ref{fig:intro-fig}; bottom).
% \TheoryOfMind (\ToM) is crucial for humans to navigate everyday encounters; 

% With the growing presence of technology in everyday social interactions, 
% the need for such \TheoryOfMind and social intelligence in AI systems becomes increasingly necessary \cite{Wang2007-xa,schuller2018age,Dhelim2021-sd}.
% Such \theoryOfMind (\ToM) abilities enable us to 
% seamlessly navigate social situations ranging from friendly interactions to complex negotiations \cite{apperly2010mindreaders}.
% seamlessly navigate everyday encounters \cite{apperly2010mindreaders}.
% and is increasingly necessary for AI systems to successfully interface or interact with us in our everyday lives \cite{Wang2007-xa,Pereira2016socialpower,schuller2018age}.
% As AI systems are increasingly prevalent in everyday lives, such \ToM abilities in these systems are going to become increasingly necessary. 
%\maarten{Short paragraph describing the two types of social intelligence we refer to (e.g., commonsense reasoning about social interaction, false belief tests}

While humans develop it naturally, \ToM and social intelligence remain elusive goals for modern AI systems \cite{Choi2022-sv}, including large neural language models (\ptlms). % despite suggestions that these models are capable of commonsense reasoning \cite{Narang2022palmsBlog}.
% \maarten{Maybe more position-y: In this work, we discuss why \ToM and social commonsense reasoning remain an elusive goal, }
% In this work, we establish that \ToM and social commonsense reasoning remains an elusive goal for modern AI systems such as large neural language models (\ptlms)\maarten{, despite suggestions that these models are capable of commonsense reasoning \cite{Narang2022palmsBlog}?}.
With advances in scaling the sizes of models and datasets, these \ptlms have proven very impressive at generating human-like language for conversational, summarization, or sentence continuation settings, often with zero to few examples to learn from \cite{brown2020language,clark2021all,Chowdhery2022palm}.
% We investigate to what extent such \ToM abilities remain a challenge for modern AI systems such as large neural language models (\ptlms).
% Recent advances \ptlms due to scaling the training data and model size have made these models quite impressive at things like conversations, summarization, etc. often with no or few training examples \cite{brown2020language,clark2021all}.
However, increasing scrutiny has shed light on the shortcomings of these \ptlms, showing that they often fall prey to spurious correlational patterns instead of displaying higher-order reasoning \cite{elkins2020can,dale2021gpt,Marcus2022-fg}.
% coherence, commonsense, factuality, or consistency
%However, there is increasing discourse/evidence that these models lack fundamental social reasoning abilities \cite{Bender2020-kt,Bisk2020-xb,elkins2020can,dale2021gpt,somethingEmpirical}.
%\maarten{Paragraph/sentence about why AI systems need this kind of intelligence? they are increasingly prevalent in society, increasingly engaged in social tasks like communication\cite{Chen2019smartCompose}, counseling \cite{Kearns2020-ty,Allen2020-cd}, educational settings \cite{breazeal2002designing,williams2019popbots}}

% \maarten{Argument: PTLMs have gotten super super good at so many tasks, like conversational }
% Despite impressive advances in neural modelling, this type of complex social reasoning remains a challenge for modern AI systems such as large neural language models (\ptlms).
% In light with EMNLP 2022's Theme of open questions in NLP, 
In line with EMNLP 2022's theme, we examine the open research question of whether and how much \ptlms---which are the backbone of most modern NLP systems---exhibit social intelligence and \ToM abilities.
% In this work, we examine and discuss the extent to which \ptlms exhibit social intelligence and \ToM abilities.
% \oldText{We show that out-of-the-box \ptlms struggle at two types of reasoning abilities that are requisites for \TheoryOfMind (shown in Fig.~\ref{fig:intro-fig}), using some of the largest \ptlms in existence \cite[\gptT;][]{brown2020language}}.
Using some of the largest English models in existence \cite[\gptT;][]{brown2020language}, we demonstrate that out-of-the-box \ptlms struggle at two types of reasoning abilities that requisites for \TheoryOfMind (shown in Fig.~\ref{fig:intro-fig}). 
\newText{We argue that these reasoning abilities are necessary but not sufficient for \TheoryOfMind, and that larger models will likely provide upper bounds on what equivalent-but-smaller models are capable of.}

% We show that out-of-the-box large LMs (\ptlms) such as \gptT \cite{brown2020language} severely lack these abilities, showing that they struggle on two \TheoryOfMind tasks.
% \maarten{Potentially here, say that these two reasoning types are necessary but not sufficient? Or say it elsewhere?}

We first assess whether \ptlms can reason about \textit{social commonsense and emotional intelligence} with respect to social interactions (\S\ref{sec:socialIQa}), using the \socialIQa benchmark \cite{sap2019socialIQa} illustrated in Fig.~\ref{fig:intro-fig} (top).
Results show our best performing few-shot \gptT setup achieving only 55\% accuracy, lagging $>$30\% behind human performance.
Furthermore, social reasoning about the protagonists of situations is easier for \gptT (5-15\% absolute difference) compared to reasoning about other secondary participants.

Second, we measure \ptlms' ability to \textit{understand other people's mental states and realities} in short stories (\S\ref{sec:ToMi}).
We use the \tomi QA benchmark \cite[illustrated in Fig.~\ref{fig:intro-fig}; bottom;][]{le-etal-2019-revisiting}, which was inspired by the classic Sally-Ann False Belief \TheoryOfMind test \cite{baron1985does}.
% uses the \tomi benchmark \cite{le-etal-2019-revisiting} to measures \ptlms' ability to \textit{understand other people's mental states and realities} in short stories (\S\ref{sec:ToMi}; illustrated in Fig. \ref{fig:intro-fig}; right), inspired by the classic Sally-Ann False Belief \TheoryOfMind test \cite{baron1985does}.
% , this benchmark contains stories about people and objects, along with questions about where an object is located (either in reality or where people think
Here, our results show that \gptT models peak at 60\% accuracy on questions about participants' mental states, compared to 90--100\% on factual questions.
% \maarten{Unclear that we need this extra sentence?}
% Surprisingly, \gptT appears equally puzzled by questions about people's mental states that require \ToM (i.e., false beliefs) vs. that require no \ToM to answer (i.e., true beliefs).
% Our results show that \gptT models perform 10-15\% worse than random chance on questions that require \ToM reasoning about false beliefs, despite being able to answer non-\ToM questions with reasonable performance.
% Surprisingly, we find that as \gptT models grow larger, their performance difference on questions that require \ToM and don't increases \ronan{not so clear}, compared to smaller models that are generally closer to random performance.

Our novel insights show that reasoning about social situations and false beliefs still presents a significant challenge for large language models, despite their seemingly impressive performance on tasks that could require social intelligence (e.g., story generation, dialogues).
% These investigations present novel insights into \gptT's \TheoryOfMind abilities and shortcomings, showing that reasoning about social situations and false beliefs still presents a significant challenge for large language models despite their impressive performance.
In \S\ref{sec:discussion-conclusion}, we first examine these shortcomings; drawing on \oldText{pragmatics theories}\newText{theories of the pragmatics of language}, 
% by examining the assumptions underlying \ptlms's data, design, and training paradigms.
% We speculate
we speculate that the type of texts in \ptlms' training datasets could substantially limit learning social intelligence. %, drawing on theories from pragmatics.
Then, we outline some possible future directions towards socially aware \ptlms, reflecting on the feasibility of interactional data selection, person-centric inductive biases, and interaction-based language learning.
% Drawing on theories from pragmatics, we discuss these shortcomings by examining the assumptions underlying \ptlms's data, design, and training paradigms,
% and outline possible future directions towards socially aware \ptlms.
Our findings suggest that only increasing the scale of \ptlms is likely not the most effective way to create socially aware AI systems, challenging a prevalent narrative in AI research \cite{Narang2022palmsBlog}.

\section{\TheoryOfMind \& Large LMs}
\label{sec:background}

% \maarten{Re-organizing this section to focus more on why \ptlms need \ToM and less on ToM background.}

\paragraph{Why do \ptlms need \TheoryOfMind?}
% \label{ssec:why-ptls-need-tom}
Social intelligence, \TheoryOfMind, and commonsense reasoning have been a longstanding but elusive goal of artificial intelligence for decades \citep{Gunning2018-xn,Choi2022-sv}.
These reasoning abilities are becoming increasingly necessary as AI assistants are used in situations that require social intelligence and Theory of Mind in order to operate effectively \cite{Wang2007-xa,Dhelim2021-sd,Langley2022-fl}.
For example, new technologies are emerging where AI is used to \textit{interact} and \emph{adapt} to users \cite{Bickmore2005-relationships,Jaques2019-gp}, e.g., voice assistants, and tutoring systems; or where AI helps \textit{enhance communication} between multiple users, e.g., email autocomplete \citep{Chen2019smartCompose}, AI-assisted counseling 
\cite{Kearns2020-ty,Allen2020-cd,sharma2021towards}, or facilitated discussion \cite{Rose2014-mooc-attrition}.

As we move beyond just asking single-turn questions to social and interactive AI assistants, % (e.g., about the weather),
higher-order reasoning becomes necessary \cite{McDonald2019-fy}.
For example, AI systems should be capable of more nuanced understanding, such as ensuring an alarm is on if someone has a job interview the next morning \cite{Dhelim2021-sd}, 
knowing to call for help when an elderly person falls \cite{pollack2005intelligent}, inferring personality and intentions in dialogues \cite{Mairesse2007-personality,wang-etal-2019-persuasion}, reasoning about public commitments \cite{Asher2013-strategic-conversation},
predicting emotional and affective states \cite{litman-forbes-riley-2004-predicting,jaques-etal-2020-human},
and incorporating empathy, interlocutor perspective, and social intelligence \cite{Kearns2020-ty,sharma2021towards}.

\paragraph{What is \TheoryOfMind?}
\TheoryOfMind (\ToM) describes the ability that we, as humans, have to ascribe and infer the mental states of others, and to predict which likely actions they are going to take \cite{apperly2010mindreaders}.%
\footnote{While Theory of Mind is well developed in most adults \cite{ganaie2015study}, reasoning and inference capabilities can be influenced by age, culture, neurodiversity, or developmental disorders \cite{korkmaz2011theory}.}
This ability is closely related to (interpersonal) social intelligence \cite{ganaie2015study}, which allows us to navigate and understand social situations ranging from simple everyday interactions to complex negotiations \cite{gardner1995kids}.

\begin{figure}[t]
    \centering
    \includegraphics[width=0.98\columnwidth,trim=3cm 3cm 3cm 2.9cm,clip]{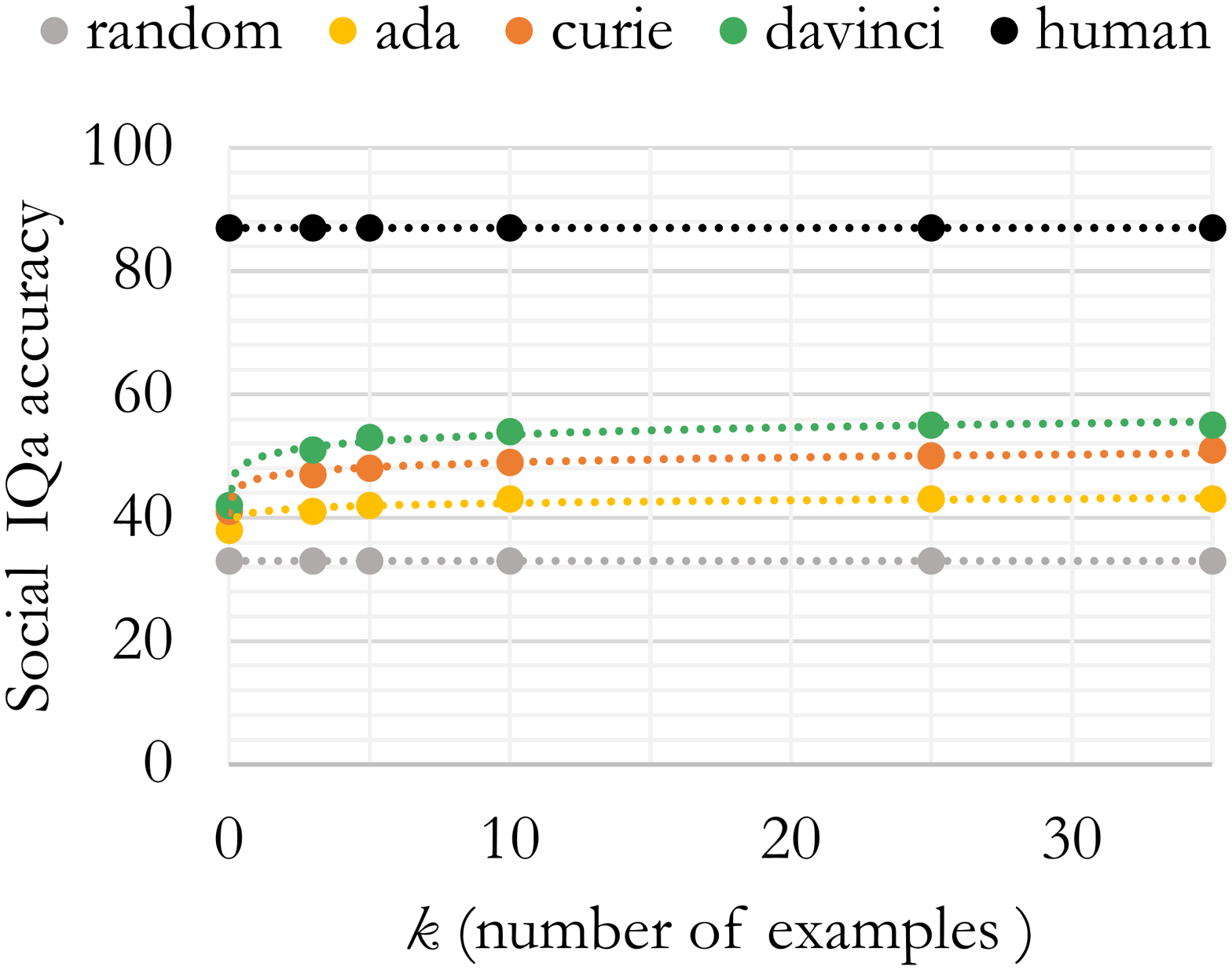}
    \caption{Accuracy on the \socialIQa dev. set, broken down by \ptlm model type and size, as well as number of few-shot examples ($k$).}
    \label{fig:siqa-acc-by-k-size}
    %\vspace{-1em}
\end{figure}

Interestingly, the development of \TheoryOfMind and language seem to happen around similar ages in children \cite{Sperber1986-ha,Wellman1992-zj,Miller2006-pg,Tauzin2018-na}.%
\footnote{The direction of the \ToM-language association is still debated \cite{De_Villiers2007-yj}.
Some researchers believe language development enables \ToM-like abilities \cite{Pyers2009-up,Rubio-Fernandez2021-gm}.
On the other hand, some argue that language develops after \ToM since preverbal infants already could possess some level of \ToM-like abilities \cite{Onishi2005-pm,Southgate2014-gi,Poulin-Dubois2018-td}.}
\oldText{Pragmatics theories}\newText{Theories of the pragmatics} of language and communication can frame our understanding of this link \cite{Rubio-Fernandez2021-gm}, positing that one needs to reason about an interlocutor's mental state (\ToM) to effectively communicate and understand language \cite{grice1975logic,Fernandez2013-su,Goodman2016-rsa,Enrici2019-mf}.%
\footnote{Most cognitive studies on this subject focus on the English language, which is not representative of the wide variation of language structures, and thus limits the cognitive conclusions one can draw about the link between language and \TheoryOfMind \cite{Blasi2022-bi}.}

\section{\socialIQa: Do \ptlms have Social Intelligence and Social Commonsense?}
\label{sec:socialIQa}
%\maarten{\socialIQa is a 3-way question answering (QA) task \cite{sap2019socialIQa}}

\newcommand{\atomic}{\textsc{Atomic}\xspace}
A crucial component of Theory-of-Mind is the ability to reason about the intents and reactions of participants of social interactions.
To measure this, we use the dev. set of the \socialIQa QA benchmark \cite{sap2019socialIQa}, which was designed to probe social and emotional intelligence in various everyday situations. 
This benchmark covers questions about \textit{nine} social reasoning dimensions, drawn from the \atomic knowledge graph \cite{sap2019atomic}.
% This benchmark was built on top of the \atomic \cite{sap2019atomic} knowledge graph, which distills social commonsense inferences about everyday events into \textit{nine} reasoning dimensions.

\socialIQa instances consist of a context, question, and three answer choices, written in English.
Each question relates to a specific reasoning dimension from \atomic: six dimensions focus on the pre- and post-conditions of the \textit{agent} or protagonist of the situation (e.g., needs, intents, reactions, next actions), and three dimensions focus on the post-conditions of \textit{other} participants involved in the situation (reaction, next action, effect).
In total, there are 1954 three-way QA tuples; see Tab.~\ref{tab:siqa-examples} for examples, and Tab.~\ref{tab:siqa-stats} in Appendix \ref{supp:siqa-deets} for per-dimension counts.

% Each question maps onto one of the \atomic reasoning dimensions, covering questions about motivations of participants of situations (xIntent: ``\textit{Why did Austin do this?}'') as well as the emotional reactions of participants (oReact: ``\textit{How would Taylor feel afterwards?}'' and xWant: ``\textit{What would Kai want to do after?}'').

% \maarten{One or two more sentences about how \socialIQa was created (crowdsourcing, ATOMIC, AF-lite) \cite{sap2019atomic}}

\begin{figure}[t]
\centering
    \includegraphics[width=0.98\columnwidth,,trim=3cm 3cm 3cm 2.9cm,clip]{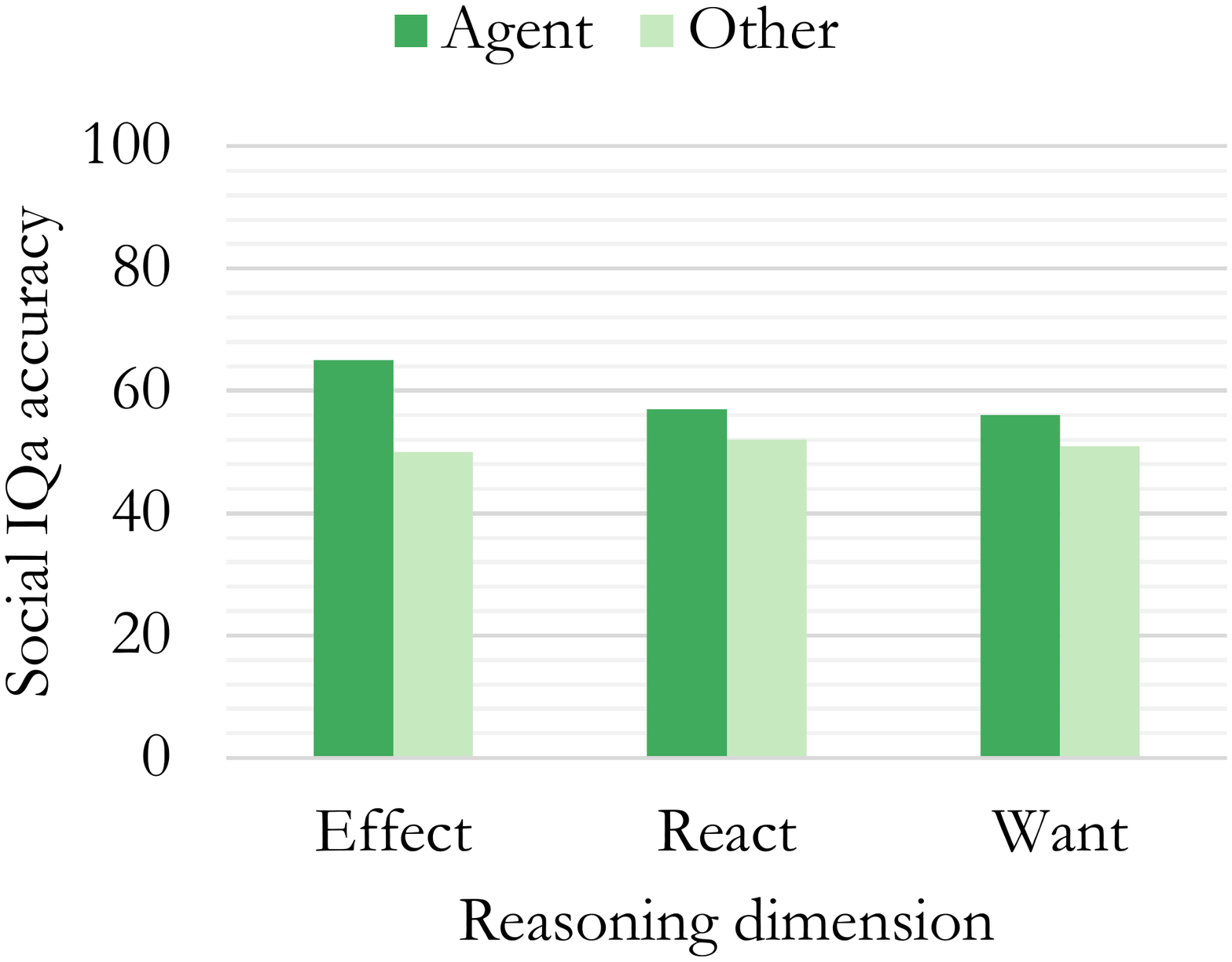}
    \caption{Comparing the accuracy of \gptDavinci (35-shot) on \socialIQa when the reasoning is about the main agent of the situation versus others. 
    %  \ronan{Can we show that it's not inherently harder? i.e. the human performance doesn't significantly decrease as the focus changes.}\maarten{Unfortunately, I don't have access to the human eval data anymore :(}
    %\vspace{-1em}
    }
    \label{fig:siqa-acc-by-char}
\end{figure}

\newcommand{\sitColWidth}{6.1cm}

\begin{table*}[t]
\centering
% \footnotesize
\small
% \scriptsize
% \renewcommand{\arraystretch}{1.15}
% \setlength{\tabcolsep}{3pt}

\begin{tabular}{@{}lll@{}lp{5.2cm}l@{}}
\toprule
& \textbf{Situation} & & & \textbf{Answers} & \textbf{Focus} \\ \toprule

\multirow{3}{*}{(a)} & \multirow{3}{\sitColWidth}{Remy was working late in his office trying to catch up. He had a big stack of papers. What does Remy need to do before this?} & \personEmoji & & Needed to be behind & \multirow{3}{*}{Agent} \\ 
 && & & Be more efficient & \\ 
 && & \robotEmoji & Finish his work & \\ \midrule
\multirow{3}{*}{(b)} & \multirow{3}{\sitColWidth}{Casey wrapped Sasha's hands around him because they are in a romantic relationship. How would you describe Casey?} & \personEmoji & & Very loving towards Sasha & \multirow{3}{*}{Agent} \\ 
 && & \robotEmoji & Wanted & \\ 
 & && & Being kept warm by Sasha & \\ \midrule
\multirow{3}{*}{(c)} & \multirow{3}{\sitColWidth}{Tracy held a baby for 9 months and then gave birth to addison. What will happen to Tracy?} & & & Throw her baby at the wall & \multirow{3}{*}{Agent} \\ 
 && & \robotEmoji & Cry & \\ 
 && \personEmoji & & Take care of her baby & \\ \midrule
\multirow{3}{*}{(d)} & \multirow{3}{\sitColWidth}{Kai gave Ash some bread so they could make a sandwich. How would Kai feel afterwards?} & \personEmoji & & Glad they helped & \multirow{3}{*}{Agent} \\
 && & & Good they get something to eat & \\ 
 && & \robotEmoji & Appreciative & \\ \midrule
 
 \multirow{3}{*}{(e)} & \multirow{3}{\sitColWidth}{Aubrey was making extra money by babysitting Tracey's kids for the summer. What will Tracy want to do next?} & & \robotEmoji & Save up for a vacation & \multirow{3}{*}{Others} \\ 
 && \multirow{1}{*}{\personEmoji} & & Let Aubrey know that they are appreciated & \\ 
% &&&& appreciated & \\
 && & & Pay off her college tuition & \\ \midrule
 		
\multirow{3}{*}{(f)} & \multirow{3}{\sitColWidth}{The people bullied Sasha all her life. But Sasha got revenge on the people. What will the people want to do next?} & & & Do whatever Sasha says & \multirow{3}{*}{Others} \\ 
 && & \robotEmoji & Get even & \\ 
 && \personEmoji & & Flee from Sasha & \\ \midrule
% \multirow{3}{*}{(f)} & \multirow{3}{\sitColWidth}{Aubrey met a creepy stranger at the park who was trying to get Aubrey to go home with them. What will happen to others?} & & \robotEmoji & Be kidnapped & \multirow{3}{*}{Others} \\ 
%  && \personEmoji & & Want to stop the stranger & \\ 
%  && & & Be saved by police & \\ \midrule
\multirow{3}{*}{(g)} & \multirow{3}{\sitColWidth}{After everyone finished their food they were going to go to a party so Kai decided to finish his food first. What will others want to do next?} & \personEmoji & & Eat their food quickly & \multirow{3}{*}{Others} \\ 
 & && \robotEmoji & Throw their food away & \\ 
 & && & Go back for a second serving & \\ \midrule
\multirow{3}{*}{(h)} & \multirow{3}{\sitColWidth}{Aubrey fed Tracy's kids lunch today when Tracy had to go to work. What will happen to Aubrey?} & & & Be grateful & \multirow{3}{*}{Agent} \\ 
 && \personEmoji & \robotEmoji & Get paid by Tracy & \\ 
 & && & Get yelled at by Tracy & \\ \midrule
\multirow{3}{*}{(i)} & \multirow{3}{\sitColWidth}{Sasha was the most popular girl in school when she accepted Jordan's invitation to go on a date. What will Jordan want to do next?} & & & Plan a best friends outing with Sasha & \multirow{3}{*}{Others} \\ 
 && \personEmoji & \robotEmoji & Plan a romantic evening with Sasha & \\ 
 && & & Go on a date with Valerie & 
 \\ \bottomrule
\end{tabular}
\caption{Examples of \socialIQa questions, which person the questions focus on (\emph{Agent}, \emph{Others}), and the human gold answers (\personEmoji) and \gptDavinci predictions (\robotEmoji).} \vspace{1em}
\label{tab:siqa-examples}
\end{table*}

\subsection{Probing \ptlms with \socialIQa}
To probe our language models, we use a $k$-shot language probing setup, following \citet{brown2020language}.
We select the answer that has the highest likelihood under the language model conditioned on the context and question, as described in Appendix~\ref{supp:gpt3-deets}.

% we use the development set of the \socialIQa, which consists of 1,954 QA tuples.
% Following \citet{brown2020language}, we use a $k$-shot language probing setup to obtain predictions.
% Specifically, we concatenate the context ($c$) and question ($q$) together with proper punctuation, and assign the model prediction to the answer ($a_i$, $i\in{1,2,3}$) with the highest conditional likelihood under the language model: $ \argmax_i p_{\text{LM}}(a_i\mid c,q, \mathcal{C}_k)$
% where $\mathcal{C}_k$ denotes the $k$ training examples, for which we provide the context, question, and correct answer concatenated.\footnote{Note that we explored various probing setups and formats, such as QA-oriented formats and normalizing by marginal likelihood of each answer ($p_{\text{LM}}(a)$), but found very little difference in performance.}

% \paragraph{Experimental Details}
To test the limits of what the models can do, 
% As an upper bound of the types of reasoning that the model can do, 
we select $k$ examples that have the same \atomic reasoning dimension as the question at hand, varying $k$ from 0 to 35 in increments of 5.
We use three \gptT model sizes: \gptAda (smallest), and \gptCurie and \gptDavinci (two largest).
% \maarten{And say which model we use.}

\subsection{\socialIQa Results}\label{ssec:siqa-results}
Shown in Fig.~\ref{fig:siqa-acc-by-k-size}, \gptT models perform substantially worse than humans (>30\% less) on \socialIQa,\footnote{
We find similar results when using \textsc{InstructGPT} \cite{ouyang2022gpt3instruct} instead of \gptDavinci.} and also worse than models finetuned on the \socialIQa training set \cite[>20\%;][]{lourie2021unicorn}.\footnote{\citet{lourie2021unicorn} achieves 83\% on the test set, as shown on the \href{https://leaderboard.allenai.org/socialiqa/submission/bsd31epbvhc9b55n46g0}{AI2 \socialIQa leaderboard}.}
\oldText{Not surprisingly, \gptDavinci reaches higher accuracies than \gptAda and \gptCurie, showing that despite model size improving performance, gains are small.}
\newText{Although it is not surprising that \gptDavinci reaches higher accuracies than \gptAda and \gptCurie, the gains are small, which suggests that increasing model size might not be enough to reach human-level accuracy. 
These findings are in line with recent BIG-Bench results on \socialIQa with the BIG-G \cite[128B parameters;][]{Srivastava2022BeyondTI} and PaLM \cite[353B parameters;][]{Chowdhery2022palm} \ptlms, which lag behind humans with 45\% and 73\% accuracy, respectively (see Fig. \ref{fig:bigbench-socialIQa} in Appendix \ref{ssec:siqa-results-more}).}

% Additionally, our results are corrobated by the contemporaneously released BIG-Bench work \cite{Srivastava2022BeyondTI}, which show $<$50\% accuracy in their best performing setting on \socialIQa (see Fig. \ref{fig:bigbench-socialIQa} in Appendix \ref{ssec:siqa-results-more}).}
%
% \maarten{Potentially add one sentence about BIG-Bench here?}

Focusing on \gptDavinci, while increasing the number of examples $k$ improves performance, the differences are marginal after $k$=10 examples (only 1\% increase from 10 to 35 examples).
\newText{This suggest that performance either plateaus or follows a logarithmic relationship with increasing number of conditioning examples.}
\oldText{This suggests a logarithmic relationship between conditioning on more examples and performance improvements.}

Finally, we examine the differences in \gptDavinci with respect to which participant is the focus.
Shown in Fig.~\ref{fig:siqa-acc-by-char}, we find that \gptDavinci performs consistently better on agent-centric questions, compared to other-oriented questions.
Shown in the example predictions in Tab.~\ref{tab:siqa-examples}, \gptDavinci often confuses which participant is being asked about.
In example (e), % in Tab. \ref{tab:siqa-examples}, 
after Aubrey babysat for Tracy, \gptDavinci fails to predict that Tracy will likely want to ``\textit{let Aubrey know they are appreciated},'' and instead mistakenly predicts that Tracy will want to ``\textit{save up for vacation},'' which is what Aubrey would likely do.
\gptDavinci displays a similar participant confusion in example (f) in Tab.~\ref{tab:siqa-examples}.

\section{\tomi: Can \ptlms Reason about Mental States and Realities?}
\label{sec:ToMi}

\newcommand{\mind}{\textsc{Mind}\xspace}
\newcommand{\facts}{\textsc{Fact}\xspace}

\newcommand{\memory}{\facts-\textsc{Mem}\xspace}
\newcommand{\reality}{\facts-\textsc{Real}\xspace}
\newcommand{\firstorder}{\mind-1st\xspace}
\newcommand{\secondorder}{\mind-2nd\xspace}
\newcommand{\falseBelief}{\textsc{Fb}\xspace}
\newcommand{\trueBelief}{\textsc{Tb}\xspace}

\newcommand{\mindFB}{\mind-\falseBelief\xspace}
\newcommand{\mindTB}{\mind-\trueBelief\xspace}

Another key component of \TheoryOfMind is the ability to reason about mental states and realities of others, recognizing that they may be different than our own mental states.
% \maarten{Shorten this and rely more on Figure 1 and Table 2.}
As a measure of this ability in humans, psychologists developed the \textit{Sally Ann false-belief test} \cite{wimmer1983beliefs}, %which depicts stories of a person moving an object while someone else is out of the room, such as those in Fig.~\ref{fig:intro-fig} (bottom) and Tab.~\ref{tab:tomi-examples}.
% These stories require \TheoryOfMind to understand that the person who was absent will not know the object's new location.
in which two people (Sally and Ann) are together in a room with a ball, a basket, and a box, and while Sally is away, Ann moves the ball from the basket to the box.
When asked where Sally will look for her ball, \TheoryOfMind allows us to infer that Sally will look in the basket (where she left the ball), instead of in the box (where the ball is, unbeknownst to Sally).

% To assess whether AI systems could pass this false belief test, \citet{grant2017can} and \citet{nematzadeh2018evaluating} used rule-based logic to create a dataset of such situations.

To measure the false-belief abilities of \ptlms, we use the \tomi QA dataset of English Sally-Ann-like stories and questions \cite{le-etal-2019-revisiting}.\footnote{\tomi is a more challenging version of the rule-based datasets by \citet{nematzadeh2018evaluating} and \citet{grant2017can}, as it contains randomly inserted distractor actions that prevent trivial reverse engineering.}
\tomi stories were created using a stochastic rule-based algorithm that samples two participants, an object of interest, and a set of locations or containers, and weaves together a story that involves an object being moved (see Tab.~\ref{tab:tomi-examples}).
All questions have two possible answers: the original object location, and the final object location. % , making this a two-way QA task.

\begin{figure}[t]
    \centering
    \includegraphics[width=.99\columnwidth,trim=0cm .2cm 0cm 0cm,clip]{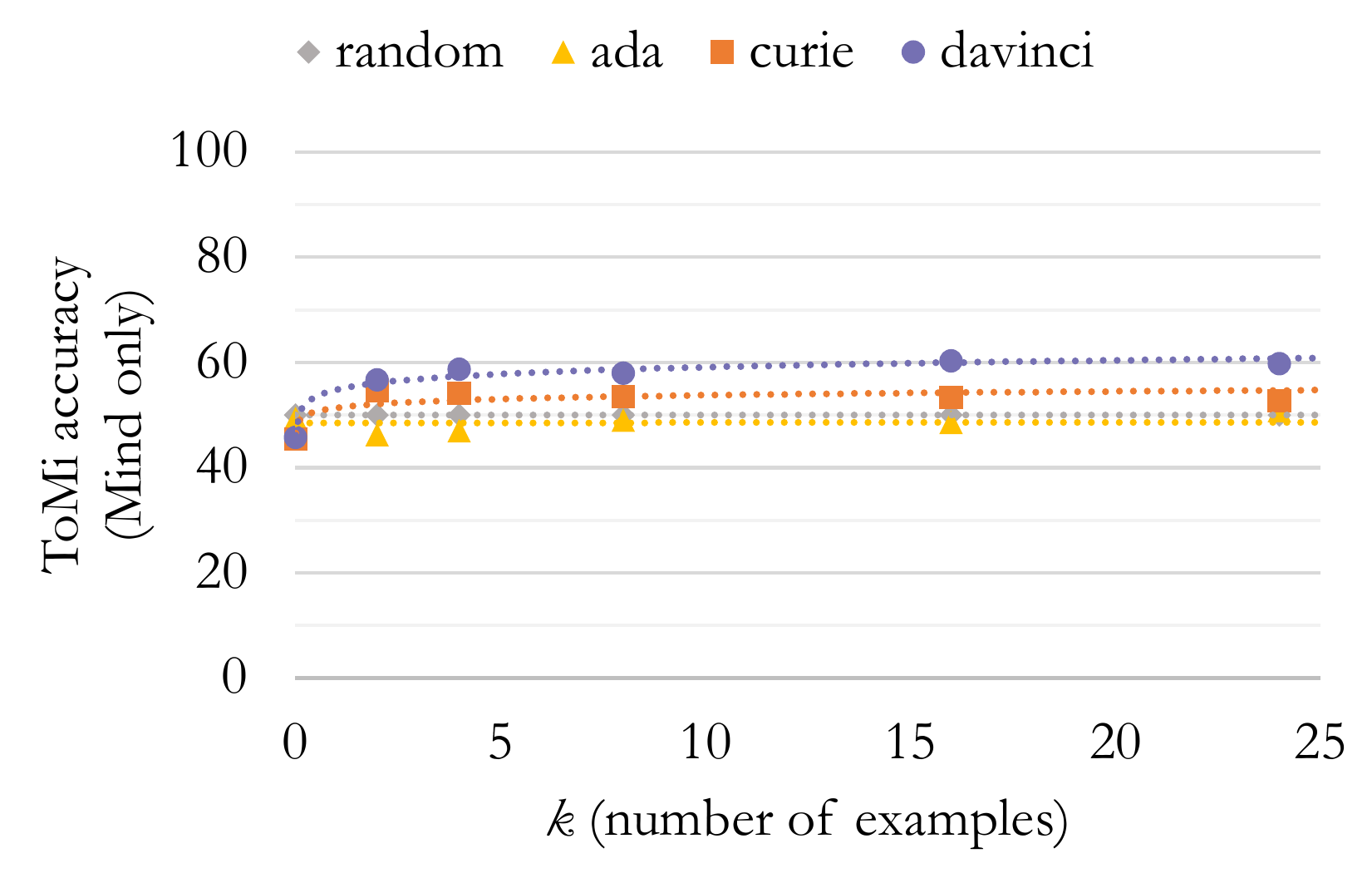}
    \caption{Accuracy on the \tomi dev. set \mind questions of varying sizes of \gptT, and with varying number of examples ($k$).}
    \label{fig:tomi-acc-by-size}
\end{figure}

We investigate how \ptlms answer the \tomi story-question pairs, distinguishing between questions about factual object locations (\facts) and questions about where participants think objects are located (i.e., their mental states; \mind).
The \facts questions either ask about the object's original (\memory) or final (\reality) location.
The \mind questions cover first-order (e.g., ``\textit{where will Abby look for the object?}''; \firstorder) and second-order beliefs (e.g., ``\textit{where does James think that Abby will look for the object?}''; \secondorder). 
We further distinguish the \mind questions between true belief (\trueBelief) and false belief (\falseBelief), i.e., stories where a participant was present or absent when an object was moved, respectively. 
% Stories vary in whether there are participants with false or true beliefs, i.e., whether participants were absent or present when an object was moved. 
% As such, \mind questions could be about true beliefs (participant was present when the object was moved; \trueBelief) or false beliefs (participant was absent and still thinks the object is in the original location; \falseBelief).

Importantly, answering the \mind questions requires \TheoryOfMind and reasoning about realities and mental states of participants---regardless of the true- or false-belief setting---whereas \facts questions do not require such \ToM.
There are a total of 1861 two-way QA pairs in our \tomi probe set, with 519 \facts and 1342 \mind questions (see Tab.~\ref{tab:tomi-stats} in Appendix \ref{supp:tomi-details} for more detailed counts).

\begin{figure}[t]
    \centering
    \includegraphics[width=.99\columnwidth,trim=3.5cm 3cm 3cm 2.8cm,clip]{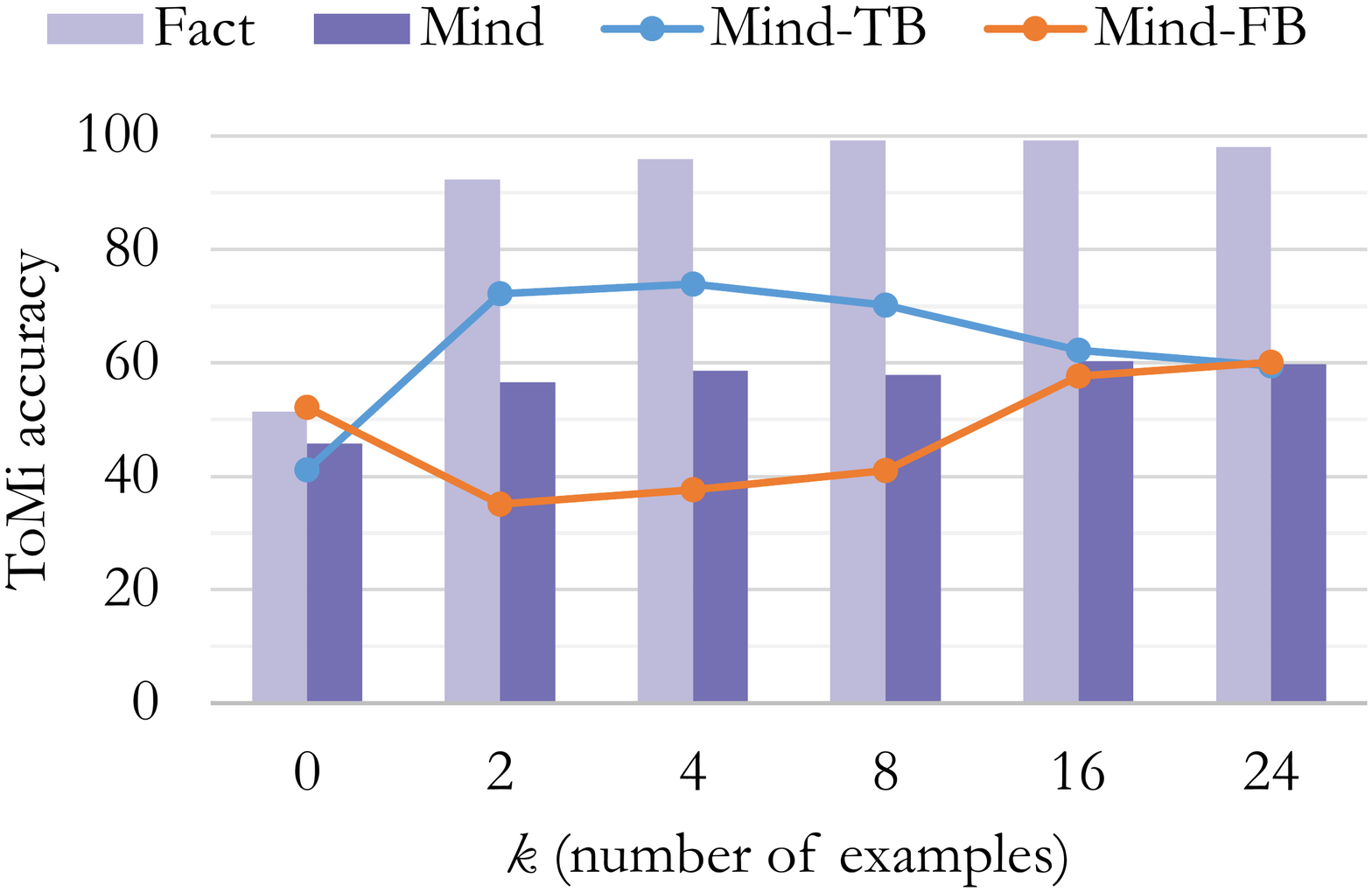}
    \caption{Accuracy of \gptDavinci by number of examples ($k$), by reasoning type (\facts vs. \mind; \mind-\trueBelief vs. \mind-\falseBelief).}
    \label{fig:tomi-acc-by-mind}
    % \vspace{-1em}
\end{figure}

\newcommand{\storyColWidth}{8.25cm}
\newcommand{\questionColWidth}{2.05cm}

\newcommand{\firstO}{M-1\xspace}
\newcommand{\secondO}{M-2\xspace}

\begin{table*}[t]
\centering
%\footnotesize
\small

\begin{tabular}{@{}cclll@{}ll@{}}
 \toprule
& Type & Story & Question &  &  & Answers \\ \midrule

 \multirow{3}{*}{(a)} & \multirow{3}{*}{\rotatebox{90}{\facts}} &
 \multirow{3}{\storyColWidth}{Sophia entered the study. Noah entered the study. The dress is in the treasure chest. Noah exited the study. Hannah entered the garden. Sophia moved the dress to the box.} &  & & \\
 && & \multirow{2}{\questionColWidth}{Where is the dress really?} & \personEmoji & \robotEmoji & box \\
 & & & &  &  & treasure chest \\ \midrule

 \multirow{3}{*}{(b)} & \multirow{3}{*}{\rotatebox{90}{\firstO-\falseBelief}} & 
 \multirow{3}{\storyColWidth}{Noah entered the garden. Nathan entered the garden. Evelyn likes the pumpkin. The banana is in the basket. Nathan exited the garden. Noah moved the banana to the suitcase.} &  
 \multirow{3}{\questionColWidth}{Where will Nathan look for the banana?} & & \\ 
 & && & \personEmoji & \robotEmoji & basket  \\
 & & & & & & suitcase \\ \midrule
 
 \multirow{4}{*}{(c)} &  \multirow{4}{*}{\rotatebox{90}{\secondO-\trueBelief}} & 
 \multirow{4}{\storyColWidth}{Lily entered the patio. Aiden is in the patio. Mila entered the patio. Mila hates the radish. The coat is in the box. Aiden moved the coat to the crate. Mila exited the patio.} &  
 \multirow{4}{\questionColWidth}{Where does Aiden think that Mila searches for the coat?} & & & \\
 & & &&  &  & \\
 & &&& \personEmoji & \robotEmoji & crate \\
 &  &&&  &  & box \\  \midrule

 \multirow{3}{*}{(d)} & \multirow{3}{*}{\rotatebox{90}{\firstO-\trueBelief}} & 
 \multirow{3}{\storyColWidth}{Elizabeth entered the cellar. Carter entered the cellar. The slippers is in the crate. Elizabeth moved the slippers to the container. Carter exited the cellar.} & 
 \multirow{3}{\questionColWidth}{Where will Carter look for the slippers?} & & & \\
 & &&& \personEmoji &  & container \\
 &  &&&  & \robotEmoji & crate \\    \midrule 
 
 \multirow{4}{*}{(e)} & \multirow{4}{*}{\rotatebox{90}{\firstO-\falseBelief}} &
 \multirow{4}{\storyColWidth}{Evelyn entered the living room. Jackson entered the playroom. James entered the playroom. The beans are in the treasure chest. James exited the playroom. Jackson moved the beans to the pantry. Jackson exited the playroom. James entered the living room.} &  
  & & & \\
 &&  & \multirow{3}{\questionColWidth}{Where will James look for the beans?} & & &  \\

 & && & \personEmoji &  & treasure chest  \\ 
 & &&  &  & \robotEmoji & pantry \\   \midrule

 \multirow{4}{*}{(f)} & \multirow{4}{*}{\rotatebox{90}{\secondO-\falseBelief}} &
 \multirow{4}{\storyColWidth}{Isla likes the potato. Ella entered the laundry. Oliver entered the laundry. The slippers are in the box. Ella exited the laundry. Oliver moved the slippers to the basket. Isla entered the office.} &  \multirow{4}{\questionColWidth}{Where does Ella think that Oliver searches for the slippers?} & & & \\
 &  &  &  &  \\
 & && & & \robotEmoji & basket \\ 
 & & && \personEmoji &  & box \\  \bottomrule

	%	pantry	treasure chest

\end{tabular}
\caption{Example stories in the \tomi dev. dataset, with \gptDavinci predictions (with $k$=16 examples) and gold answers.
``Type'' denotes reasoning type, \firstO and \secondO denote \firstorder and \secondorder, resp.
% \daniel{I might be confused (or have bad ToM!) but the true answer for the first M-1-TB seems wrong? Nathan last saw the banana in the basket. And M-1-FB too, since  James last saw the beans in the treasure chest.}\maarten{Good catch, these were my errors from copy pasting!}
\vspace{1em}
}
\label{tab:tomi-examples}
\end{table*}

\subsection{Probing \ptlms with \tomi}

We use the $k$-shot probing setup to test this \ToM component in \ptlms, with $k \in \{2,4,8,16,24\}$.
We select $k$ examples of the same reasoning type (i.e., \memory, \firstorder, etc.), ensuring a 50-50 split between true- and false-belief examples for the \mind questions.
% the \mind examples such that half are true belief and half are false belief. 
% \maarten{Maybe add some data stats?}
As before, we test \gptAda, \gptCurie, and \gptDavinci.

\subsection{\tomi Results}\label{ssec:tomi-results}

% \maarten{Yejin: weird to first highlight 75\% when in the abstract we highlight the difference between fact and mind questions... Maybe we can re-order the text (and charts) to focus first on reasoning type and then the scale experiments? Or maybe we can even just plot the accuracy by model size for only the mind questions? And we can move the overall accuracy plot to the appendix.}

Shown in Fig.~\ref{fig:tomi-acc-by-size}, our results indicate that \gptT models struggle substantially with the \tomi questions related to mental states (\mind), reaching 60\% accuracy in the best setup.
As expected, the best performance is reached with \gptDavinci compared to smaller models which do not surpass 55\% accuracy; however, as before, the gains from scaling up \gptT are very small.
% As expected, the largest model \gptDavinci performs the best, compared to smaller models which do not surpass 55\%; however, 
Similarly, increasing the number of few-shot examples beyond $k=4$ does not substantially improve performance, corroborating findings on \socialIQa.

% With a maximum accuracy of 60\%, the largest model (\gptDavinci) does 
% \gptDavinci struggles substantially with correctly answering the \tomi questions related to mental states (\mind), achieving less 
% , achieving less than 75\% accuracy.
% As expected, larger models (71\% with \gptDavinci) perform better than smaller models (57\% with \gptAda), and increasing the number of examples $k$ does not help substantially for $k>$8 (Figure \ref{fig:tomi-acc-by-size}).

% Examining questions by reasoning type, 
Further examining \gptDavinci with respect to question types, 
we show that the model struggles substantially more with questions about mental states (55--60\% for $k>0$) compared to factual questions (90--100\% for $k>0$; Fig.~\ref{fig:tomi-acc-by-mind}; columns).
% \maarten{Maybe we should plot that: bargraph of overall performance and fact and mind, with different model sizes ? }
Furthermore, the difference between performance on \mindTB and \mindFB questions shows an interesting pattern when conditioning on an increasing number of examples $k$ (Fig.~\ref{fig:tomi-acc-by-mind}; lines): \gptDavinci's \mindTB accuracy first increases, peaks at $k=4$, then decreases. 
This peak seems to be due to the model defaulting to the most recent object location (i.e., the correct \mindTB answer), as illustrated in example (e) in Tab. \ref{tab:tomi-examples}.
Apparent in Fig.~\ref{fig:tomi-pctReal-by-k} in Appendix \ref{supp:tomi-details}, this recency bias is a phenomenon that has been previously documented in \ptlms \cite{oconnor-andreas-2021-context}.
% this seems to be due to the model defaulting to the most recent object location (i.e., the correct \mindTB answer), which has been described as recency bias \cite{oconnor-andreas-2021-context}.
% This shows that \gptDavinci first learns a strong prior to answer with the final object location (which is the answer for all \mindTB questions since all participants were present when the object was moved), and then with increasing $k$ starts to predict the original object location (answer for \mindFB questions).\maarten{Just say recency bias and leave it at that.}
% \daniel{the reason for this is unclear to me, since I'm assuming Mind-TB and Mind-FB are equally common in the prompts? IIRC sensitivity to the last thing mentioned is part of the reason, which would explain the Mind-TB prior but unclear on how this fits with Mind-FB, unless Mind-FB are more common in the prompts.}
% \daniel{how robust are these results to different choices and orderings of the prompts?}\maarten{Unclear, but the ordering is randomized for each example, so probably not that important?}
In general, \gptDavinci's comparably poor performance for \mindTB and \mindFB questions at $k>8$ suggests that it cannot properly answer questions about participants' mental states and realities.
% Overall, adding additional examples beyond $k$=16 does not increase performance (Figure \ref{fig:tomi-acc-by-size}).
% Examining \gptDavinci's performance more closely, we find that 

% Additionally, \gptDavinci performs somewhat worse on \secondorder compared to \firstorder (Fig. \ref{fig:tomi-acc-by-examples-order} in Appendix \ref{supp:tomi-details}), mirroring how humans struggle with increasingly higher-order \ToM questions \cite{Valle2015-yb}.

% \maarten{
% 1) Larger models are better than smaller models.
% 2) Looking at the best performing model, factual questions (\facts) reach nearly 100\%, but \mind questions barely reach 60\%.
% 3) Interesting trend of adding more examples, where \gptDavinci shifts from a strong prior to predict the real location which is often the answer for \trueBelief questions, to a more muddled confusion about both \trueBelief and \falseBelief.
% }

% \clearpage

\section{Discussion: Towards NLP with Neural Theory of Mind}
\label{sec:discussion-conclusion}
% \maarten{Comments from IRL meeting:
% Frame the discussion more positive, as exciting future directions. How can we bias models towards having ToM? data, arch/representations, training objective
% }

Most humans develop social intelligence and \TheoryOfMind naturally.
However, in this work, we showed that these abilities do not emerge automatically in large-pretrained language models. 
These shortcomings contrast with the wealth of successes of \ptlms at a variety of tasks, including tasks that potentially require social intelligence. 
For example, \gptT has been shown to generate stories with emotional arcs that are virtually indistinguishable from human-written stories \cite{clark2021all}.
Additionally, recent work has used \gptT to generate social commonsense knowledge related to protagonists of situations \cite{west2022symbolic}.
While those findings suggest some level of social and emotional intelligence in \ptlms, our explorations highlight the limits of these abilities, and raise the open question: \textit{how can we create NLP systems with true social intelligence and \TheoryOfMind?}

To begin answering this question, we first discuss the current \ptlms training paradigm (\S\ref{ssec:limitations-of-paradigm}), drawing from \oldText{pragmatics theories}\newText{theories of pragmatics} to examine why these models are not learning social intelligence efficiently.
Then, we outline some possible future directions to bias models towards \TheoryOfMind (\S\ref{ssec:future-directions}), through person-centric neural architectures, data selection, and training objectives.
% discuss pragmatics-informed hypotheses to explain these limitations, as well as some possible future directions to bias models towards \TheoryOfMind.

% \maarten{Focus on how social intelligence is a goal, maybe re-iterate some of the points from section 2 about how this is going to be necessary. Then discuss why this limitation is showing up;
% It's obvious that LMs should have social intelligence, and need to discuss that they kind of do things that require social intelligence already. However, there are still some big missing gaps.}

% conclusion paragraph, summarizing the findings briefly

% pop higher level, describe that these conclusions could be somewhat surprising in light of recent successes of PTLMs at tasks that seem quite social intelligence related

% reiterate that true social intelligence is needed, and that there are some fundamental limitations 

% Despite massive successes in other domains, our work highlighted some shortcomings of \ptlms at reasoning about social intelligence and \TheoryOfMind.
% Specifically, \gptT's social intelligence as measured by \socialIQa lags behind humans (>30\%), and struggles to answer \tomi questions about mental states (55-60\%) compared factual questions (90--100\%).
% We also saw that participant-centric reasoning is a particular challenge for \gptT.
% On \socialIQa, it often got participants of situations confused, and performed better for questions related to the main character of situations.
% With \tomi, \gptT performed increasingly worse from factual to first-order to second-order mental state questions.

\newcommand{\static}{static\xspace}
\newcommand{\Static}{Static\xspace}

\subsection{The Pragmatics of ``\Static'' Text}
\label{ssec:limitations-of-paradigm}
To understand why \ptlms are still struggling with social intelligence, % and \ToM, %\TheoryOfMind, 
we examine \ptlms' training paradigm through the lens of \textit{pragmatics}.
As discussed in \S\ref{sec:background}, pragmatics provides a connection between language development and \TheoryOfMind \cite{Sperber1986-ha,Miller2006-pg,Tauzin2018-na}: learning to communicate effectively with language requires reasoning about what our interlocutor knows or does not know \cite{grice1975logic,Fernandez2013-su,Goodman2016-rsa,Enrici2019-mf}.\footnote{
Note here that, in contrast to other work \cite{Bender2020-kt,Bisk2020-xb}, we do not focus on whether \ptlms ``understand'' language, instead we examine whether \ptlms can answer questions about the emotions and mental states of participants of situations.}

% This limited \TheoryOfMind ability from \ptlms could be explained by examining \ptlms' training data of through a pragmatic lens.
% Specifically, 
% dfried: I'm not sure about `predominant', changed wording to hedge a bit.
\oldText{Humans' predominant use of language}\newText{One major use of language by people} is to communicate about relationships and personal experiences \cite{Clark1989-discourse,Dunbar1993-ew}.
This is fundamentally different from the training data of \ptlms, which consists of language found in what we call \textit{\static} texts: documents that are written for a general audience and are relatively self-contained and topically focused \cite[e.g., news articles, books, Wikipedia articles;][]{gao2020pile,dodge2021documentingC4}.
% \daniel{I like that \emph{\static} is concise, but I don't feel it captures all the limitations that we're outlining here. I also feel that general audience and topically focused don't exclude the other things we lay out - eg you could imagine topically focused/general audience conversations that have most/all of the features below. Maybe something like \emph{observational}, or \emph{egocentric}, or \emph{non-social}?} \maarten{How about ``unidirectional''?}
% Most \ptlms are trained on \textit{\static} text, which we define as documents that are written for a general audience and are relatively  self-contained and topically focused \cite[e.g., news articles, books, Wikipedia articles;][]{gao2020pile,dodge2021documentingC4}.
Such \static text is typically written such that readers only require the language itself as input, which they then combine with their world knowledge and commonsense to understand its meaning \cite{Graesser1994-cc}.

If AI systems are to learn social intelligence and \TheoryOfMind, we posit that \static text has certain limitations, from a pragmatics lens, outlined below. %  We outline those below.
% When it comes to learning social commonsense and social intelligence, we posit that the \static nature of this text creates certain limitations, through the lens of pragmatics.

\paragraph{Reporting bias.}
Following Grice's maxim of quantity \cite{grice1975logic}, \static text often avoids redundancy by omitting content that is known by both the author and the reader \cite{clark1991grounding}.
% \daniel{mention common ground, and cite a Clark paper?}\maarten{good idea, can you try incorporating? \cite{clark1991grounding} }
Also known as reporting bias \cite{Gordon2013reportingbias,Lucy2017AreDR}, this phenomenon likely limits \ptlms' ability to learn social commonsense knowledge from \static text.

% \paragraph{Lack of alternatives.}
\paragraph{Lack of communicative intent and alternatives.}
% \daniel{Thinking about this more in conjunction with another paper, I'm inclined to change this to change this to \emph{lack of communicative intent}, but still use the same examples we have (and end in the same way).}\maarten{I see your point, but wondering how to describe these concepts in this context... Tried below}
A corollary to reporting bias, \static text does not provide any direct access to \textit{communicative intent} (why words were used) or to \textit{alternatives} (which words were not used, and why).
% A corollary to reporting bias, \static text does not provide any direct access to \textit{alternative utterances} that the author could have chosen to write, but did not.
This reasoning about intents, alternatives, and their implications
% which are 
\oldText{
is a core component of how people draw pragmatic inferences about their interlocutor's mental states 
\cite{grice1975logic,Horn1984-implicature,Levinson2000-implicature}
}
\newText{is highly predictive of the pragmatic inferences people draw about their interlocutors \cite{Goodman2016-rsa}}
  --- for example, when someone answers \emph{Where does Taylor live?} with \emph{Somewhere in the U.S.}, it implies that they likely do not know or do not want to share the exact location, since, if they did, they would have been more specific.
% \maarten{Tried to add a sentence about how that could affect models, Daniel plz double check}
This poses a likely limitation that \ptlms only learn what words are used, but not which words were not used, and why.
% \maarten{One sentence about how this could affect models?}
% \maarten{Daniel, can you add a sentence about how \static text only contains the ``what happened'' piece of info, not the ``what else could have happened but didn't'' piece, i.e., the alternatives/counterfactuals that you brought up as being an important piece of pragmatics?}

\paragraph{Lack of communicative effects.} % \maarten{Daniel please double check that I'm using backchannel properly?}
% \daniel{perhaps change this to \emph{lack of communicative effects}. This would be a bit more general --- backchannel is one specific way that communicative effects are conveyed but there are others. It would also parallel the proposed structure for para 2 (could also reorder paras to put these next to each other)}\maarten{Done, sounds good!}
Language is primarily learned \cite{wells1981learning,Tomasello2005-intentions} and used \cite{Clark1996-using} in collaborative and interactive settings \cite{Clark1989-discourse}, which allow interlocutors to 
\oldText{backchannel, i.e.,} % dfried: backchannel is specific to speech settings, I think; we can leave out that phrase to make it more general
give immediate feedback to each other on whether their language was understood \cite{Clark2004-monitoring} or should be adjusted \cite{Krauss1966-feedback}, and observe the perlocutionary effects that their language has on their partners \cite{Austin1975-do-things-with-words}.
% \maarten{Again, tried to add a sentence about how this could affect models, Daniel plz double check}
Since \static text has no such feedback, \ptlms learn from all texts, as if they were all equally understandable by readers.
% \maarten{Potentially, can we make a third but succinct claim about how language primarily used for communication/interaction \cite{wells1981learning}; \static text is only one half of the speaker-listener dyad; and interactions allow for immediate feedback of whether what we're saying could/should be ajusted (i.e., perlocutionary aspects of language)? Daniel, if you could give that a go?}

\paragraph{Centering theory.}
% Yejin: Centering theory: surface text is gonna give information more on the person in the focus, and thus there will be more information on that person. 
% Learning to distinguish characters in \static text is challenging due to text often having a single focus or perspective.
At any given time, most text focuses on describing one protagonist and their relation to their surroundings, according to Centering Theory \cite{Grosz1995-hm}.
As such, main characters and their mental states are more likely to be \textit{described}, whereas other participants might only be \textit{mentioned}.
Additionally, main characters or protagonists are more likely to be referred to with pronouns, whereas secondary characters with their names.
% According to Centering Theory \cite{Grosz1995-hm}, the centrality of a character in a discourse is often marked by their being referred to with pronouns, whereas secondary characters are often referred to by name.

Thus, a model trained purely on \static text might not learn to reason about social intelligence or mental states and realities of different characters of situations; they might not even inherently learn to resolve coreference for multiple characters \cite{sakaguchi2020winogrande}.
\newText{In fact, challenges of coreference resolution could explain why \gptT models struggle on \socialIQa which contains questions with pronouns, and centering theory and main character biases in \static text could explain why models find non-protagonist questions more challenging. 
On the other hand, \tomi does not contain any pronouns, and thus requires social intelligence beyond coreference resolution.}
% \maarten{Maybe here, mention that this could explain \socialIQa results, since a non-trivial portion of the questions there have pronouns that need to be resolved to characters; however, \tomi does not have any coref issues since all characters are mentioned by name.}
% resolve coreference for multiple characters inherently \cite{sakaguchi2020winogrande}, much less reasoning about social intelligence or mental states and realities of different characters of situations.

% \clearpage
\subsection{Future directions towards \ptlms with \TheoryOfMind}
\label{ssec:future-directions}

% \maarten{Something like: }
While there is no one best path towards \ptlms with social intelligence and \TheoryOfMind, it seems likely that progress will require challenging the standard paradigm of training on \static text with the language modeling objective.
Based on our findings and the limitations we discussed, we reflect on some possible directions forward.
\paragraph{Beyond \static text as training data?}
% 1. Maybe just changing the data will help the models learn ToM and social intelligence? What if data contained more elaborations of social commonsense, how characters feel, etc. Maybe that would teach the model everything it needs? However, a key component for language models seems to be the scale, which can make a big difference; so getting this kind of data at scale would likely take a long time. Possibly, we could use smaller amounts of data that has this information and finetune the PTLMs on that (or use RL); that has been known to work for \cite{Zhou2021PretrainingTT}.

% Due to these limitations, we posit that \ptlms could more efficiently learn \TheoryOfMind abilities through non-\static training data, entity-centric neural architectures and social training objectives.
% Similar to how visual groundings can help with learning physical knowledge \cite{Zhang2022-nk}, grounding training data in its social context
% (e.g., communicative intents, speaker identities) could enable better learning of \TheoryOfMind abilities.
% Echoing arguments by \citet{Bender2020-kt,Bisk2020-xb}, grounding the text used to train \ptlms in its social context (e.g.,  communicative intents, speaker identities) could enable better learning on \TheoryOfMind abilities.
% \maarten{Challenge is getting this context/metadata then? Can we use the example from \cite{Zhou2021PretrainingTT} as an example that it may be fine to finetune PTLMs on better data to get them to exhibit \TheoryOfMind like abilities better?}
Perhaps the key is in the data: the knowledge contained in \static text might be too limited for models to learn social intelligence, for reasons described in \S\ref{ssec:limitations-of-paradigm}
Socially grounded text (containing elaborations of communicative intents, character mental states, speaker identities, etc.) could enable more efficient learning of \TheoryOfMind abilities \cite{Bender2020-kt,Bisk2020-xb,hovy2021importance}, similar to how visual groundings can help with learning physical knowledge \cite{Zhang2022-nk}. 
\newText{Examples of such datasets include ``Social Stories,'' which are devised to help individuals with autism improve their interpersonal skills \cite{gray1995teaching}, or the Story Commonsense \cite{rashkin2018modeling} and GLUCOSE \cite{mostafazadeh2020glucose} commonsense-annotated story datasets.}
% Echoing arguments by \citet{Bender2020-kt,Bisk2020-xb}, this could be accomplished by exploring more interactional training data, such as dialogues in situated contexts that explicitly require reasoning about interlocutor mental states \cite{Bara2021-xd,Zhu2021-zj}. 
Alternatively, perhaps interactional texts, such as dialogues and other datasets that were explicitly created to require reasoning about mental states, could help with neural \TheoryOfMind \cite{Bara2021-xd}.
% \maarten{Need to take this further, be more concrete, and say that learning during interaction is not the same as learning from interactional data? Also address the challenge of how to get the data?}

Nevertheless, the scale of training datasets seems to be crucial for \ptlms \cite{kaplan2020scaling,Chowdhery2022palm}, which poses a challenge: text datasets rich in social intelligence and interactions are not easily found naturally \newText{due to reporting biases,} and they are costly to create \cite{rashkin2018modeling,mostafazadeh2020glucose}.
\newText{Promising results on commonsense reasoning suggest a possible hybrid approach: \ptlms could be jointly or sequentially trained on \static text and commonsense knowledge bases or socially grounded or interactional text \citep{bosselut2019comet,hwang2021comet}, first trained on \static text and then enhanced for commonsense knowledge via reinforcement learning \cite{Zhou2021PretrainingTT}.}
% \maarten{Todo: address reporting bias more directly, hybrid approaches to static text + social commonsense knowledge}
% Promising results on commonsense reasoning \cite{bosselut2019comet,Zhou2021PretrainingTT}
% suggest a possible hybrid approach: learning from large amounts of \static text followed by using finetuning or reinforcement learning on smaller amounts of socially grounded or interactional text.

% \maarten{Need to take this further, be more concrete, and say that learning during interaction is not the same as learning from interactional data? Also address the challenge of how to get the data?}

%%%%%%%%%%%%%%%%%%%%%%%%%%%%%%%%%%%%%%%%%%%%%%%%%
\paragraph{Person-centric neural inductive biases?}
% 2. So maybe we also need to add inductive biases to the model architecture / training objective; possibly, this could be done supervisedly, but again we don't have the data. So it would have to be done unsupervisedly. these could be inspired by neuroscience \cite{Langley2022-fl}, or could be based on investigations of what concepts neural models learn in latent ways through pretraining, and honing in on those in some ways. This would need more research, and could potentially require also returning to recurrent models. Also, this 
While more socially grounded training data could help, 
% Changing the training data might only be one piece of the puzzle; 
\ptlms might also learn social intelligence better if they are designed with person-centric inductive biases and training objectives.
Hinting at this, prior work has shown that training entity-centric neural architectures on text with entity coreference information yields more entity-aware \ptlms, both in recurrent \cite{Henaff2017-ov,Ji2017-iq,Yang2017-uk,Liu2019-kn} and Transformer-based models \cite{Fevry2020-vm,De_Cao2020-ve,Rosset2020-bj,Zhang2022-du}.
% \daniel{if you wanted to include a limitation, you could say that such approaches are often difficult to scale}\maarten{Agreed, but added it to the next sentence/§}

However, \TheoryOfMind and social intelligence require much richer social grounding than coreference chains, which is challenging to obtain for supervised settings, especially at the scale that \ptlms require. 
Thus, unsupervised approaches to adding inductive biases to models could be a promising solution.
% Unfortunately, such approaches would have to be largely unsupervised, due to the lack of large amounts of socially grounded text.
Future work could look to cognitive science and neuroscience research for possible directions \cite{Langley2022-fl}, such as exploring \ptlms' equivalents of human concept cells \cite[i.e., sets of neurons that activate for important people or concepts;][]{Bowers2017-hp,Calvo_Tapia2020-ll}.

Alternatively, examining the internal or latent representations of \ptlms could point to future directions towards inductive biases for neural \TheoryOfMind.
% Alternatively, research that probes the internal representations of \ptlms could point to future directions towards inductive biases for neural \TheoryOfMind.
As an example, recent work has found evidence of latent representations of grounded semantics in models trained only on \static text \cite{Li2021-mp}, which can be tied to real-world grounding with a small amount of additional supervised training \cite{Patel2022-mapping}. 
Future work might similarly analyze deep learning models for representations of \TheoryOfMind, toward augmenting the models with structure or objectives that surface and strengthen these representations.
% Perhaps similar results could emerge from probing for evidence of social and pragmatic representations in existing models.
% We encourage future work to continue to probe for evidence of social and pragmatic representations in existing models, and 
% , and consider additional human-in-the-loop or multi-agent pretraining and finetuning objectives to help surface and strengthen these capabilities, such as those explored by \citet{Hawkins2019-adaptation}, \citet{lazaridou-etal-2020-multi}, \citet{Zhu2021-zj}, and \citet{Wang2022-vi}.

%%%%%%%%%%%%%%%%%%%%%%%%%%%%%%%%%%%%%%%%%%%%%%%%%%
\paragraph{Interactive and experiential grounding?}
It is possible, nevertheless, that socially grounded data and person-centric inductive biases will not suffice. 
% \maarten{Better link to counterfactuals/alternatives, and make more concrete suggestions: predict others' actions, then observe and compare, and learn why predictions were wrong.}
Some researchers have argued that language understanding could only emerge from interactions and experiences \cite{Bender2020-kt,Bisk2020-xb}.
\newText{Likely,}\oldText{Perhaps} this applies to \TheoryOfMind and social intelligence as well\newText{, due to lack of communicative intents and alternatives in \static text}.
\newText{Future work could explore approaches grounded more explicitly in interaction, intents, and alternatives, e.g., by explicitly predicting possible next steps and learning why predictions were wrong.} 
In fact, promising research has shown that using an interactive learning or multi-agent communication paradigm can enable some \TheoryOfMind capabilities of models \cite{Hawkins2019-adaptation,lazaridou-etal-2020-multi,Zhu2021-zj,Wang2022-vi}. 

However, there are limits to the types of \TheoryOfMind that can be learned from interactive simulations, which are often task-specific \cite[e.g., describing objects in an image;][]{lazaridou-etal-2020-multi,steinert2022emergent}.
Furthermore, models that were trained in interactive simulation settings often struggle to generalize beyond the simulation environment \cite{Ludwin-Peery2021-hx,Mu2021-dq}.
Based on promising results by \citet{lazaridou-etal-2020-multi,Zhu2021-zj}, future work \oldText{could further explore hybrid approaches that combine pretraining with interactive learning settings to create generalizable \ptlms with neural \TheoryOfMind.}\newText{might create generalizable \ptlms with neural \TheoryOfMind through hybrid approaches that combine pretraining with interactive learning: updating models trained on static text using supervision either from humans \cite{stiennon2020feedback,ouyang2022gpt3instruct,scheurer2022feedback} or from proxies for human behavior or social environments \cite{ammanabrolu-etal-2022-situated,ammanabrolu-etal-2022-aligning} based on broad coverage LLMs~\cite{perez2022redteaming}.} 
\paragraph{Probing and evaluating \ToM}
% \daniel{possibly reorder the paragraph so that it ends with your call for evaluation/ development on social intelligence benchmarks that's currently 566-569 (so that we're not ending by talking about non-ToM/non-social grounding)}
% \maarten{Todo: polish and add cites.. Daniel can you help?}
While neural \TheoryOfMind and social intelligence may remain an elusive goal for some time, developing measures of those abilities in systems can be done in tandem.
We encourage further research in developing benchmarks that measure specific social abilities in \ptlms \cite[e.g.,][]{sap2019socialIQa,zadeh2019social}, especially those that minimize annotation artifacts and spurious correlations \cite{schwartz2017effect,gururangan-etal-2018-annotation,le-etal-2019-revisiting}.
Additionally, we encourage further investigations into probing the latent knowledge within \ptlms \cite{Tenney2019-fn,Li2021-mp} or examining how \ptlms handle entities and people \cite{Onoe2022-ew,Schuster2022-by}, which could shed light onto better data choices and inductive biases towards neural \TheoryOfMind and social intelligence.

\section{Conclusion}
We explore the open question of whether and how much modern large-scale language models (\ptlms) can reason about social intelligence and \TheoryOfMind.
Our results show that out-of-the-box \ptlms struggle substantially with these abilities\newText{, which we argue are necessary but not sufficient aspects of Theory of Mind}.
% Despite massive successes in other domains, our work highlighted some shortcomings of \ptlms at reasoning about social intelligence and \TheoryOfMind.
Specifically, \gptT's social intelligence as measured by \socialIQa lags behind humans (>30\%), and the model struggles to answer \tomi questions about mental states (55-60\%) compared to factual questions (90--100\%).
In light of these shortcomings, we critically examine the large language model pretraining paradigm from a pragmatics-based perspective, and discuss possible directions towards enabling true social intelligence in NLP systems.

\medskip\noindent
\newText{We make our preprocessed datasets available at \url{http://maartensap.com/neuralToM}.}
% We also saw that participant-centric reasoning is a particular challenge for \gptT.
% On \socialIQa, it often got participants of situations confused, and performed better for questions related to the main character of situations.
% With \tomi, \gptT performed increasingly worse from factual to first-order to second-order mental state questions.

\section{Limitations}

Our work focuses on investigating the \TheoryOfMind abilities in large pretrained language models, but we focus on accessing \gptT \cite{brown2020language} through an API, since we do not have access to some of the larger models out there \cite[PaLM;][]{Chowdhery2022palm} nor do we have the computational resources to run an open-source version of \gptT \cite[OPT;][]{zhang2022opt}. 
We hypothesize that results would not be drastically different with such models, based on the low accuracy displayed on \socialIQa in the recently released BIG-Bench experiments \cite{Srivastava2022BeyondTI}.
Nevertheless, we hope developers of larger \ptlms will investigate these \ToM abilities to confirm or refute our findings.

We measure the ability to answer questions about people's mental states using \tomi, which is an automatically constructed corpus of stories involving people, objects, and locations. 
The automatic nature of the creation process could induce biases and artifacts, such as objects being in locations that are plausible but not typical (e.g., bananas in a closet), which could influence model's ability to answer questions properly. 
Based on the near-perfect accuracy on the factual questions, however, this may not be a significant issue.
Future work should investigate more naturalistic settings to probe this ability in \ptlms.

A potential limitation of our work is that models could latch onto surface patterns and spurious correlations in our two datasets.
For example, theoretically, a model prompted with many \tomi examples may be able to reverse-engineer the data creation algorithm to find the solution to each question.
However, this would be a bigger limitation if our claims were that \ptlms \textit{do} have social intelligence and \TheoryOfMind; instead, given that our results show low performance on these tasks even though they are potentially easier due to correlational patterns, this would indicate that \ptlms have potentially even less reasoning abilities.

Additionally, while we operationalize our measure of social intelligence and \TheoryOfMind through two specific tasks, \socialIQa and \tomi, these abilities are much broader.
\newText{As noted earlier, we view these benchmarks as necessary but not sufficient conditions for \ptlms to have \ToM; solving the benchmarks does not imply that \ptlms have \ToM, but \ptlms with \ToM should be able to solve them.}
We hope that future research will further investigate other aspects of \TheoryOfMind abilities in large pretrained LMs, drawing on social science research.
For example, future work could make use of the ``unexpected content'' task \cite{gopnik1988children} or the ``George Washington University Social Intelligence Test'' \cite{hunt1928measurement} to measure the social intelligence of \ptlms.

\newText{Finally, the focus on English language \ptlms and benchmarks for \TheoryOfMind is another limitation of our work. 
Echoing recent cognitive science work that argues the need for non-English cognitive science investigations \cite{Blasi2022-bi}. 
Specifically, false-belief abilities are greatly influenced by language structure and grammar \cite{Boeg_Thomsen2021-ut,Zhang2022-ez}.
}

% \daniel{consider adding limitations of the directions we propose, or moving them here when they're in the main text (with a note perhaps at beginning of Sec 5 that we outline limitations in this section). e.g. person-centric architectures / inductive biases generally are difficult to scale. Interaction can be brittle (as is currently in the main text)}
% \maarten{I see what you're saying, but I think those limitations are worthy of being discussed in the main text, cause (1) it's relevant to bring them up there so that people don't get annoyed at us throwing out ideas without discussing the challenges and (2) it adds to the narrative tension / flow, by adding the like "well maybe nothing works, that's why it's an open problem"?}

% \maarten{Given the recent debate about AI sentience, should we make a paragraph discussing that we outline these points towards NLP systems with social intelligence and \TheoryOfMind?}
\subsection*{Broader Sociotechnical Implications}
AI systems are part of a broader sociotechnical system that also involves individual motivations and societal norms \cite{Johnson2017-si}. 
As such, per a contextualist view of AI \cite[instead of utopian or dystopian;][]{barbour1992ethics}, we envision AI systems with social intelligence and \TheoryOfMind being used in ways that enhance human's lives, autonomy, and agency \cite{Chan2022-pz}. 
In parallel, we strongly support the development and research of policy and regulation, to prevent misuses of AI with social intelligence \citep{Wischmeyer2020-bz,Crawford2021-kz,Reich2021-xw}.

\section*{Acknowledgements}
\newText{We would like to thank Jack Hessel, Rowan Zellers, Jena D. Hwang, Prithviraj Ammanabrolu for their feedback on preliminary versions of this work, and Anna Jafarpour and Noah Goodman for fruitful cognitive science discussions about the research.
We also thank the anonymous reviewers for their thoughtful comments.
This research was supported by the Allen Institute for AI and the DARPA MCS program through NIWC Pacific (N66001-19-2-4031).
}

\bibliography{0-main-arxiv}

\begin{thebibliography}{135}
\expandafter\ifx\csname natexlab\endcsname\relax\def\natexlab#1{#1}\fi

\bibitem[{Allen(2020)}]{Allen2020-cd}
Summer Allen. 2020.
\newblock \href {http://dx.doi.org/10.1109/MPULS.2020.2993657} {Artificial
  intelligence and the future of psychiatry}.
\newblock \emph{IEEE pulse}, 11(3):2--6.

\bibitem[{Ammanabrolu et~al.(2022{\natexlab{a}})Ammanabrolu, Jia, and
  Riedl}]{ammanabrolu-etal-2022-situated}
Prithviraj Ammanabrolu, Renee Jia, and Mark Riedl. 2022{\natexlab{a}}.
\newblock \href {https://doi.org/10.18653/v1/2022.acl-long.557} {Situated
  dialogue learning through procedural environment generation}.
\newblock In \emph{Proceedings of the 60th Annual Meeting of the Association
  for Computational Linguistics (Volume 1: Long Papers)}, pages 8099--8116,
  Dublin, Ireland. Association for Computational Linguistics.

\bibitem[{Ammanabrolu et~al.(2022{\natexlab{b}})Ammanabrolu, Jiang, Sap,
  Hajishirzi, and Choi}]{ammanabrolu-etal-2022-aligning}
Prithviraj Ammanabrolu, Liwei Jiang, Maarten Sap, Hannaneh Hajishirzi, and
  Yejin Choi. 2022{\natexlab{b}}.
\newblock \href {https://doi.org/10.18653/v1/2022.naacl-main.439} {Aligning to
  social norms and values in interactive narratives}.
\newblock In \emph{Proceedings of the 2022 Conference of the North American
  Chapter of the Association for Computational Linguistics: Human Language
  Technologies}, pages 5994--6017, Seattle, United States. Association for
  Computational Linguistics.

\bibitem[{Apperly(2010)}]{apperly2010mindreaders}
Ian Apperly. 2010.
\newblock \emph{Mindreaders: the cognitive basis of" theory of mind"}.
\newblock Psychology Press.

\bibitem[{Asher and Lascarides(2013)}]{Asher2013-strategic-conversation}
Nicholas Asher and Alex Lascarides. 2013.
\newblock Strategic conversation.
\newblock \emph{Semantics and Pragmatics}, 6.

\bibitem[{Austin(1975)}]{Austin1975-do-things-with-words}
John~Langshaw Austin. 1975.
\newblock \emph{How to Do Things with Words}.
\newblock Clarendon Press.

\bibitem[{Bara et~al.(2021)Bara, CH-Wang, and Chai}]{Bara2021-xd}
Cristian-Paul Bara, Sky CH-Wang, and Joyce Chai. 2021.
\newblock \href {http://arxiv.org/abs/2109.06275} {{MindCraft}: Theory of mind
  modeling for situated dialogue in collaborative tasks}.
\newblock In \emph{{EMNLP}}.

\bibitem[{Barbour(1992)}]{barbour1992ethics}
Ian~G Barbour. 1992.
\newblock Ethics in an age of technology.
\newblock \emph{The Gifford lectures}.

\bibitem[{Baron-Cohen et~al.(1985)Baron-Cohen, Leslie, and
  Frith}]{baron1985does}
Simon Baron-Cohen, Alan~M Leslie, and Uta Frith. 1985.
\newblock Does the autistic child have a “theory of mind”?
\newblock \emph{Cognition}, 21(1):37--46.

\bibitem[{Bender and Koller(2020)}]{Bender2020-kt}
Emily~M Bender and Alexander Koller. 2020.
\newblock \href {https://www.aclweb.org/anthology/2020.acl-main.463.pdf}
  {Climbing towards {NLU}: On meaning, form, and understanding in the age of
  data}.
\newblock In \emph{Proc. of {ACL}}.

\bibitem[{Bickmore and Picard(2005)}]{Bickmore2005-relationships}
Timothy~W Bickmore and Rosalind~W Picard. 2005.
\newblock Establishing and maintaining long-term human-computer relationships.
\newblock \emph{ACM Transactions on Computer-Human Interaction},
  12(2):293--327.

\bibitem[{Bisk et~al.(2020)Bisk, Holtzman, Thomason, Andreas, Bengio, Chai,
  Lapata, Lazaridou, May, Nisnevich, Pinto, and Turian}]{Bisk2020-xb}
Yonatan Bisk, Ari Holtzman, Jesse Thomason, Jacob Andreas, Yoshua Bengio, Joyce
  Chai, Mirella Lapata, Angeliki Lazaridou, Jonathan May, Aleksandr Nisnevich,
  Nicolas Pinto, and Joseph Turian. 2020.
\newblock \href {https://aclanthology.org/2020.emnlp-main.703} {Experience
  grounds language}.
\newblock In \emph{{EMNLP}}, pages 8718--8735, Online. Association for
  Computational Linguistics.

\bibitem[{Blasi et~al.(2022)Blasi, Henrich, Adamou, Kemmerer, and
  Majid}]{Blasi2022-bi}
Dami{\'a}n~E Blasi, Joseph Henrich, Evangelia Adamou, David Kemmerer, and Asifa
  Majid. 2022.
\newblock Over-reliance on english hinders cognitive science.
\newblock \emph{Trends Cogn. Sci.}

\bibitem[{Boeg~Thomsen et~al.(2021)Boeg~Thomsen, Theakston, Kandemirci, and
  Brandt}]{Boeg_Thomsen2021-ut}
Ditte Boeg~Thomsen, Anna Theakston, Birsu Kandemirci, and Silke Brandt. 2021.
\newblock Do complement clauses really support false-belief reasoning? a
  longitudinal study with english-speaking 2- to 3-year-olds.
\newblock \emph{Dev. Psychol.}, 57(8):1210--1227.

\bibitem[{Bosselut et~al.(2019)Bosselut, Rashkin, Sap, Malaviya, Celikyilmaz,
  and Choi}]{bosselut2019comet}
Antoine Bosselut, Hannah Rashkin, Maarten Sap, Chaitanya Malaviya, Asli
  Celikyilmaz, and Yejin Choi. 2019.
\newblock \href {https://www.aclweb.org/anthology/P19-1470} {Comet: Commonsense
  transformers for automatic knowledge graph construction}.
\newblock In \emph{ACL}.

\bibitem[{Bowers(2017)}]{Bowers2017-hp}
Jeffrey~S Bowers. 2017.
\newblock \href {https://doi.org/10.1080/23273798.2016.1267782} {Grandmother
  cells and localist representations: a review of current thinking}.
\newblock \emph{Language, Cognition and Neuroscience}, 32(3):257--273.

\bibitem[{Brown et~al.(2020)Brown, Mann, Ryder, Subbiah, Kaplan, Dhariwal,
  Neelakantan, Shyam, Sastry, Askell, Agarwal, Herbert-Voss, Krueger, Henighan,
  Child, Ramesh, Ziegler, Wu, Winter, Hesse, Chen, Sigler, Litwin, Gray, Chess,
  Clark, Berner, McCandlish, Radford, Sutskever, and
  Amodei}]{brown2020language}
Tom~B Brown, Benjamin Mann, Nick Ryder, Melanie Subbiah, Jared Kaplan, Prafulla
  Dhariwal, Arvind Neelakantan, Pranav Shyam, Girish Sastry, Amanda Askell,
  Sandhini Agarwal, Ariel Herbert-Voss, Gretchen Krueger, Tom Henighan, Rewon
  Child, Aditya Ramesh, Daniel~M Ziegler, Jeffrey Wu, Clemens Winter,
  Christopher Hesse, Mark Chen, Eric Sigler, Mateusz Litwin, Scott Gray,
  Benjamin Chess, Jack Clark, Christopher Berner, Sam McCandlish, Alec Radford,
  Ilya Sutskever, and Dario Amodei. 2020.
\newblock Language models are {Few-Shot} learners.
\newblock In \emph{NeurIPS}.

\bibitem[{Bubeck et~al.(2023)Bubeck, Chandrasekaran, Eldan, Gehrke, Horvitz,
  Kamar, Lee, Lee, Li, Lundberg et~al.}]{bubeck2023sparks}
S{\'e}bastien Bubeck, Varun Chandrasekaran, Ronen Eldan, Johannes Gehrke, Eric
  Horvitz, Ece Kamar, Peter Lee, Yin~Tat Lee, Yuanzhi Li, Scott Lundberg,
  et~al. 2023.
\newblock Sparks of artificial general intelligence: Early experiments with
  gpt-4.
\newblock \emph{arXiv preprint arXiv:2303.12712}.

\bibitem[{Calvo~Tapia et~al.(2020)Calvo~Tapia, Tyukin, and
  Makarov}]{Calvo_Tapia2020-ll}
Carlos Calvo~Tapia, Ivan Tyukin, and Valeri~A Makarov. 2020.
\newblock \href {http://dx.doi.org/10.1038/s41598-020-64466-7} {Universal
  principles justify the existence of concept cells}.
\newblock \emph{Scientific reports}, 10(1):7889.

\bibitem[{Chan(2022)}]{Chan2022-pz}
Anastasia Chan. 2022.
\newblock \href {https://doi.org/10.1007/s43681-022-00148-6} {{GPT-3} and
  {InstructGPT}: technological dystopianism, utopianism, and ``contextual''
  perspectives in {AI} ethics and industry}.
\newblock \emph{AI and Ethics}.

\bibitem[{Chen et~al.(2019)Chen, Lee, Bansal, Cao, Zhang, Lu, Tsay, Wang, Dai,
  Chen, Sohn, and Wu}]{Chen2019smartCompose}
Mia~Xu Chen, Benjamin~N Lee, Gagan Bansal, Yuan Cao, Shuyuan Zhang, Justin Lu,
  Jackie Tsay, Yinan Wang, Andrew~M Dai, Zhifeng Chen, Timothy Sohn, and
  Yonghui Wu. 2019.
\newblock \href {https://doi.org/10.1145/3292500.3330723} {Gmail smart compose:
  {Real-Time} assisted writing}.
\newblock In \emph{Proceedings of the 25th {ACM} {SIGKDD} International
  Conference on Knowledge Discovery \& Data Mining}, KDD '19, pages 2287--2295,
  New York, NY, USA. Association for Computing Machinery.

\bibitem[{Choi(2022)}]{Choi2022-sv}
Yejin Choi. 2022.
\newblock \href {http://dx.doi.org/10.1162/DAED_a_01906} {The curious case of
  commonsense intelligence}.
\newblock \emph{Daedalus}, 151(2):139--155.

\bibitem[{Chowdhery et~al.(2022)Chowdhery, Narang, Devlin, Bosma, Mishra,
  Roberts, Barham, Won, Sutton, Gehrmann, Schuh, Shi, Tsvyashchenko, Maynez,
  Rao, Barnes, Tay, Shazeer, Prabhakaran, Reif, Du, Hutchinson, Pope, Bradbury,
  Austin, Guy, Pengcheng, Duke, Levskaya, Ghemawat, Dev, Michalewski, Garcia,
  Misra, Robinson, Fedus, Zhou, Ippolito, Luan, Lim, Zoph, Spiridonov, Sepassi,
  Dohan, Agrawal, Omernick, Dai, Pillai, Pellat, Lewkowycz, Moreira, Child,
  Polozov, Lee, Zhou, Wang, Saeta, Diaz, Firat, Catasta, Wei, Meier-Hellstern,
  Eck, Dean, Petrov, and Fiedel}]{Chowdhery2022palm}
Aakanksha Chowdhery, Sharan Narang, Jacob Devlin, Maarten Bosma, Gaurav Mishra,
  Adam Roberts, Paul Barham, Hyung Won, Chung~Charles Sutton, Sebastian
  Gehrmann, Parker Schuh, Kensen Shi, Sasha Tsvyashchenko, Joshua Maynez,
  Abhishek Rao, Parker Barnes, Yi~Tay, Noam Shazeer, Vinodkumar Prabhakaran,
  Emily Reif, Nan Du, Ben Hutchinson, Reiner Pope, James Bradbury, Jacob
  Austin, Michael~Isard Guy, Gur-Ari Pengcheng, Yin~Toju Duke, Anselm Levskaya,
  Sanjay Ghemawat, Sunipa Dev, Henryk Michalewski, Xavier Garcia, Vedant Misra,
  Kevin Robinson, Liam Fedus, Denny Zhou, Daphne Ippolito, David Luan,
  Hyeontaek Lim, Barret Zoph, Alexander Spiridonov, Ryan Sepassi, David Dohan,
  Shivani Agrawal, Mark Omernick, Andrew~M Dai, Thanumalayan~Sankaranarayana
  Pillai, Marie Pellat, Aitor Lewkowycz, Erica Moreira, Rewon Child, Oleksandr
  Polozov, Katherine Lee, Zongwei Zhou, Xuezhi Wang, Brennan Saeta, Mark Diaz,
  Orhan Firat, Michele Catasta, Jason Wei, Kathy Meier-Hellstern, Douglas Eck,
  Jeff Dean, Slav Petrov, and Noah Fiedel. 2022.
\newblock \href
  {https://storage.googleapis.com/pathways-language-model/PaLM-paper.pdf}
  {{PaLM}: Scaling language modeling with pathways}.
\newblock Technical report, Google.

\bibitem[{Clark et~al.(2021)Clark, August, Serrano, Haduong, Gururangan, and
  Smith}]{clark2021all}
Elizabeth Clark, Tal August, Sofia Serrano, Nikita Haduong, Suchin Gururangan,
  and Noah~A Smith. 2021.
\newblock All that’s ‘human’is not gold: Evaluating human evaluation of
  generated text.
\newblock In \emph{ACL}.

\bibitem[{Clark(1996)}]{Clark1996-using}
Herbert~H Clark. 1996.
\newblock \emph{Using Language}.
\newblock Cambridge University Press.

\bibitem[{Clark and Brennan(1991)}]{clark1991grounding}
Herbert~H Clark and Susan~E Brennan. 1991.
\newblock Grounding in communication.

\bibitem[{Clark and Krych(2004)}]{Clark2004-monitoring}
Herbert~H Clark and Meredyth~A Krych. 2004.
\newblock Speaking while monitoring addressees for understanding.
\newblock \emph{J. Mem. Lang.}, 50(1):62--81.

\bibitem[{Clark and Schaefer(1989)}]{Clark1989-discourse}
Herbert~H Clark and Edward~F Schaefer. 1989.
\newblock Contributing to discourse.
\newblock \emph{Cognitive Science}, 13(2):259--294.

\bibitem[{Crawford(2021)}]{Crawford2021-kz}
Kate Crawford. 2021.
\newblock \href
  {https://www.degruyter.com/document/doi/10.12987/9780300252392/html}
  {\emph{Atlas of {AI}}}.
\newblock Yale University Press.

\bibitem[{Dale(2021)}]{dale2021gpt}
Robert Dale. 2021.
\newblock Gpt-3: What’s it good for?
\newblock \emph{Natural Language Engineering}, 27(1):113--118.

\bibitem[{De~Cao et~al.(2020)De~Cao, Izacard, Riedel, and
  Petroni}]{De_Cao2020-ve}
Nicola De~Cao, Gautier Izacard, Sebastian Riedel, and Fabio Petroni. 2020.
\newblock \href {http://arxiv.org/abs/2010.00904} {Autoregressive entity
  retrieval}.
\newblock In \emph{{ICLR}}.

\bibitem[{de~Villiers(2007)}]{De_Villiers2007-yj}
Jill de~Villiers. 2007.
\newblock \href {http://dx.doi.org/10.1016/j.lingua.2006.11.006} {The interface
  of language and theory of mind}.
\newblock \emph{Lingua. International review of general linguistics. Revue
  internationale de linguistique generale}, 117(11):1858--1878.

\bibitem[{Dhelim et~al.(2021)Dhelim, Ning, Farha, Chen, Atzori, and
  Daneshmand}]{Dhelim2021-sd}
Sahraoui Dhelim, Huansheng Ning, Fadi Farha, Liming Chen, Luigi Atzori, and
  Mahmoud Daneshmand. 2021.
\newblock \href {http://dx.doi.org/10.1109/JIOT.2021.3081556} {{IoT-Enabled}
  social relationships meet artificial social intelligence}.
\newblock \emph{IEEE Internet of Things Journal}, 8(24):17817--17828.

\bibitem[{Dodge et~al.(2021)Dodge, Sap, Marasović, Agnew, Ilharco, Groeneveld,
  Mitchell, and Gardner}]{dodge2021documentingC4}
Jesse Dodge, Maarten Sap, Ana Marasović, William Agnew, Gabriel Ilharco, Dirk
  Groeneveld, Margaret Mitchell, and Matt Gardner. 2021.
\newblock Documenting large webtext corpora: A case study on the colossal clean
  crawled corpus.
\newblock In \emph{EMNLP}.

\bibitem[{Dunbar(1993)}]{Dunbar1993-ew}
R~I~M Dunbar. 1993.
\newblock \href
  {https://www.cambridge.org/core/services/aop-cambridge-core/content/view/4290FF4D7362511136B9A15A96E74FEF/S0140525X00032325a.pdf/div-class-title-coevolution-of-neocortical-size-group-size-and-language-in-humans-div.pdf}
  {Coevolution of neocortical size, group size and language in humans}.
\newblock \emph{The Behavioral and brain sciences}, 16(4):681--694.

\bibitem[{Elkins and Chun(2020)}]{elkins2020can}
Katherine Elkins and Jon Chun. 2020.
\newblock Can gpt-3 pass a writer’s turing test?
\newblock \emph{Journal of Cultural Analytics}, 1(1):17212.

\bibitem[{Enrici et~al.(2019)Enrici, Bara, and Adenzato}]{Enrici2019-mf}
Ivan Enrici, Bruno~G Bara, and Mauro Adenzato. 2019.
\newblock \href
  {https://www.jbe-platform.com/content/journals/10.1075/pc.19010.enr} {Theory
  of mind, pragmatics and the brain: Converging evidence for the role of
  intention processing as a core feature ofhuman communication}.
\newblock \emph{Pragmatics \& Cognition}, 26(1):5--38.

\bibitem[{Fern{\'a}ndez(2013)}]{Fernandez2013-su}
Camila Fern{\'a}ndez. 2013.
\newblock \href {https://doi.org/10.1177/0142723711422633} {Mindful
  storytellers: Emerging pragmatics and theory of mind development}.
\newblock \emph{First language}, 33(1):20--46.

\bibitem[{F{\'e}vry et~al.(2020)F{\'e}vry, Soares, FitzGerald, Choi, and
  Kwiatkowski}]{Fevry2020-vm}
Thibault F{\'e}vry, Livio~Baldini Soares, Nicholas FitzGerald, Eunsol Choi, and
  Tom Kwiatkowski. 2020.
\newblock \href {http://arxiv.org/abs/2004.07202} {Entities as experts: Sparse
  memory access with entity supervision}.
\newblock In \emph{{EMNLP}}.

\bibitem[{Ganaie and Mudasir(2015)}]{ganaie2015study}
MY~Ganaie and Hafiz Mudasir. 2015.
\newblock {A Study of Social Intelligence \& Academic Achievement of College
  Students of District Srinagar, J\&K, India}.
\newblock \emph{Journal of American Science}, 11(3):23--27.

\bibitem[{Gao et~al.(2020)Gao, Biderman, Black, Golding, Hoppe, Foster, Phang,
  He, Thite, Nabeshima et~al.}]{gao2020pile}
Leo Gao, Stella Biderman, Sid Black, Laurence Golding, Travis Hoppe, Charles
  Foster, Jason Phang, Horace He, Anish Thite, Noa Nabeshima, et~al. 2020.
\newblock The pile: An 800gb dataset of diverse text for language modeling.
\newblock \emph{arXiv preprint arXiv:2101.00027}.

\bibitem[{Gardner et~al.(1995)Gardner, Hanson, and Hamilton}]{gardner1995kids}
Howard Gardner, Robert~M Hanson, and Steve Hamilton. 1995.
\newblock \emph{How Are Kids SMART?: Multiple Intelligences in the Classroom}.
\newblock National Professional Resources, Incorporated.

\bibitem[{Goodman and Frank(2016)}]{Goodman2016-rsa}
Noah~D Goodman and Michael~C Frank. 2016.
\newblock Pragmatic language interpretation as probabilistic inference.
\newblock \emph{Trends in Cognitive Science}, 20(11):818--829.

\bibitem[{Gopnik and Astington(1988)}]{gopnik1988children}
Alison Gopnik and Janet~W Astington. 1988.
\newblock Children's understanding of representational change and its relation
  to the understanding of false belief and the appearance-reality distinction.
\newblock \emph{Child development}, pages 26--37.

\bibitem[{Gordon and Van~Durme(2013)}]{Gordon2013reportingbias}
Jonathan Gordon and Benjamin Van~Durme. 2013.
\newblock \href {https://doi.org/10.1145/2509558.2509563} {Reporting bias and
  knowledge acquisition}.
\newblock In \emph{Proceedings of the 2013 Workshop on Automated Knowledge Base
  Construction}, AKBC '13, pages 25--30, New York, NY, USA. ACM.

\bibitem[{Graesser et~al.(1994)Graesser, Singer, and
  Trabasso}]{Graesser1994-cc}
A~C Graesser, M~Singer, and T~Trabasso. 1994.
\newblock \href {https://www.ncbi.nlm.nih.gov/pubmed/7938337} {Constructing
  inferences during narrative text comprehension}.
\newblock \emph{Psychological review}, 101(3):371--395.

\bibitem[{Grant et~al.(2017)Grant, Nematzadeh, and Griffiths}]{grant2017can}
Erin Grant, Aida Nematzadeh, and Thomas~L Griffiths. 2017.
\newblock How can memory-augmented neural networks pass a false-belief task?
\newblock In \emph{CogSci}.

\bibitem[{Gray(1995)}]{gray1995teaching}
Carol~A Gray. 1995.
\newblock Teaching children with autism to “read” social situations.
\newblock \emph{Teaching children with autism: Strategies to enhance
  communication and socialization}, pages 219--242.

\bibitem[{Grice(1975)}]{grice1975logic}
Herbert~P Grice. 1975.
\newblock Logic and conversation.
\newblock In \emph{Speech acts}, pages 41--58. Brill.

\bibitem[{Grosz et~al.(1995)Grosz, Joshi, and Weinstein}]{Grosz1995-hm}
Barbara~J Grosz, Aravind~K Joshi, and Scott Weinstein. 1995.
\newblock \href {https://aclanthology.org/J95-2003} {{C}entering: A framework
  for modeling the local coherence of discourse}.
\newblock \emph{Computational Linguistics}, 21(2):203--225.

\bibitem[{Gunning(2018)}]{Gunning2018-xn}
David Gunning. 2018.
\newblock \href {http://arxiv.org/abs/1810.07528} {Machine common sense concept
  paper}.

\bibitem[{Gururangan et~al.(2018)Gururangan, Swayamdipta, Levy, Schwartz,
  Bowman, and Smith}]{gururangan-etal-2018-annotation}
Suchin Gururangan, Swabha Swayamdipta, Omer Levy, Roy Schwartz, Samuel Bowman,
  and Noah~A. Smith. 2018.
\newblock \href {https://doi.org/10.18653/v1/N18-2017} {Annotation artifacts in
  natural language inference data}.
\newblock In \emph{Proceedings of the 2018 Conference of the North {A}merican
  Chapter of the Association for Computational Linguistics: Human Language
  Technologies, Volume 2 (Short Papers)}, pages 107--112, New Orleans,
  Louisiana. Association for Computational Linguistics.

\bibitem[{Hawkins et~al.(2019)Hawkins, Kwon, Sadigh, and
  Goodman}]{Hawkins2019-adaptation}
Robert~D Hawkins, Minae Kwon, Dorsa Sadigh, and Noah~D Goodman. 2019.
\newblock Continual adaptation for efficient machine communication.
\newblock In \emph{{CoNLL}}.

\bibitem[{Henaff et~al.(2017)Henaff, Weston, Szlam, Bordes, and
  LeCun}]{Henaff2017-ov}
Mikael Henaff, Jason Weston, Arthur Szlam, Antoine Bordes, and Yann LeCun.
  2017.
\newblock \href {http://openreview.net/pdf?id=rJTKKKqeg} {Tracking the world
  state with recurrent entity networks}.
\newblock In \emph{{ICLR}}.

\bibitem[{Hovy and Yang(2021)}]{hovy2021importance}
Dirk Hovy and Diyi Yang. 2021.
\newblock The importance of modeling social factors of language: Theory and
  practice.
\newblock In \emph{NAACL}. Association for Computational Linguistics.

\bibitem[{Hunt(1928)}]{hunt1928measurement}
Thelma Hunt. 1928.
\newblock The measurement of social intelligence.
\newblock \emph{Journal of Applied Psychology}, 12(3):317.

\bibitem[{Hwang et~al.(2021)Hwang, Bhagavatula, Le~Bras, Da, Sakaguchi,
  Bosselut, and Choi}]{hwang2021comet}
Jena~D Hwang, Chandra Bhagavatula, Ronan Le~Bras, Jeff Da, Keisuke Sakaguchi,
  Antoine Bosselut, and Yejin Choi. 2021.
\newblock {(COMET-)ATOMIC2020}: On symbolic and neural commonsense knowledge
  graphs.
\newblock In \emph{AAAI}.

\bibitem[{Jaques et~al.(2020)Jaques, Shen, Ghandeharioun, Ferguson, Lapedriza,
  Jones, Gu, and Picard}]{jaques-etal-2020-human}
Natasha Jaques, Judy~Hanwen Shen, Asma Ghandeharioun, Craig Ferguson, Agata
  Lapedriza, Noah Jones, Shixiang Gu, and Rosalind Picard. 2020.
\newblock \href {https://doi.org/10.18653/v1/2020.emnlp-main.327}
  {Human-centric dialog training via offline reinforcement learning}.
\newblock In \emph{Proceedings of the 2020 Conference on Empirical Methods in
  Natural Language Processing (EMNLP)}, pages 3985--4003, Online. Association
  for Computational Linguistics.

\bibitem[{Jaques(2019)}]{Jaques2019-gp}
Natasha~M Jaques. 2019.
\newblock \href {https://dspace.mit.edu/handle/1721.1/129901?show=full}
  {\emph{Social and affective machine learning}}.
\newblock Ph.D. thesis, Massachusetts Institute of Technology.

\bibitem[{Ji et~al.(2017)Ji, Tan, Martschat, Choi, and Smith}]{Ji2017-iq}
Yangfeng Ji, Chenhao Tan, Sebastian Martschat, Yejin Choi, and Noah~A Smith.
  2017.
\newblock \href {http://aclweb.org/anthology/D17-1195} {Dynamic entity
  representations in neural language models}.
\newblock In \emph{{EMNLP}}, pages 1831--1840, Copenhagen, Denmark. Association
  for Computational Linguistics.

\bibitem[{Johnson and Verdicchio(2017)}]{Johnson2017-si}
Deborah~G Johnson and Mario Verdicchio. 2017.
\newblock \href {https://doi.org/10.1007/s11023-017-9417-6} {Reframing {AI}
  discourse}.
\newblock \emph{Minds and Machines}, 27(4):575--590.

\bibitem[{Kaplan et~al.(2020)Kaplan, McCandlish, Henighan, Brown, Chess, Child,
  Gray, Radford, Wu, and Amodei}]{kaplan2020scaling}
Jared Kaplan, Sam McCandlish, Tom Henighan, Tom~B Brown, Benjamin Chess, Rewon
  Child, Scott Gray, Alec Radford, Jeffrey Wu, and Dario Amodei. 2020.
\newblock Scaling laws for neural language models.
\newblock \emph{arXiv preprint arXiv:2001.08361}.

\bibitem[{Kearns et~al.(2020)Kearns, Kaura, Divina, Vo, Si, Ward, and
  Yuwen}]{Kearns2020-ty}
William~R Kearns, Neha Kaura, Myra Divina, Cuong Vo, Dong Si, Teresa Ward, and
  Weichao Yuwen. 2020.
\newblock \href {https://doi.org/10.1145/3334480.3382902} {A {Wizard-of-Oz}
  interface and persona-based methodology for collecting health counseling
  dialog}.
\newblock In \emph{{CHI} Extended Abstracts}, CHI EA '20, pages 1--9, New York,
  NY, USA. Association for Computing Machinery.

\bibitem[{Korkmaz(2011)}]{korkmaz2011theory}
Baris Korkmaz. 2011.
\newblock Theory of mind and neurodevelopmental disorders of childhood.
\newblock \emph{Pediatr Res}, 69(5 Pt 2):101R--8R.

\bibitem[{Kosinski(2023)}]{kosinski2023theory}
Michal Kosinski. 2023.
\newblock Theory of mind may have spontaneously emerged in large language
  models.
\newblock \emph{arXiv preprint arXiv:2302.02083}.

\bibitem[{Krauss and Weinheimer(1966)}]{Krauss1966-feedback}
R~M Krauss and S~Weinheimer. 1966.
\newblock Concurrent feedback, confirmation, and the encoding of referents in
  verbal communication.
\newblock \emph{J. Pers. Soc. Psychol.}, 4(3):343--346.

\bibitem[{Langley et~al.(2022)Langley, Cirstea, Cuzzolin, and
  Sahakian}]{Langley2022-fl}
Christelle Langley, Bogdan~Ionut Cirstea, Fabio Cuzzolin, and Barbara~J
  Sahakian. 2022.
\newblock \href {https://www.frontiersin.org/article/10.3389/frai.2022.778852}
  {Theory of mind and preference learning at the interface of cognitive
  science, neuroscience, and {AI}: A review}.
\newblock \emph{Frontiers in Artificial Intelligence and Applications}, 5.

\bibitem[{Lazaridou et~al.(2020)Lazaridou, Potapenko, and
  Tieleman}]{lazaridou-etal-2020-multi}
Angeliki Lazaridou, Anna Potapenko, and Olivier Tieleman. 2020.
\newblock \href {https://doi.org/10.18653/v1/2020.acl-main.685} {Multi-agent
  communication meets natural language: Synergies between functional and
  structural language learning}.
\newblock In \emph{Proceedings of the 58th Annual Meeting of the Association
  for Computational Linguistics}, pages 7663--7674, Online. Association for
  Computational Linguistics.

\bibitem[{Le et~al.(2019)Le, Boureau, and Nickel}]{le-etal-2019-revisiting}
Matthew Le, Y-Lan Boureau, and Maximilian Nickel. 2019.
\newblock \href {https://doi.org/10.18653/v1/D19-1598} {Revisiting the
  evaluation of theory of mind through question answering}.
\newblock In \emph{Proceedings of the 2019 Conference on Empirical Methods in
  Natural Language Processing and the 9th International Joint Conference on
  Natural Language Processing (EMNLP-IJCNLP)}, pages 5872--5877, Hong Kong,
  China. Association for Computational Linguistics.

\bibitem[{Lenci(2023)}]{Lenci2023-yg}
Alessandro Lenci. 2023.
\newblock \href {http://arxiv.org/abs/2303.04229} {Understanding natural
  language understanding systems. a critical analysis}.

\bibitem[{Li et~al.(2021)Li, Nye, and Andreas}]{Li2021-mp}
Belinda~Z Li, Maxwell Nye, and Jacob Andreas. 2021.
\newblock \href {https://aclanthology.org/2021.acl-long.143/} {Implicit
  representations of meaning in neural language models}.
\newblock In \emph{{ACL}}, pages 1813--1827.

\bibitem[{Litman and Forbes-Riley(2004)}]{litman-forbes-riley-2004-predicting}
Diane~J. Litman and Kate Forbes-Riley. 2004.
\newblock \href {https://doi.org/10.3115/1218955.1219000} {Predicting student
  emotions in computer-human tutoring dialogues}.
\newblock In \emph{Proceedings of the 42nd Annual Meeting of the Association
  for Computational Linguistics ({ACL}-04)}, pages 351--358, Barcelona, Spain.

\bibitem[{Liu et~al.(2019)Liu, Zettlemoyer, and Eisenstein}]{Liu2019-kn}
Fei Liu, Luke Zettlemoyer, and Jacob Eisenstein. 2019.
\newblock \href {https://www.aclweb.org/anthology/P19-1593} {The referential
  reader: A recurrent entity network for anaphora resolution}.
\newblock In \emph{{ACL}}, pages 5918--5925, Stroudsburg, PA, USA. Association
  for Computational Linguistics.

\bibitem[{Lourie et~al.(2021)Lourie, Le~Bras, Bhagavatula, and
  Choi}]{lourie2021unicorn}
Nicholas Lourie, Ronan Le~Bras, Chandra Bhagavatula, and Yejin Choi. 2021.
\newblock Unicorn on rainbow: A universal commonsense reasoning model on a new
  multitask benchmark.
\newblock In \emph{AAAI}.

\bibitem[{Lucy and Gauthier(2017)}]{Lucy2017AreDR}
Li~Lucy and Jon Gauthier. 2017.
\newblock Are distributional representations ready for the real world?
  evaluating word vectors for grounded perceptual meaning.
\newblock In \emph{RoboNLP@ACL}.

\bibitem[{Ludwin-Peery et~al.(2021)Ludwin-Peery, Bramley, Davis, and
  Gureckis}]{Ludwin-Peery2021-hx}
Ethan Ludwin-Peery, Neil~R Bramley, Ernest Davis, and Todd~M Gureckis. 2021.
\newblock \href {http://dx.doi.org/10.1016/j.cogpsych.2021.101396} {Limits on
  simulation approaches in intuitive physics}.
\newblock \emph{Cognitive psychology}, 127:101396.

\bibitem[{Mairesse et~al.(2007)Mairesse, Walker, Mehl, and
  Moore}]{Mairesse2007-personality}
F~Mairesse, M~A Walker, M~R Mehl, and R~K Moore. 2007.
\newblock Using linguistic cues for the automatic recognition of personality in
  conversation and text.
\newblock \emph{Journal of Artificial Intelligence Research}, 30:457--500.

\bibitem[{Marcus(2022)}]{Marcus2022-fg}
Gary Marcus. 2022.
\newblock \href {https://nautil.us/deep-learning-is-hitting-a-wall-14467/}
  {Deep learning is hitting a wall}.
\newblock \emph{Nautilus}.

\bibitem[{Marcus and Davis(2023)}]{Marcus2023-sq}
Gary Marcus and Ernest Davis. 2023.
\newblock How not to test {GPT-3}.

\bibitem[{McDonald and Pearson(2019)}]{McDonald2019-fy}
Kelsey~R McDonald and John~M Pearson. 2019.
\newblock \href
  {https://www.sciencedirect.com/science/article/pii/S2352154618301979}
  {Cognitive bots and algorithmic humans: toward a shared understanding of
  social intelligence}.
\newblock \emph{Current Opinion in Behavioral Sciences}, 29:55--62.

\bibitem[{Miller(2006)}]{Miller2006-pg}
Carol~A Miller. 2006.
\newblock \href {http://dx.doi.org/10.1044/1058-0360(2006/014)} {Developmental
  relationships between language and theory of mind}.
\newblock \emph{American journal of speech-language pathology / American
  Speech-Language-Hearing Association}, 15(2):142--154.

\bibitem[{Mostafazadeh et~al.(2020)Mostafazadeh, Kalyanpur, Moon, Buchanan,
  Berkowitz, Biran, and Chu-Carroll}]{mostafazadeh2020glucose}
Nasrin Mostafazadeh, Aditya Kalyanpur, Lori Moon, David Buchanan, Lauren
  Berkowitz, Or~Biran, and Jennifer Chu-Carroll. 2020.
\newblock Glucose: Generalized and contextualized story explanations.
\newblock In \emph{Proceedings of the 2020 Conference on Empirical Methods in
  Natural Language Processing (EMNLP)}, pages 4569--4586.

\bibitem[{Mu and Goodman(2021)}]{Mu2021-dq}
Jesse Mu and Noah Goodman. 2021.
\newblock \href {http://arxiv.org/abs/2106.02668} {Emergent communication of
  generalizations}.
\newblock In \emph{{NeurIPS}}.

\bibitem[{Narang and Chowdhery(2022)}]{Narang2022palmsBlog}
Sharan Narang and Aakanksha Chowdhery. 2022.
\newblock \href
  {https://ai.googleblog.com/2022/04/pathways-language-model-palm-scaling-to.html}
  {Pathways language model ({PaLM)}: Scaling to 540 billion parameters for
  breakthrough performance}.
\newblock
  \url{https://ai.googleblog.com/2022/04/pathways-language-model-palm-scaling-to.html}.
\newblock Accessed: 2022-5-5.

\bibitem[{Nematzadeh et~al.(2018)Nematzadeh, Burns, Grant, Gopnik, and
  Griffiths}]{nematzadeh2018evaluating}
Aida Nematzadeh, Kaylee Burns, Erin Grant, Alison Gopnik, and Thomas~L
  Griffiths. 2018.
\newblock Evaluating theory of mind in question answering.
\newblock \emph{arXiv preprint arXiv:1808.09352}.

\bibitem[{O{'}Connor and Andreas(2021)}]{oconnor-andreas-2021-context}
Joe O{'}Connor and Jacob Andreas. 2021.
\newblock \href {https://doi.org/10.18653/v1/2021.acl-long.70} {What context
  features can transformer language models use?}
\newblock In \emph{Proceedings of the 59th Annual Meeting of the Association
  for Computational Linguistics and the 11th International Joint Conference on
  Natural Language Processing (Volume 1: Long Papers)}, pages 851--864, Online.
  Association for Computational Linguistics.

\bibitem[{Onishi and Baillargeon(2005)}]{Onishi2005-pm}
Kristine~H Onishi and Ren{\'e}e Baillargeon. 2005.
\newblock \href {http://dx.doi.org/10.1126/science.1107621} {Do 15-month-old
  infants understand false beliefs?}
\newblock \emph{Science}, 308(5719):255--258.

\bibitem[{Onoe et~al.(2022)Onoe, Zhang, Choi, and Durrett}]{Onoe2022-ew}
Yasumasa Onoe, Michael J~Q Zhang, Eunsol Choi, and Greg Durrett. 2022.
\newblock \href {http://arxiv.org/abs/2205.02832} {Entity cloze by date: What
  {LMs} know about unseen entities}.
\newblock In \emph{Findings of {NAACL}}.

\bibitem[{{OpenAI}(2023)}]{OpenAI2023gpt4}
{OpenAI}. 2023.
\newblock {GPT-4} technical report.
\newblock Technical report.

\bibitem[{Ouyang et~al.(2022)Ouyang, Wu, Jiang, Almeida, Wainwright, Mishkin,
  Zhang, Agarwal, Slama, Ray et~al.}]{ouyang2022gpt3instruct}
Long Ouyang, Jeff Wu, Xu~Jiang, Diogo Almeida, Carroll~L Wainwright, Pamela
  Mishkin, Chong Zhang, Sandhini Agarwal, Katarina Slama, Alex Ray, et~al.
  2022.
\newblock Training language models to follow instructions with human feedback.
\newblock \emph{arXiv preprint arXiv:2203.02155}.

\bibitem[{Patel and Pavlick(2022)}]{Patel2022-mapping}
Roma Patel and Ellie Pavlick. 2022.
\newblock Mapping language models to grounded conceptual spaces.
\newblock In \emph{International Conference on Learning Representations}.

\bibitem[{Pereira et~al.(2016)Pereira, Prada, and
  Santos}]{Pereira2016socialpower}
Gon{\c c}alo Pereira, Rui Prada, and Pedro~A Santos. 2016.
\newblock Integrating social power into the decision-making of cognitive
  agents.
\newblock \emph{Artificial Intelligence}, 241:1--44.

\bibitem[{Perez et~al.(2022)Perez, Huang, Song, Cai, Ring, Aslanides, Glaese,
  McAleese, and Irving}]{perez2022redteaming}
Ethan Perez, Saffron Huang, Francis Song, Trevor Cai, Roman Ring, John
  Aslanides, Amelia Glaese, Nat McAleese, and Geoffrey Irving. 2022.
\newblock Red teaming language models with language models.

\bibitem[{Pollack(2005)}]{pollack2005intelligent}
Martha~E Pollack. 2005.
\newblock Intelligent technology for an aging population: The use of ai to
  assist elders with cognitive impairment.
\newblock \emph{AI magazine}, 26(2):9--9.

\bibitem[{Poulin-Dubois and Yott(2018)}]{Poulin-Dubois2018-td}
Diane Poulin-Dubois and Jessica Yott. 2018.
\newblock \href {http://dx.doi.org/10.1111/desc.12600} {Probing the depth of
  infants' theory of mind: disunity in performance across paradigms}.
\newblock \emph{Developmental science}, 21(4):e12600.

\bibitem[{Premack and Woodruff(1978)}]{Premack1978-sh}
David Premack and Guy Woodruff. 1978.
\newblock \href
  {https://www.cambridge.org/core/journals/behavioral-and-brain-sciences/article/does-the-chimpanzeehave-a-theory-of-mind/1E96B02CD9850016B7C93BC6D2FEF1D0}
  {Does the chimpanzee have a theory of mind?}
\newblock \emph{The Behavioral and brain sciences}, 1(4):515--526.

\bibitem[{Pyers and Senghas(2009)}]{Pyers2009-up}
Jennie~E Pyers and Ann Senghas. 2009.
\newblock Language promotes false-belief understanding: evidence from learners
  of a new sign language.
\newblock \emph{Psychol. Sci.}, 20(7):805--812.

\bibitem[{Rashkin et~al.(2018)Rashkin, Bosselut, Sap, Knight, and
  Choi}]{rashkin2018modeling}
Hannah Rashkin, Antoine Bosselut, Maarten Sap, Kevin Knight, and Yejin Choi.
  2018.
\newblock \href {https://www.aclweb.org/anthology/P18-1213} {Modeling naive
  psychology of characters in simple commonsense stories}.
\newblock In \emph{ACL}.

\bibitem[{Reich et~al.(2021)Reich, Sahami, and Weinstein}]{Reich2021-xw}
Rob Reich, Mehran Sahami, and Jeremy~M Weinstein. 2021.
\newblock \emph{System error: Where big tech went wrong and how we can reboot}.
\newblock Hodder \& Stoughton.

\bibitem[{Ros{\'e} et~al.(2014)Ros{\'e}, Carlson, Yang, Wen, Resnick, Goldman,
  and Sherer}]{Rose2014-mooc-attrition}
Carolyn~Penstein Ros{\'e}, Ryan Carlson, Diyi Yang, Miaomiao Wen, Lauren
  Resnick, Pam Goldman, and Jennifer Sherer. 2014.
\newblock Social factors that contribute to attrition in {MOOCs}.
\newblock In \emph{Proceedings of the first {ACM} conference on Learning @
  Scale}, L@S '14, pages 197--198, New York, NY, USA. Association for Computing
  Machinery.

\bibitem[{Rosset et~al.(2020)Rosset, Xiong, Phan, Song, Bennett, and
  Tiwary}]{Rosset2020-bj}
Corby Rosset, Chenyan Xiong, Minh Phan, Xia Song, Paul Bennett, and Saurabh
  Tiwary. 2020.
\newblock \href {http://arxiv.org/abs/2007.00655} {{Knowledge-Aware} language
  model pretraining}.
\newblock In \emph{{NeurIPS}}.

\bibitem[{Rubio-Fernandez(2021)}]{Rubio-Fernandez2021-gm}
Paula Rubio-Fernandez. 2021.
\newblock Pragmatic markers: the missing link between language and theory of
  mind.
\newblock \emph{Synthese}, 199(1):1125--1158.

\bibitem[{Sakaguchi et~al.(2020)Sakaguchi, Le~Bras, Bhagavatula, and
  Choi}]{sakaguchi2020winogrande}
Keisuke Sakaguchi, Ronan Le~Bras, Chandra Bhagavatula, and Yejin Choi. 2020.
\newblock Winogrande: An adversarial winograd schema challenge at scale.
\newblock In \emph{Proceedings of the AAAI Conference on Artificial
  Intelligence}.

\bibitem[{Sap et~al.(2019{\natexlab{a}})Sap, Le~Bras, Allaway, Bhagavatula,
  Lourie, Rashkin, Roof, Smith, and Choi}]{sap2019atomic}
Maarten Sap, Ronan Le~Bras, Emily Allaway, Chandra Bhagavatula, Nicholas
  Lourie, Hannah Rashkin, Brendan Roof, Noah~A Smith, and Yejin Choi.
  2019{\natexlab{a}}.
\newblock Atomic: An atlas of machine commonsense for if-then reasoning.
\newblock In \emph{Proceedings of the AAAI Conference on Artificial
  Intelligence}.

\bibitem[{Sap et~al.(2019{\natexlab{b}})Sap, Rashkin, Chen, LeBras, and
  Choi}]{sap2019socialIQa}
Maarten Sap, Hannah Rashkin, Derek Chen, Ronan LeBras, and Yejin Choi.
  2019{\natexlab{b}}.
\newblock \href {https://www.aclweb.org/anthology/D19-1454} {Social iqa:
  Commonsense reasoning about social interactions}.
\newblock In \emph{EMNLP}.
\newblock Data downloaded from
  \url{http://maartensap.com/social-iqa/data/socialIQa_v1.4_withDims.tgz}.

\bibitem[{Scheurer et~al.(2022)Scheurer, Campos, Chan, Chen, Cho, and
  Perez}]{scheurer2022feedback}
J{\'e}r{\'e}my Scheurer, Jon~Ander Campos, Jun~Shern Chan, Angelica Chen,
  Kyunghyun Cho, and Ethan Perez. 2022.
\newblock Training language models with language feedback.
\newblock In \emph{{ACL} Workshop on Learning with Natural Language
  Supervision. 2022.}

\bibitem[{Schuster and Linzen(2022)}]{Schuster2022-by}
Sebastian Schuster and Tal Linzen. 2022.
\newblock \href {http://arxiv.org/abs/2205.03472} {When a sentence does not
  introduce a discourse entity, transformer-based models still sometimes refer
  to it}.
\newblock In \emph{{NAACL}}.

\bibitem[{Schwartz et~al.(2017)Schwartz, Sap, Konstas, Zilles, Choi, and
  Smith}]{schwartz2017effect}
Roy Schwartz, Maarten Sap, Ioannis Konstas, Li~Zilles, Yejin Choi, and Noah~A
  Smith. 2017.
\newblock \href {https://www.aclweb.org/anthology/K17-1004} {The effect of
  different writing tasks on linguistic style: A case study of the roc story
  cloze task}.
\newblock In \emph{CoNLL}.

\bibitem[{Sharma et~al.(2021)Sharma, Lin, Miner, Atkins, and
  Althoff}]{sharma2021towards}
Ashish Sharma, Inna~W Lin, Adam~S Miner, David~C Atkins, and Tim Althoff. 2021.
\newblock Towards facilitating empathic conversations in online mental health
  support: A reinforcement learning approach.
\newblock In \emph{Proceedings of the Web Conference 2021}, pages 194--205.

\bibitem[{Southgate and Vernetti(2014)}]{Southgate2014-gi}
Victoria Southgate and Angelina Vernetti. 2014.
\newblock \href {http://dx.doi.org/10.1016/j.cognition.2013.08.008}
  {Belief-based action prediction in preverbal infants}.
\newblock \emph{Cognition}, 130(1):1--10.

\bibitem[{Sperber and Wilson(1986)}]{Sperber1986-ha}
Dan Sperber and Deirdre Wilson. 1986.
\newblock \href
  {https://citeseerx.ist.psu.edu/viewdoc/download?doi=10.1.1.233.793&rep=rep1&type=pdf}
  {\emph{Relevance: Communication and cognition}}, volume 142.
\newblock Citeseer.

\bibitem[{Srivastava et~al.(2022)Srivastava, Rastogi, Rao, Shoeb, Abid, Fisch,
  Brown, Santoro, Gupta, Garriga-Alonso, Kluska, Lewkowycz, Agarwal, Power,
  Ray, Warstadt, Kocurek, Safaya, Tazarv, Xiang, Parrish, Nie, Hussain, Askell,
  Dsouza, Rahane, Iyer, Andreassen, Santilli, Stuhlmuller, Dai, La, Lampinen,
  Zou, Jiang, Chen, Vuong, Gupta, Gottardi, Norelli, Venkatesh, Gholamidavoodi,
  Tabassum, Menezes, Kirubarajan, Mullokandov, Sabharwal, Herrick, Efrat,
  Erdem, Karakacs, Roberts, Loe, Zoph, Bojanowski, Ozyurt, Hedayatnia,
  Neyshabur, Inden, Stein, Ekmekci, Lin, Howald, Diao, Dour, Stinson, Argueta,
  Ram'irez, Singh, Rathkopf, Meng, Baral, Wu, Callison-Burch, Waites, Voigt,
  Manning, Potts, Ramirez, Rivera, Siro, Raffel, Ashcraft, Garbacea, Sileo,
  Garrette, Hendrycks, Kilman, Roth, Freeman, Khashabi, Levy, Gonz'alez,
  Hernandez, Chen, Ippolito, Gilboa, Dohan, Drakard, Jurgens, Datta, Ganguli,
  Emelin, Kleyko, Yuret, Chen, Tam, Hupkes, Misra, Buzan, Mollo, Yang, Lee,
  Shutova, Cubuk, Segal, Hagerman, Barnes, Donoway, Pavlick, Rodol{\`a}, Lam,
  Chu, Tang, Erdem, Chang, Chi, Dyer, Jerzak, Kim, Manyasi, Zheltonozhskii,
  Xia, Siar, Mart'inez-Plumed, Happ'e, Chollet, Rong, Mishra, Winata, de~Melo,
  Kruszewski, Parascandolo, Mariani, Wang, Jaimovitch-L'opez, Betz, Gur-Ari,
  Galijasevic, Kim, Rashkin, Hajishirzi, Mehta, Bogar, Shevlin, Sch{\"u}tze,
  Yakura, Zhang, Wong, Ng, Noble, Jumelet, Geissinger, Kernion, Hilton, Lee,
  Fisac, Simon, Koppel, Zheng, Zou, Koco'n, Thompson, Kaplan, Radom,
  Sohl-Dickstein, Phang, Wei, Yosinski, Novikova, Bosscher, Marsh, Kim, Taal,
  Engel, Alabi, Xu, Song, Tang, Waweru, Burden, Miller, Balis, Berant,
  Frohberg, Rozen, Hern{\'a}ndez-Orallo, Boudeman, Jones, Tenenbaum, Rule,
  Chua, Kanclerz, Livescu, Krauth, Gopalakrishnan, Ignatyeva, Markert, Dhole,
  Gimpel, Omondi, Mathewson, Chiafullo, Shkaruta, Shridhar, McDonell,
  Richardson, Reynolds, Gao, Zhang, Dugan, Qin, Contreras-Ochando, Morency,
  Moschella, Lam, Noble, Schmidt, He, Col'on, Metz, cSenel, Bosma, Sap, ter
  Hoeve, Andrea, Farooqi, Faruqui, Mazeika, Baturan, Marelli, Maru, Quintana,
  Tolkiehn, Giulianelli, Lewis, Potthast, Leavitt, Hagen, Schubert,
  Baitemirova, Arnaud, McElrath, Yee, Cohen, Gu, Ivanitskiy, Starritt, Strube,
  Swkedrowski, Bevilacqua, Yasunaga, Kale, Cain, Xu, Suzgun, Tiwari, Bansal,
  Aminnaseri, Geva, Gheini, MukundVarma, Peng, Chi, Lee, Krakover, Cameron,
  Roberts, Doiron, Nangia, Deckers, Muennighoff, Keskar, Iyer, Constant,
  Fiedel, Wen, Zhang, Agha, Elbaghdadi, Levy, Evans, Casares, Doshi, Fung,
  Liang, Vicol, Alipoormolabashi, Liao, Liang, Chang, Eckersley, Htut, Hwang,
  Milkowski, Patil, Pezeshkpour, Oli, Mei, LYU, Chen, Banjade, Rudolph,
  Gabriel, Habacker, Delgado, Milli{\`e}re, Garg, Barnes, Saurous, Arakawa,
  Raymaekers, Frank, Sikand, Novak, Sitelew, Lebras, Liu, Jacobs, Zhang,
  Salakhutdinov, Chi, Lee, Stovall, Teehan, Yang, Singh, Mohammad, Anand,
  Dillavou, Shleifer, Wiseman, Gruetter, Bowman, Schoenholz, Han, Kwatra, Rous,
  Ghazarian, Ghosh, Casey, Bischoff, Gehrmann, Schuster, Sadeghi, Hamdan, Zhou,
  Srivastava, Shi, Singh, Asaadi, Gu, Pachchigar, Toshniwal, Upadhyay, Debnath,
  Shakeri, Thormeyer, Melzi, Reddy, Makini, hwan Lee, Torene, Hatwar, Dehaene,
  Divic, Ermon, Biderman, Lin, Prasad, Piantadosi, Shieber, Misherghi,
  Kiritchenko, Mishra, Linzen, Schuster, Li, Yu, Ali, Hashimoto, Wu, Desbordes,
  Rothschild, Phan, Wang, Nkinyili, Schick, Kornev, Telleen-Lawton, Tunduny,
  Gerstenberg, Chang, Neeraj, Khot, Shultz, Shaham, Misra, Demberg, Nyamai,
  Raunak, Ramasesh, Prabhu, Padmakumar, Srikumar, Fedus, Saunders, Zhang,
  Vossen, Ren, Tong, Wu, Shen, Yaghoobzadeh, Lakretz, Song, Bahri, Choi, Yang,
  Hao, Chen, Belinkov, Hou, Hou, Bai, Seid, Xinran, Zhao, Wang, Wang, Wang, and
  Wu}]{Srivastava2022BeyondTI}
Aarohi Srivastava, Abhinav Rastogi, Abhishek~B Rao, Abu Awal~Md Shoeb, Abubakar
  Abid, Adam Fisch, Adam~R. Brown, Adam Santoro, Aditya Gupta, Adri{\`a}
  Garriga-Alonso, Agnieszka Kluska, Aitor Lewkowycz, Akshat Agarwal, Alethea
  Power, Alex Ray, Alex Warstadt, Alexander~W. Kocurek, Ali Safaya, Ali Tazarv,
  Alice Xiang, Alicia Parrish, Allen Nie, Aman Hussain, Amanda Askell, Amanda
  Dsouza, Ameet~Annasaheb Rahane, Anantharaman~S. Iyer, Anders~Johan
  Andreassen, Andrea Santilli, Andreas Stuhlmuller, Andrew~M. Dai, Andrew~D.
  La, Andrew~Kyle Lampinen, Andy Zou, Angela Jiang, Angelica Chen, Anh Vuong,
  Animesh Gupta, Anna Gottardi, Antonio Norelli, Anu Venkatesh, Arash
  Gholamidavoodi, Arfa Tabassum, Arul Menezes, Arun Kirubarajan, Asher
  Mullokandov, Ashish Sabharwal, Austin Herrick, Avia Efrat, Aykut Erdem, Ayla
  Karakacs, Bridget~R. Roberts, Bao~Sheng Loe, Barret Zoph, Bartlomiej
  Bojanowski, Batuhan Ozyurt, Behnam Hedayatnia, Behnam Neyshabur, Benjamin
  Inden, Benno Stein, Berk Ekmekci, Bill~Yuchen Lin, Blake~Stephen Howald,
  Cameron Diao, Cameron Dour, Catherine Stinson, Cedrick Argueta, C'esar~Ferri
  Ram'irez, Chandan Singh, Charles Rathkopf, Chenlin Meng, Chitta Baral, Chiyu
  Wu, Chris Callison-Burch, Chris Waites, Christian Voigt, Christopher~D.
  Manning, Christopher Potts, Cindy~Tatiana Ramirez, Clara~E. Rivera, Clemencia
  Siro, Colin Raffel, Courtney Ashcraft, Cristina Garbacea, Damien Sileo,
  Daniel~H Garrette, Dan Hendrycks, Dan Kilman, Dan Roth, Daniel Freeman,
  Daniel Khashabi, Daniel Levy, Daniel Gonz'alez, Danny Hernandez, Danqi Chen,
  Daphne Ippolito, Dar Gilboa, David Dohan, D.~Drakard, David Jurgens,
  Debajyoti Datta, Deep Ganguli, Denis Emelin, Denis Kleyko, Deniz Yuret, Derek
  Chen, Derek Tam, Dieuwke Hupkes, Diganta Misra, Dilyar Buzan, Dimitri~Coelho
  Mollo, Diyi Yang, Dong-Ho Lee, Ekaterina Shutova, Ekin~Dogus Cubuk, Elad
  Segal, Eleanor Hagerman, Elizabeth Barnes, Elizabeth~P. Donoway, Ellie
  Pavlick, Emanuele Rodol{\`a}, Emma~FC Lam, Eric Chu, Eric Tang, Erkut Erdem,
  Ernie Chang, Ethan~A. Chi, Ethan Dyer, Ethan Jerzak, Ethan Kim, Eunice~Engefu
  Manyasi, Evgenii Zheltonozhskii, Fan Xia, Fatemeh Siar, Fernando
  Mart'inez-Plumed, Francesca Happ'e, François Chollet, Frieda Rong, Gaurav
  Mishra, Genta~Indra Winata, Gerard de~Melo, Germ{\'a}n Kruszewski,
  Giambattista Parascandolo, Giorgio Mariani, Gloria Wang, Gonzalo
  Jaimovitch-L'opez, Gregor Betz, Guy Gur-Ari, Hana Galijasevic, Han~Sol Kim,
  Hannah Rashkin, Hanna Hajishirzi, Harsh Mehta, Hayden Bogar, Henry Shevlin,
  Hinrich Sch{\"u}tze, Hiromu Yakura, Hongming Zhang, Hubert Wong, Ian Aik-Soon
  Ng, Isaac Noble, Jaap Jumelet, Jack Geissinger, John Kernion, Jacob Hilton,
  Jaehoon Lee, Jaime~Fern{\'a}ndez Fisac, J.~Brooker Simon, James Koppel, James
  Zheng, James Zou, Jan Koco'n, Jana Thompson, Jared Kaplan, Jarema Radom,
  Jascha Sohl-Dickstein, Jason Phang, Jason Wei, Jason Yosinski, Jekaterina
  Novikova, Jelle Bosscher, Jenni Marsh, Jeremy Kim, Jeroen Taal, Jesse Engel,
  Jesujoba~Oluwadara Alabi, Jiacheng Xu, Jiaming Song, Jillian Tang, Jane~W
  Waweru, John Burden, John Miller, John~U. Balis, Jonathan Berant, Jorg
  Frohberg, Jos Rozen, Jos{\'e} Hern{\'a}ndez-Orallo, Joseph Boudeman, Joseph
  Jones, Joshua~B. Tenenbaum, Joshua~S. Rule, Joyce Chua, Kamil Kanclerz, Karen
  Livescu, Karl Krauth, Karthik Gopalakrishnan, Katerina Ignatyeva, Katja
  Markert, Kaustubh~D. Dhole, Kevin Gimpel, Kevin~Ochieng’ Omondi,
  Kory~Wallace Mathewson, Kristen Chiafullo, Ksenia Shkaruta, Kumar Shridhar,
  Kyle McDonell, Kyle Richardson, Laria Reynolds, Leo Gao, Li~Zhang, Liam
  Dugan, Lianhui Qin, Lidia Contreras-Ochando, Louis-Philippe Morency, Luca
  Moschella, Luca Lam, Lucy Noble, Ludwig Schmidt, Luheng He, Luis~Oliveros
  Col'on, Luke Metz, Lutfi~Kerem cSenel, Maarten Bosma, Maarten Sap, Maartje
  ter Hoeve, Madotto Andrea, Maheen~Saleem Farooqi, Manaal Faruqui, Mantas
  Mazeika, Marco Baturan, Marco Marelli, Marco Maru, M~Quintana, Marie
  Tolkiehn, Mario Giulianelli, Martha Lewis, Martin Potthast, Matthew Leavitt,
  Matthias Hagen, M'aty'as Schubert, Medina Baitemirova, Melissa Arnaud,
  Melvin~Andrew McElrath, Michael~A. Yee, Michael Cohen, Mi~Gu, Michael~I.
  Ivanitskiy, Michael Starritt, Michael Strube, Michal Swkedrowski, Michele
  Bevilacqua, Michihiro Yasunaga, Mihir Kale, Mike Cain, Mimee Xu, Mirac
  Suzgun, Monica Tiwari, Mohit Bansal, Moin Aminnaseri, Mor Geva, Mozhdeh
  Gheini, T~MukundVarma, Nanyun Peng, Nathan Chi, Nayeon Lee, Neta Gur-Ari
  Krakover, Nicholas Cameron, Nicholas~S. Roberts, Nicholas Doiron, Nikita
  Nangia, Niklas Deckers, Niklas Muennighoff, Nitish~Shirish Keskar, Niveditha
  Iyer, Noah Constant, Noah Fiedel, Nuan Wen, Oliver Zhang, Omar Agha, Omar
  Elbaghdadi, Omer Levy, Owain Evans, Pablo Casares, Parth Doshi, Pascale Fung,
  Paul~Pu Liang, Paul Vicol, Pegah Alipoormolabashi, Peiyuan Liao, Percy Liang,
  Peter~W. Chang, Peter Eckersley, Phu~Mon Htut, Pi-Bei Hwang, P.~Milkowski,
  Piyush~S. Patil, Pouya Pezeshkpour, Priti Oli, Qiaozhu Mei, QING LYU, Qinlang
  Chen, Rabin Banjade, Rachel~Etta Rudolph, Raefer Gabriel, Rahel Habacker,
  Ram'on~Risco Delgado, Rapha{\"e}l Milli{\`e}re, Rhythm Garg, Richard Barnes,
  Rif~A. Saurous, Riku Arakawa, Robbe Raymaekers, Robert Frank, Rohan Sikand,
  Roman Novak, Roman Sitelew, Ronan Lebras, Rosanne Liu, Rowan Jacobs, Rui
  Zhang, Ruslan Salakhutdinov, Ryan Chi, Ryan Lee, Ryan Stovall, Ryan Teehan,
  Rylan Yang, Sahib~J. Singh, Saif~M. Mohammad, Sajant Anand, Sam Dillavou, Sam
  Shleifer, Sam Wiseman, Samuel Gruetter, Sam Bowman, Samuel~S. Schoenholz,
  Sanghyun Han, Sanjeev Kwatra, Sarah~A. Rous, Sarik Ghazarian, Sayan Ghosh,
  Sean Casey, Sebastian Bischoff, Sebastian Gehrmann, Sebastian Schuster,
  Sepideh Sadeghi, Shadi~Sameh Hamdan, Sharon Zhou, Shashank Srivastava, Sherry
  Shi, Shikhar Singh, Shima Asaadi, Shixiang~Shane Gu, Shubh Pachchigar,
  Shubham Toshniwal, Shyam Upadhyay, Shyamolima Debnath, Siamak Shakeri, Simon
  Thormeyer, Simone Melzi, Siva~Kumar Reddy, Sneha~Priscilla Makini, Soo hwan
  Lee, Spencer~Bradley Torene, Sriharsha Hatwar, Stanislas Dehaene, Stefan
  Divic, Stefano Ermon, Stella~Rose Biderman, Stephanie~C. Lin, Stephen Prasad,
  Steven~T. Piantadosi, Stuart~M. Shieber, Summer Misherghi, Svetlana
  Kiritchenko, Swaroop Mishra, Tal Linzen, Tal Schuster, Tao Li, Tao Yu,
  Tariq~A. Ali, Tatsuo Hashimoto, Te-Lin Wu, Theo Desbordes, Theodore
  Rothschild, Thomas Phan, Tianle Wang, Tiberius Nkinyili, Timo Schick, T.~N.
  Kornev, Timothy Telleen-Lawton, Titus Tunduny, Tobias Gerstenberg, Trenton
  Chang, Trishala Neeraj, Tushar Khot, Tyler~O. Shultz, Uri Shaham, Vedant
  Misra, Vera Demberg, Victoria Nyamai, Vikas Raunak, Vinay~V. Ramasesh,
  Vinay~Uday Prabhu, Vishakh Padmakumar, Vivek Srikumar, William Fedus, William
  Saunders, William Zhang, W~Vossen, Xiang Ren, Xiaoyu~F Tong, Xinyi Wu, Xudong
  Shen, Yadollah Yaghoobzadeh, Yair Lakretz, Yang Song, Yasaman Bahri, Ye~Ji
  Choi, Yichi Yang, Yiding Hao, Yifu Chen, Yonatan Belinkov, Yu~Hou, Yu~Hou,
  Yuntao Bai, Zachary Seid, Zhao Xinran, Zhuoye Zhao, Zi~Fu Wang, Zijie~Jay
  Wang, Zirui Wang, and Ziyi Wu. 2022.
\newblock \href {https://arxiv.org/abs/2206.04615} {Beyond the imitation game:
  Quantifying and extrapolating the capabilities of language models}.

\bibitem[{Steinert-Threlkeld et~al.(2022)Steinert-Threlkeld, Zhou, Liu, and
  Downey}]{steinert2022emergent}
Shane Steinert-Threlkeld, Xuhui Zhou, Zeyu Liu, and CM~Downey. 2022.
\newblock Emergent communication fine-tuning (ec-ft) for pretrained language
  models.
\newblock In \emph{Emergent Communication Workshop at ICLR 2022}.

\bibitem[{Stiennon et~al.(2020)Stiennon, Ouyang, Wu, Ziegler, Lowe, Voss,
  Radford, Amodei, and Christiano}]{stiennon2020feedback}
Nisan Stiennon, Long Ouyang, Jeffrey Wu, Daniel Ziegler, Ryan Lowe, Chelsea
  Voss, Alec Radford, Dario Amodei, and Paul~F Christiano. 2020.
\newblock \href
  {https://proceedings.neurips.cc/paper/2020/file/1f89885d556929e98d3ef9b86448f951-Paper.pdf}
  {Learning to summarize with human feedback}.
\newblock In \emph{Advances in Neural Information Processing Systems},
  volume~33, pages 3008--3021. Curran Associates, Inc.

\bibitem[{Tauzin and Gergely(2018)}]{Tauzin2018-na}
Tibor Tauzin and Gy{\"o}rgy Gergely. 2018.
\newblock \href {http://dx.doi.org/10.1038/s41598-018-27804-4} {Communicative
  mind-reading in preverbal infants}.
\newblock \emph{Scientific reports}, 8(1):9534.

\bibitem[{Tenney et~al.(2019)Tenney, Das, and Pavlick}]{Tenney2019-fn}
Ian Tenney, Dipanjan Das, and Ellie Pavlick. 2019.
\newblock \href {http://arxiv.org/abs/1905.05950} {{BERT} rediscovers the
  classical {NLP} pipeline}.
\newblock In \emph{{ACL}}.

\bibitem[{Tomasello et~al.(2005)Tomasello, Carpenter, Call, Behne, and
  Moll}]{Tomasello2005-intentions}
Michael Tomasello, Malinda Carpenter, Josep Call, Tanya Behne, and Henrike
  Moll. 2005.
\newblock Understanding and sharing intentions: the origins of cultural
  cognition.
\newblock \emph{Behavioral and Brain Sciences}, 28(5):675--91; discussion
  691--735.

\bibitem[{Ullman(2023)}]{ullman2023large}
Tomer Ullman. 2023.
\newblock Large language models fail on trivial alterations to theory-of-mind
  tasks.
\newblock \emph{arXiv preprint arXiv:2302.08399}.

\bibitem[{Valle et~al.(2015)Valle, Massaro, Castelli, and
  Marchetti}]{Valle2015-yb}
Annalisa Valle, Davide Massaro, Ilaria Castelli, and Antonella Marchetti. 2015.
\newblock \href {http://dx.doi.org/10.5964/ejop.v11i1.829} {Theory of mind
  development in adolescence and early adulthood: The growing complexity of
  recursive thinking ability}.
\newblock \emph{European journal of psychological assessment: official organ of
  the European Association of Psychological Assessment}, 11(1):112--124.

\bibitem[{Wang et~al.(2007)Wang, Carley, Zeng, and Mao}]{Wang2007-xa}
Fei-Yue Wang, Kathleen~M Carley, Daniel Zeng, and Wenji Mao. 2007.
\newblock \href {http://dx.doi.org/10.1109/MIS.2007.41} {Social computing: From
  social informatics to social intelligence}.
\newblock \emph{IEEE intelligent systems}, 22(2):79--83.

\bibitem[{Wang et~al.(2019)Wang, Shi, Kim, Oh, Yang, Zhang, and
  Yu}]{wang-etal-2019-persuasion}
Xuewei Wang, Weiyan Shi, Richard Kim, Yoojung Oh, Sijia Yang, Jingwen Zhang,
  and Zhou Yu. 2019.
\newblock \href {https://doi.org/10.18653/v1/P19-1566} {Persuasion for good:
  Towards a personalized persuasive dialogue system for social good}.
\newblock In \emph{Proceedings of the 57th Annual Meeting of the Association
  for Computational Linguistics}, pages 5635--5649, Florence, Italy.
  Association for Computational Linguistics.

\bibitem[{Wang et~al.(2022)Wang, Zhong, Xu, and Wang}]{Wang2022-vi}
Yuanfei Wang, Fangwei Zhong, Jing Xu, and Yizhou Wang. 2022.
\newblock \href {http://arxiv.org/abs/2111.09189} {{ToM2C}: Target-oriented
  multi-agent communication and cooperation with theory of mind}.
\newblock In \emph{{ICLR}}. arxiv.org.

\bibitem[{Wellman(1992)}]{Wellman1992-zj}
Henry~M Wellman. 1992.
\newblock \href {https://psycnet.apa.org/fulltext/1990-98613-000.pdf} {The
  child's theory of mind}.
\newblock \emph{The MIT Press series in learning, development, and conceptual
  change.}, 358.

\bibitem[{Wells and Bridges(1981)}]{wells1981learning}
Gordon Wells and Allayne Bridges. 1981.
\newblock \emph{Learning through interaction: volume 1: the study of language
  development}, volume~1.
\newblock Cambridge University Press.

\bibitem[{West et~al.(2022)West, Bhagavatula, Hessel, Hwang, Jiang, Bras, Lu,
  Welleck, and Choi}]{west2022symbolic}
Peter West, Chandra Bhagavatula, Jack Hessel, Jena~D Hwang, Liwei Jiang,
  Ronan~Le Bras, Ximing Lu, Sean Welleck, and Yejin Choi. 2022.
\newblock Symbolic knowledge distillation: from general language models to
  commonsense models.
\newblock In \emph{NAACL}.

\bibitem[{Wimmer and Perner(1983)}]{wimmer1983beliefs}
Heinz Wimmer and Josef Perner. 1983.
\newblock Beliefs about beliefs: Representation and constraining function of
  wrong beliefs in young children's understanding of deception.
\newblock \emph{Cognition}, 13(1):103--128.

\bibitem[{Wischmeyer and Rademacher(2020)}]{Wischmeyer2020-bz}
Thomas Wischmeyer and Timo Rademacher, editors. 2020.
\newblock \href {https://link.springer.com/book/10.1007/978-3-030-32361-5}
  {\emph{Regulating Artificial Intelligence}}.
\newblock Springer, Cham.

\bibitem[{Yang et~al.(2017)Yang, Blunsom, Dyer, and Ling}]{Yang2017-uk}
Zichao Yang, Phil Blunsom, Chris Dyer, and Wang Ling. 2017.
\newblock \href {http://arxiv.org/abs/1611.01628} {{Reference-Aware} language
  models}.
\newblock In \emph{{EMNLP}}.

\bibitem[{Zadeh et~al.(2019)Zadeh, Chan, Liang, Tong, and
  Morency}]{zadeh2019social}
Amir Zadeh, Michael Chan, Paul~Pu Liang, Edmund Tong, and Louis-Philippe
  Morency. 2019.
\newblock Social-iq: A question answering benchmark for artificial social
  intelligence.
\newblock In \emph{Proceedings of the IEEE/CVF Conference on Computer Vision
  and Pattern Recognition}, pages 8807--8817.

\bibitem[{Zhang et~al.(2022{\natexlab{a}})Zhang, Van~Durme, Li, and
  Stengel-Eskin}]{Zhang2022-nk}
Chenyu Zhang, Benjamin Van~Durme, Zhuowan Li, and Elias Stengel-Eskin.
  2022{\natexlab{a}}.
\newblock \href {http://arxiv.org/abs/2205.01850} {Visual commonsense in
  pretrained unimodal and multimodal models}.
\newblock In \emph{{NAACL} 2022}.

\bibitem[{Zhang et~al.(2022{\natexlab{b}})Zhang, Roller, Goyal, Artetxe, Chen,
  Chen, Dewan, Diab, Li, Lin et~al.}]{zhang2022opt}
Susan Zhang, Stephen Roller, Naman Goyal, Mikel Artetxe, Moya Chen, Shuohui
  Chen, Christopher Dewan, Mona Diab, Xian Li, Xi~Victoria Lin, et~al.
  2022{\natexlab{b}}.
\newblock Opt: Open pre-trained transformer language models.
\newblock \emph{arXiv preprint arXiv:2205.01068}.

\bibitem[{Zhang and Zhou(2022)}]{Zhang2022-ez}
Xiaowen Zhang and Peng Zhou. 2022.
\newblock Linguistic cues facilitate children's understanding of
  belief-reporting sentences.
\newblock \emph{First Lang.}, 42(1):51--80.

\bibitem[{Zhang et~al.(2022{\natexlab{c}})Zhang, Yu, Zhu, and
  Jiang}]{Zhang2022-du}
Zhihan Zhang, Wenhao Yu, Chenguang Zhu, and Meng Jiang. 2022{\natexlab{c}}.
\newblock A unified {Encoder-Decoder} framework with entity memory.
\newblock In \emph{{EMNLP}}.

\bibitem[{Zhou et~al.(2021)Zhou, Lee, Selvam, Lee, Lin, and
  Ren}]{Zhou2021PretrainingTT}
Wangchunshu Zhou, Dong-Ho Lee, Ravi~Kiran Selvam, Seyeon Lee, Bill~Yuchen Lin,
  and Xiang Ren. 2021.
\newblock Pre-training text-to-text transformers for concept-centric common
  sense.
\newblock \emph{ICLR}, abs/2011.07956.

\bibitem[{Zhu et~al.(2021)Zhu, Neubig, and Bisk}]{Zhu2021-zj}
Hao Zhu, Graham Neubig, and Yonatan Bisk. 2021.
\newblock \href {http://proceedings.mlr.press/v139/zhu21d.html} {Few-shot
  language coordination by modeling theory of mind}.
\newblock In \emph{Proceedings of the 38th International Conference on Machine
  Learning}, volume 139 of \emph{Proceedings of Machine Learning Research},
  pages 12901--12911. PMLR.

\end{thebibliography}

\newpage
\clearpage
\appendix
\section{\socialIQa Details}
\label{supp:siqa-deets}

\begin{table}[t]
\centering
\footnotesize
    \begin{tabular}{@{}lp{4.8cm}r@{}}
    \toprule
        dimension & definition & count \\ \midrule
        \multicolumn{3}{c}{\textit{Before the event}}\\
        ~~xIntent & Why does X cause the event &  238           \\
        ~~xNeed & What does X need to do before the event   & 228           \\
        ~~xAttr  & How would X be described?  & 287           \\\midrule
        \multicolumn{3}{c}{\textit{After the event}} \\
        \textit{Effect}& &  \textit{218} \\
        ~~xEffect  & What effects does the event have on X?  & 99            \\
        ~~oEffect  & What effects does the event have on others?    & 119           \\
        \textit{React} && \textit{415} \\
        ~~xReact & How does X feel after the event?     & 223           \\
        ~~oReact & How do others feel after the event?   & 192           \\
        \textit{Want} & & \textit{568} \\
        ~~xWant & What would X likely want to do after the event? & 338           \\
        ~~oWant & What would others likely want to do after the event?  & 230     \\\midrule
        
        \textbf{total} &&  \textbf{1954} \\ \bottomrule
    \end{tabular}
\caption{\socialIQa dev. set statistics, broken down by question reasoning type and their definitions from \atomic.}
\label{tab:siqa-stats}
\end{table}

% \paragraph{Data Access}
% One of the authors of this paper is also a creator of the \socialIQa dataset, which is publicly available.\footnote{\url{https://leaderboard.allenai.org/socialiqa/submissions/get-started}}

\subsection{Data Preprocessing}
We downloaded the \socialIQa training and dev. datasets from the publicly available \socialIQa website.\footnote{\url{http://maartensap.com/social-iqa/data/socialIQa_v1.4_withDims.tgz}} 
This version of the \socialIQa dataset contains the original \atomic dimensions that workers were prompted with to create a question, as well as the correspondence between questions and which character they focus on (agent or other).
To ensure consistency, for each context, question, and answer, we normalize the casing to start with a capital letter if the text does not already.

% \maarten{Include details on how we linked each trn/dev question to the person it is referring to? How do we say we got access to internal versions that came from ATOMIC?}

\subsection{Further \socialIQa results}
\label{ssec:siqa-results-more}
In addition to results discussed in \S\ref{ssec:siqa-results}, we report further \socialIQa results here.

\paragraph{\socialIQa broken down by reasoning dimension.}

\begin{figure}[t]
     \centering
     \includegraphics[width=.9\columnwidth,trim=3cm 3cm 3cm 3cm,clip]{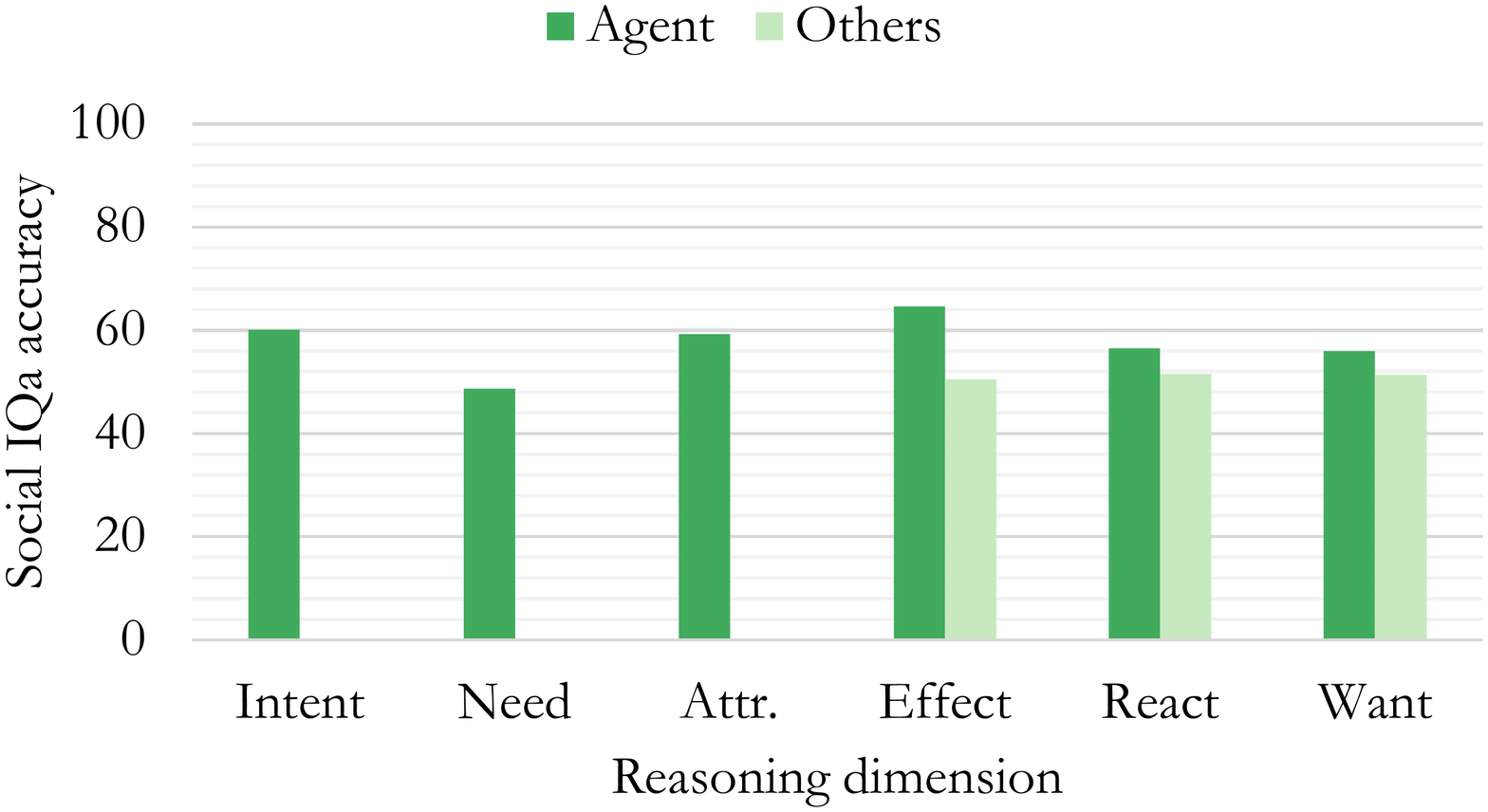}
     \caption{Comparing the accuracy of \gptDavinci (35-shot) when over all nine reasoning dimensions.}
     \label{fig:siqa-acc-by-dims}
\end{figure}

We break down the best performing \gptDavinci (35-shot) setup by reasoning dimension.
Shown in Fig. \ref{fig:siqa-acc-by-dims}, we find that \gptDavinci struggles most with questions related to what people needed to do before a situation could take place (Need).
Conversely, questions related to a situation's agent's intent (Intent) and the effect of the situation on the agent (Effect) are seemingly easier for \gptDavinci.
\newText{Future work should explore \ptlms's reasoning abilities along each of these dimensions in further detail.}

\begin{figure*}[t]
    \centering
    \includegraphics[width=.95\textwidth,clip,trim=2.2cm 4.2cm 2.2cm 4.2cm]{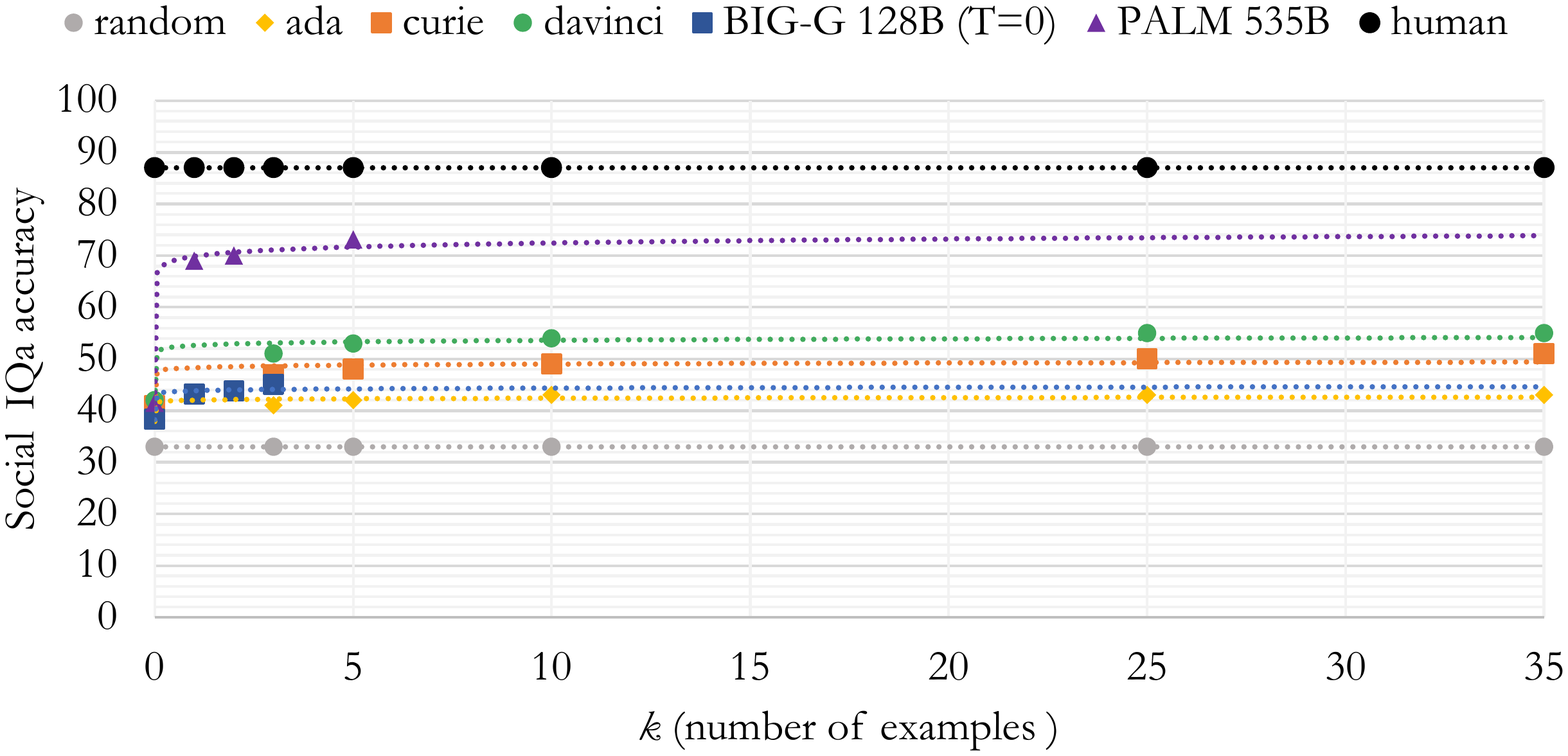}
    \caption{
    Expanded version of Fig.~\ref{fig:siqa-acc-by-k-size}, depicting the accuracy on the \socialIQa dev. set, broken down by \ptlm model type and size, as well as number of few-shot examples ($k$).
    Here, we also include the accuracy results of the PaLM \cite{Chowdhery2022palm} and BIG-G \cite{Srivastava2022BeyondTI} \ptlms, taken from the \texttt{\href{https://github.com/google/BIG-bench/tree/cffb32cdfa42658dddce1b78be82a46a63c03440/bigbench/benchmark_tasks/social_iqa/}{BIG-Bench github}} repository on 2022-11-10.
    % Expanded version of accuracy with results PaLM \cite{Chowdhery2022palm} Accuracy results from the BIG-Bench \cite{Srivastava2022BeyondTI} \socialIQa experiments. 
    % PALM and BIG-G accuracies taken on 2022-11-10 from .
    }
    \label{fig:bigbench-socialIQa}
\end{figure*}

\paragraph{BIG-Bench and PaLM results on \socialIQa.}
To further corroborate that \ptlms struggle with \socialIQa, we \newText{show the performance of the non-publicly available BIG-G \cite{Srivastava2022BeyondTI} and PaLM \cite{Chowdhery2022palm} \ptlms, along with the \gptT models, in Fig.~\ref{fig:bigbench-socialIQa}.
Both models are proprietary \ptlms developed and tested on the 200+ datasets in BIG-Bench by Google / DeepMind.}
% These results were compiled as part of BIG-Bench \cite{Srivastava2022BeyondTI}, a benchmark of over 200 tasks for measuring zero- or few-shot abilities of \ptlms of various scales.
% plot from the contemporaneously released BIG-Bench \cite{Srivastava2022BeyondTI}, a benchmark of over 200 tasks for measuring zero- or few-shot abilities of \ptlms of various scales.
% In their experiments, BIG-Bench authors ran several large \ptlms (both in-house and publicly available) in up to a 5-shot setting.

While they are not discussed in the main BIG-Bench paper, the \socialIQa results for few-shot settings up to $k$=3 for BIG-G and $k$=5 for PaLM can be found on the
\texttt{\href{https://github.com/google/BIG-bench/tree/cffb32cdfa42658dddce1b78be82a46a63c03440/bigbench/benchmark_tasks/social_iqa/}{BIG-Bench github}} website (accessed on 2022-11-10).
% \texttt{\href{https://github.com/google/BIG-bench/tree/586baacd1de83946d183955494c8db4db571a76e/bigbench/benchmark_tasks/social_iqa/}{the BIG-Bench github website}}. 
Plotted in Fig.~\ref{fig:bigbench-socialIQa}, \newText{both the BIG-G and PaLM \ptlms lag behind humans with 45\% and 73\% peak accuracy, respectively.}
\oldText{results show that the \ptlms tested by BIG-Bench authors achieve even lower performance ($<$50\% accuracy) compared to the 35-shot setting we explore in our work (55\% accuracy).}

% \begin{figure}[t]
%     \centering
%     \includegraphics[width=.9\columnwidth]{graphics/BigBench-SocialIQa-accuracy.pdf}
%     \caption{Accuracy results from the BIG-Bench \cite{Srivastava2022BeyondTI} \socialIQa experiments. 
%     Figure taken on 2022-06-15 from \texttt{\href{https://github.com/google/BIG-bench/tree/586baacd1de83946d183955494c8db4db571a76e/bigbench/benchmark_tasks/social_iqa/}{the BIG-Bench github}}.}
%     \label{fig:bigbench-socialIQa}
% \end{figure}

% \clearpage

\section{\tomi Details}
\label{supp:tomi-details}
\subsection{Data Preprocessing}
We generated \tomi stories using \texttt{\href{https://github.com/facebookresearch/ToMi}{the github repository}} provided by \citet{le-etal-2019-revisiting}.
% We generated the stories using the code provided by \citet{le-etal-2019-revisiting}.\footnote{\url{https://github.com/facebookresearch/ToMi}}
The code generated 5994 training and 5994 dev. stories.
From those, we removed the story-question pairs which wrongly answered \ToM-requiring questions from an omniscient perspective (i.e., answered \mind-\falseBelief questions from an omniscient perspective instead of the perspective of the character) which we noticed upon manual data inspection.\footnote{We do not know why these datapoints were generated.}
After this filtering, 5190 training and 5170 dev. stories remained.

For the final \tomi dev. set, we used stratified sampling to obtain similar numbers of story-question pairs for all types (\reality, \memory, \firstorder-\falseBelief, \firstorder-\trueBelief, \secondorder-\falseBelief and \secondorder-\trueBelief).
The exact counts are shown in Tab.~\ref{tab:tomi-stats}.
\newText{We release our final preprocessed \tomi dev. dataset at \url{http://maartensap.com/neuralToM/ToMi-finalNeuralTOM.csv}}
\oldText{We plan to make our processing code and \tomi story corpus publicly available upon acceptance.}

% \maarten{Include details about how we removed any story that assumed an omniscient reader, which for some reason the code generates. As a conservative estimate, we remove all story-question pairs where a story has someone absent while an object is moved and the question is first or second order.}

\begin{table}[t]
\centering
\footnotesize
    \begin{tabular}{@{}lr@{}}
    \toprule
    ~~\facts                       & 519  \\
    ~~~~\memory                    & 278  \\
    ~~~~\reality                   & 241  \\
    ~~\mind                       & 1342 \\
    ~~~~\mind-\trueBelief          & 778  \\
    ~~~~~~\firstorder-\trueBelief   & 389  \\
    ~~~~~~\secondorder-\trueBelief  & 389  \\
    ~~~~\mind-\falseBelief         & 564  \\
    ~~~~~~\firstorder-\falseBelief  & 231  \\
    ~~~~~~\secondorder-\falseBelief & 333  \\ \midrule
    total                        & 1861 \\ \bottomrule
    \end{tabular}
\caption{\tomi dev. set statistics, broken down by question reasoning type.}
\label{tab:tomi-stats}
\end{table}

\subsection{Further \tomi results}
\begin{figure}[h]
     \centering
     \includegraphics[width=\columnwidth,trim=6cm 2cm 5cm 2cm,clip]{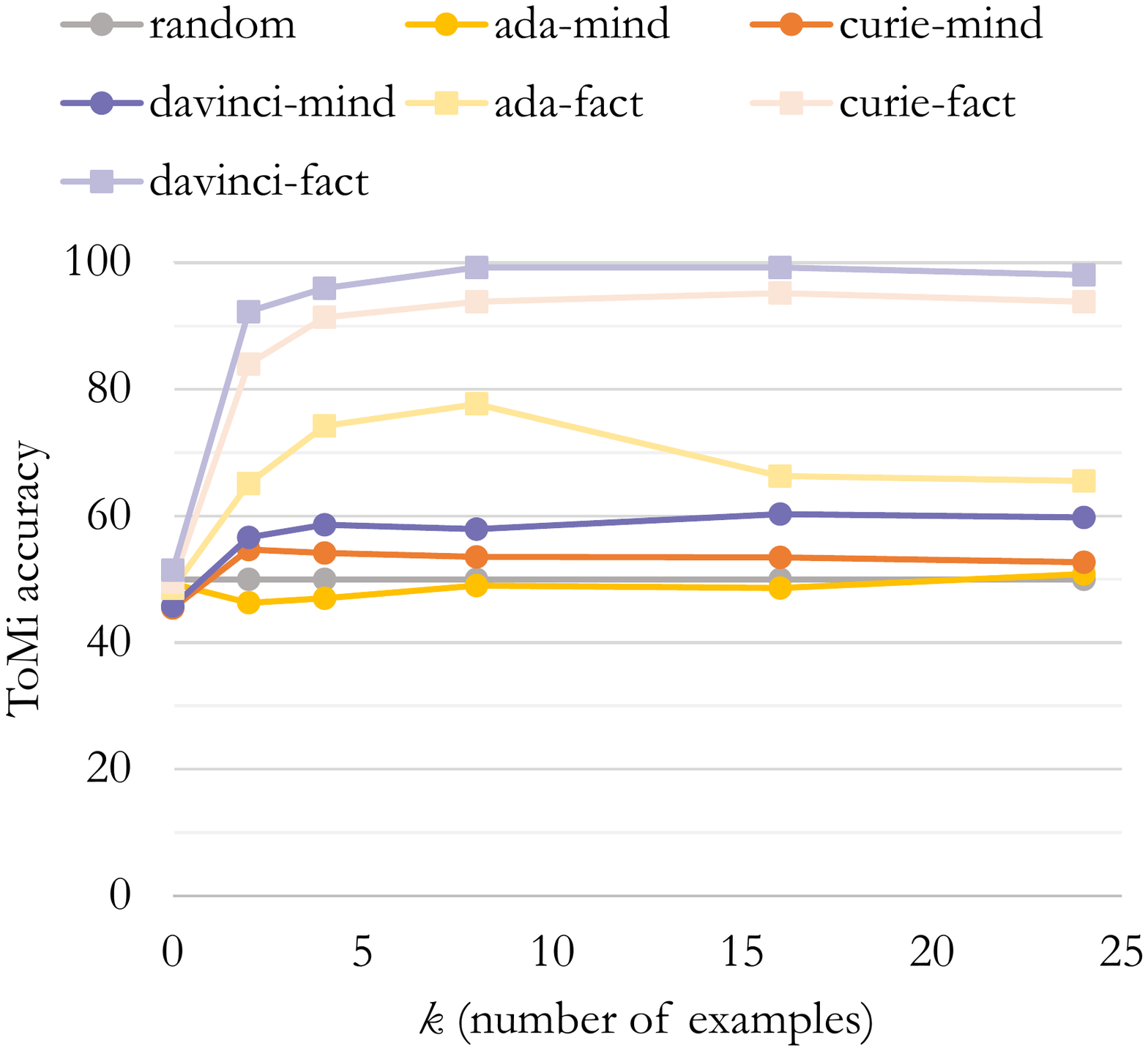}
     \caption{Examining the accuracy of \gptT of different sizes with different number of few-shot examples ($k$) on \tomi-\mind vs. \tomi-\facts questions.}
     \label{fig:tomi-acc-by-examples-size-mindVsFact}
\end{figure}

\begin{figure}[h]
     \centering
     \includegraphics[width=.9\columnwidth,trim=3cm 3cm 3cm 3cm,clip]{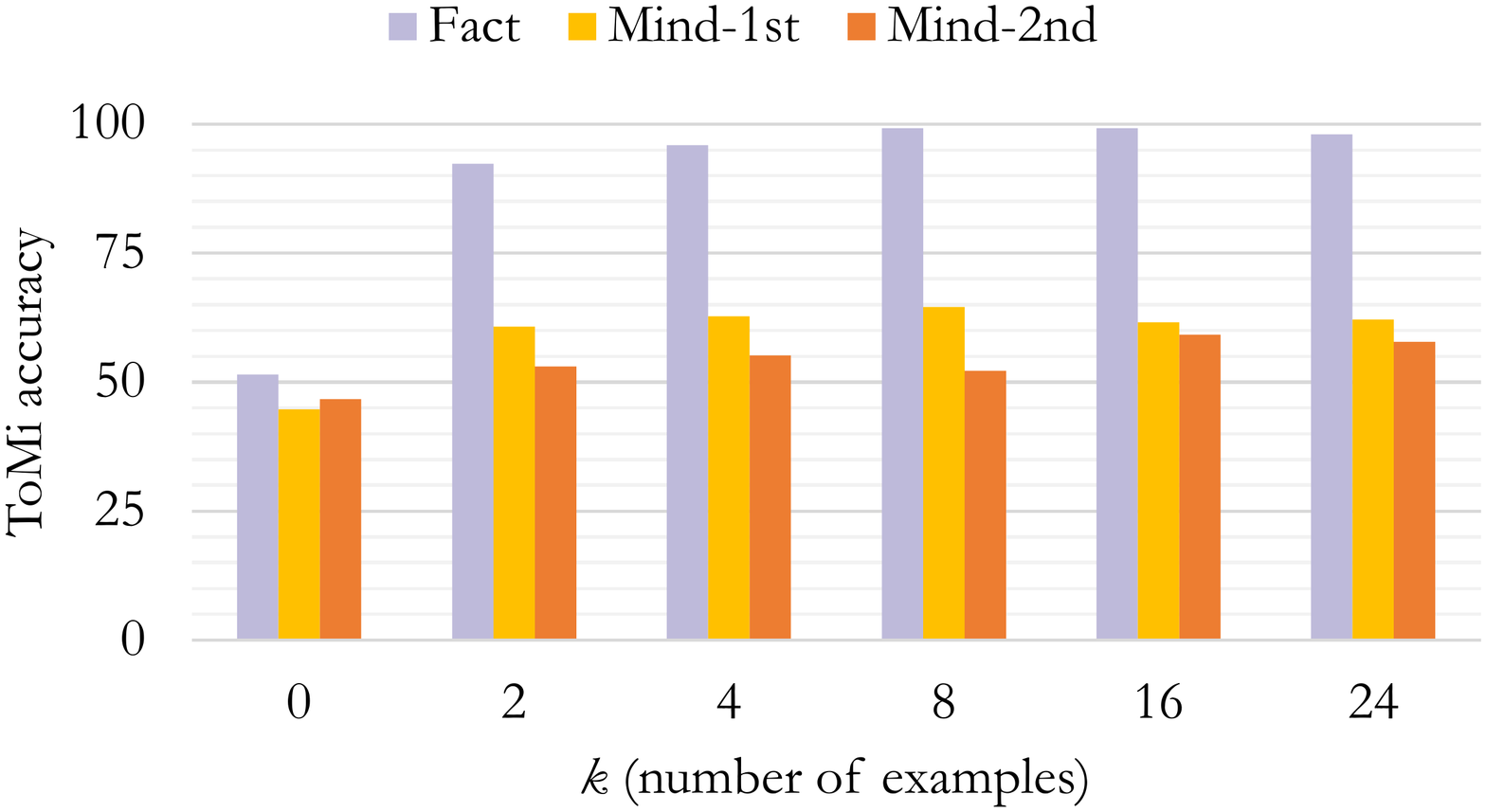}
     \caption{Comparing the accuracy of \gptDavinci by the question reasoning type, specifically \facts vs. \firstorder vs. \secondorder.}
     \label{fig:tomi-acc-by-examples-order}
\end{figure}

\begin{figure}[h]
    \centering
    \includegraphics[width=.9\columnwidth,trim=3cm 3cm 3cm 3cm,clip]{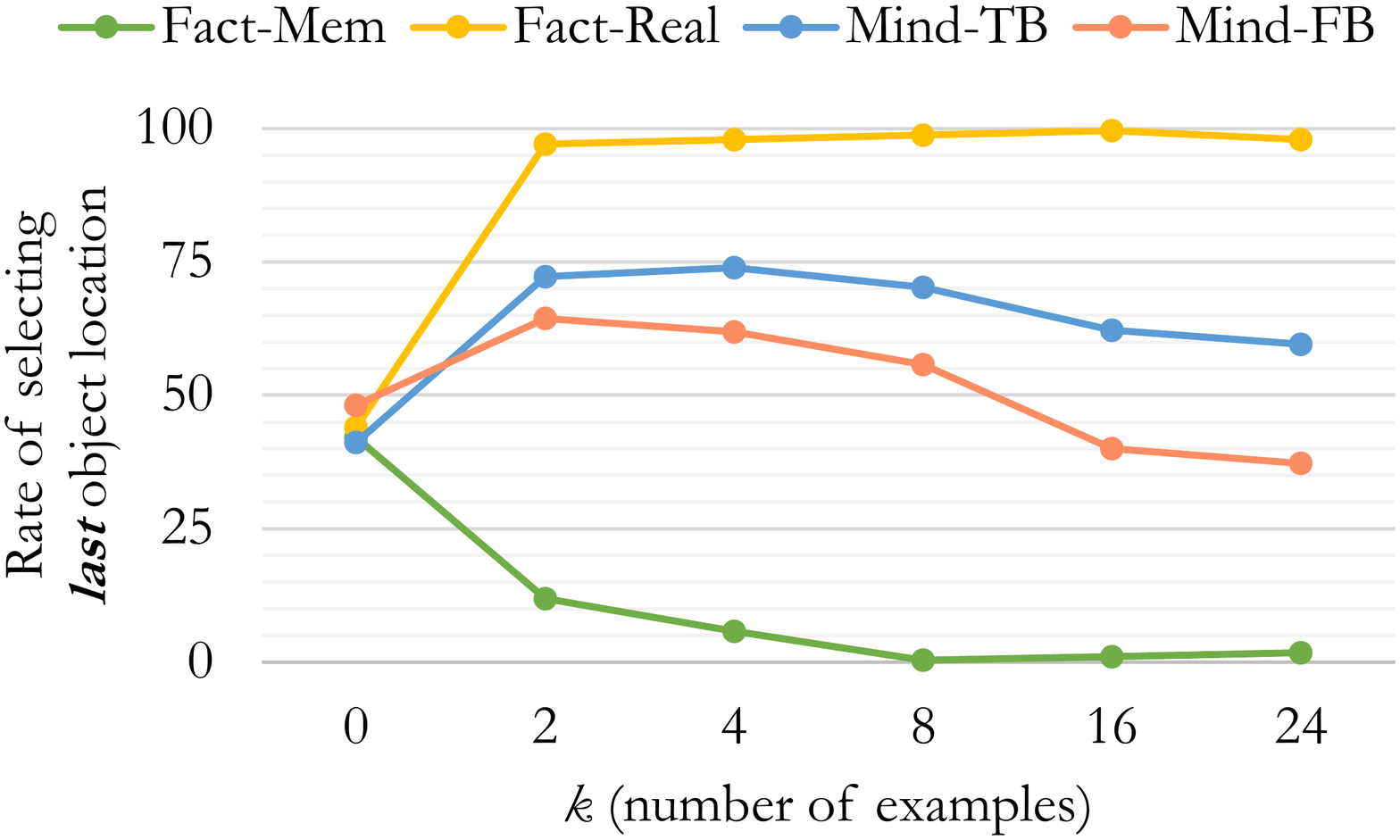}
    \caption{We plot the proportion of examples for which \gptDavinci selects the last object location (i.e., in ``reality'').}
    \label{fig:tomi-pctReal-by-k}
\end{figure}

Shown in Fig.~\ref{fig:tomi-acc-by-examples-size-mindVsFact}-\ref{fig:tomi-pctReal-by-k}, we provide additional results to supplement those in \S\ref{ssec:tomi-results}.

\paragraph{Performance by model size, number of examples, and \mind versus \facts.}
In Fig.~\ref{fig:tomi-acc-by-examples-size-mindVsFact}, we show the different accuracies that \gptT models of various sizes, prompted with various number of examples, for \tomi \mind and \facts questions.
This plot shows the same accuracies as Fig.~\ref{fig:tomi-acc-by-size}, with the addition of the \facts accuracies.
These results show that in the few-shot prompting setup, \gptCurie and \gptDavinci can achieve near perfect performance on factual questions about object locations (\facts), but struggle substantially more on questions related to mental states (\mind).
Surprisingly, \gptAda struggles with both factual and mental state questions, possibly due to its smaller size.

\paragraph{Performance by question order.}
In Fig.~\ref{fig:tomi-acc-by-examples-order}, we break the \gptDavinci performance down by \ToM order (i.e., \firstorder, \secondorder).
Results show that with a number of examples between 2 and 16, \gptDavinci performs better on \firstorder questions (e.g., ``Where will Sally look for the ball?'') and struggles more with \secondorder questions (e.g., ``Where does Ann think that Sally will look for the ball?'').
This difference is somewhat diminished but still present for $k$=24 few-shot examples.
These results somewhat mirror how humans struggle with increasingly higher-order \ToM questions \cite{Valle2015-yb}.

\paragraph{Recency bias in predictions.}
We further examine the results from \S\ref{ssec:tomi-results}, looking at \gptDavinci's rate of predicting the location where the object was moved to (i.e., \reality).
Shown in Fig.~\ref{fig:tomi-pctReal-by-k}, \gptDavinci accurately learns to almost always predict the last object location for \facts-\reality questions, and almost never for \facts-\memory locations.

Interestingly, the rates of selecting the last object location for \mind questions follows a concave pattern. 
This helps shed light onto the concave accuracy pattern seen in Fig.~\ref{fig:tomi-acc-by-mind} for \mind-\trueBelief (and convex pattern for \mind-\falseBelief).
Likely, in the few-shot setting with $2<k<8$, \gptDavinci defaults to the most recently mentioned object location due to recency bias, which  has been previously documented in \ptlms \cite{oconnor-andreas-2021-context}.

\section{\gptT Access and Probing Details}\label{supp:gpt3-deets}
To probe our language models, we use a $k$-shot language probing setup, following \citet{brown2020language}.
% we use the development set of the \socialIQa, which consists of 1,954 QA tuples.
% Following \citet{brown2020language}, we use a $k$-shot language probing setup to obtain predictions.
Specifically, we concatenate the context ($c$) and question ($q$) together with proper punctuation, and assign the model prediction to the answer ($a_i$, $i\in{1,2,3}$) with the highest conditional likelihood under the language model: $ \argmax_i p_{\text{LM}}(a_i\mid c,q, \mathcal{C}_k)$
where $\mathcal{C}_k$ denotes the $k$ training examples, for which we provide the context, question, and correct answer concatenated.
Note that we explored various probing setups and formats, such as QA-oriented formats and normalizing by marginal likelihood of each answer $p_{\text{LM}}(a)$ \citep[as also explored in][]{brown2020language}, but found very little difference in performance.

We access \gptT through the \href{https://openai.com/api/}{OpenAI API}.

% \vspace{35em}

\newpage
% \onecolumn
% \vspace{2em}
% \Large{Revisiting}

\newcommand{\newtext}[1]{{\color{green!50!black}#1}\xspace}

\newcommand{\davincitwo}{GPT-3.5-IFT\xspace}
\newcommand{\davincithree}{GPT-3.5-RLHF\xspace}
\newcommand{\gptturbo}{GPT-3.5-Turbo\xspace}
\newcommand{\gptfour}{GPT-4\xspace}

\begin{table*}[t!]
    \centering
    \small
    \begin{tabular}{@{}cccp{7.5cm}@{}}
    \toprule    
         model & OpenAI endpoint & Probing type & Description \\ \midrule
         \gptDavinci & \texttt{davinci} & LM, MC & GPT-3 model, pretrained solely on the language modeling objective on large amounts of web text \cite{brown2020language}.\\
         \davincitwo & \texttt{text-davinci-002} & LM, MC & GPT-3 based model, finetuned to follow instructions and generate ``high-quality'' text using supervised finetuning \cite{ouyang2022gpt3instruct}.\\
         \davincithree  & \texttt{text-davinci-003} & LM, MC & Similar to GPT-3.5-IFT, but finetuned with reinforcement learning from human feedback (RLHF) instead of supervised finetuning \cite{ouyang2022gpt3instruct}. \\
         \gptturbo & \texttt{gpt-3.5-turbo-0301} & MC & Version of GPT-3.5-RLHF that is optimized for chat at 1/10th the computational cost (snapshot model from March 1st 2023) \cite{ouyang2022gpt3instruct}. \\
         \gptfour & \texttt{gpt-4-0314} & MC & Newest model from OpenAI, claimed to be more capable than previous models (snapshot model from March 14th 2023) \cite{OpenAI2023gpt4}. \\ \bottomrule
    \end{tabular}
    \caption{List of new OpenAI models and APIs that we query with our \socialIQa and \tomi tasks. Probing type denotes whether the API allows for scoring candidate answers independently (LM) or if it requires showing candidate answers in multiple choice format (MC).}
    \label{tab:openai-models}
\end{table*}

\section{What About ChatGPT or GPT-4? Effect of Instruction-tuning \& RLFH}

% \twocolumn

% \noindent\newtext{}

Our analyses in this paper have focused on large language models that are simply trained on the language modeling objective (e.g., GPT-2, GPT-3). 
However, in recent years, many improvements in \ptlms have come from \textit{instruction-finetuning} (IFT) and reinforcement learning from human feedback (RLHF), which are the key to the success of ChatGPT \cite[or GPT-3.5;][]{ouyang2022gpt3instruct} and GPT-4 \cite{OpenAI2023gpt4}.
Despite the opacity of how these models were trained,\footnote{OpenAI has stated that they will not be releasing any useful details. In their system report for GPT-4, they state: ``\textit{Given both the competitive landscape and the safety implications of large-scale models like GPT-4, this report contains no further details about the architecture (including model size), hardware, training compute, dataset construction, training method, or similar}" \cite{OpenAI2023gpt4}.} 
the question of whether they have better neural Theory of Mind than LM-objective-only \ptlms is still of interest.

We quantify the performance of the newer set of OpenAI models (listed in Table \ref{tab:openai-models}) on a randomly selected 400 examples from the \socialIQa and \tomi datasets. 
Since these models have been increasingly capable, we focus on examining only their performance in zero-shot settings, as a stress-test of their \ToM abilities. %\maarten{maybe we can motivate this better?}
% Due to the new models being optimized for chat, t
The newer APIs (\gptturbo, \gptfour) no longer provide the ability to score sequences, and thus, prevent the language modeling probing setup used in \S\ref{sec:socialIQa} and \S\ref{sec:ToMi}.
As such, we also compare two types of probing: language modeling as described in \S\ref{supp:gpt3-deets} (\textit{LM probing}), and multiple-choice probing, where we provide the two or three answer candidates prepended with letter choices (A, B, C) and prompt the model to generate the letter for the correct answer (\textit{MC probing}).\footnote{For example, for \socialIQa, we prepend the example with the instructions ``\texttt{You are a multiple-choice answering system that responds with either A, B, or C.}''}

\vspace{1em}
\subsection{\socialIQa: Social Intelligence and Social Commonsense}
\begin{figure*}[t]
    \centering
    \begin{subfigure}[b]{.95\columnwidth}
        \centering
        \includegraphics[height=4.2cm,clip,trim=2cm 4cm 2cm 4cm]{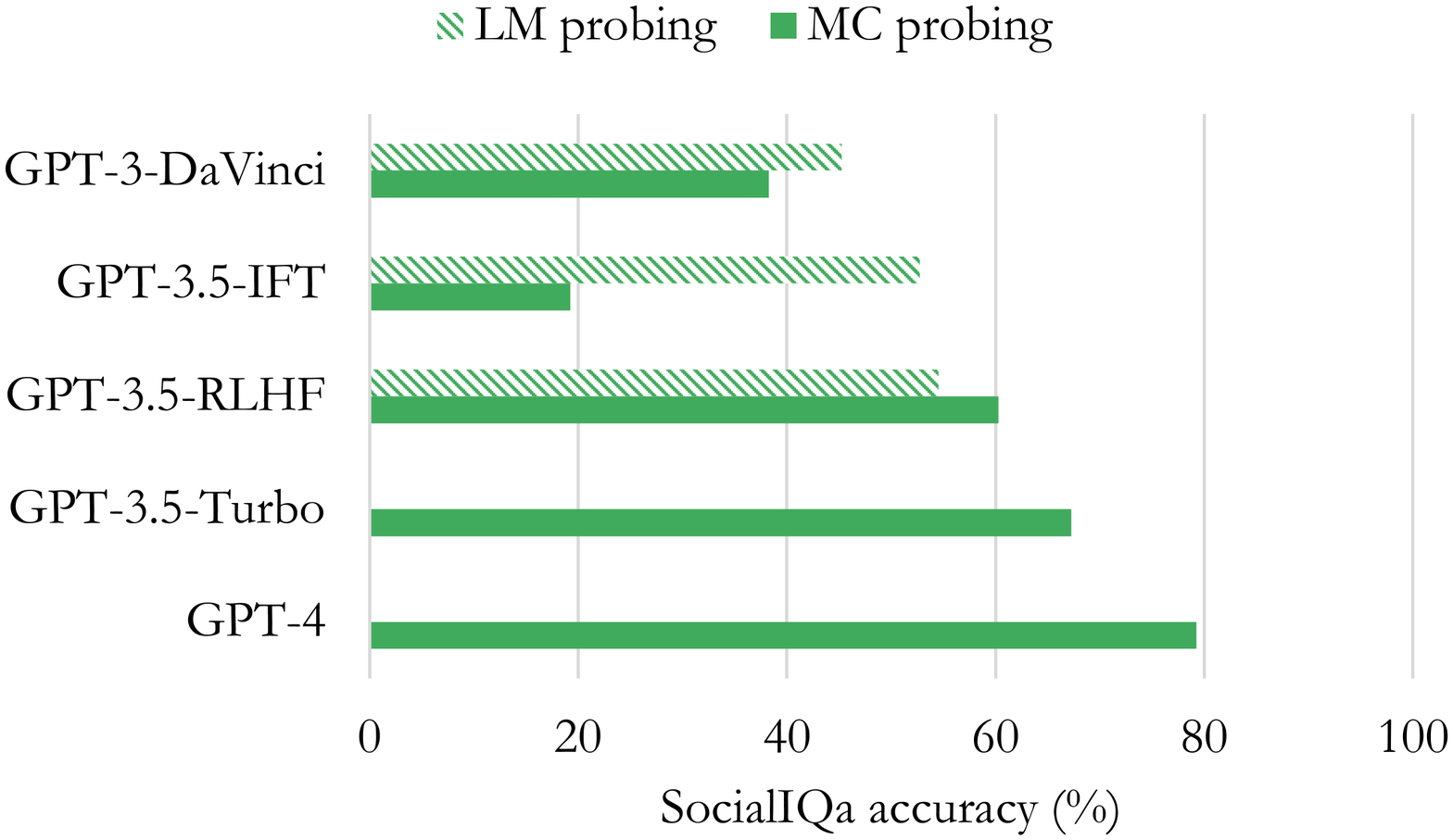}
        \caption{\socialIQa accuracy with different probing methods.}
        \label{subfig:gpt4-siqa-accs}
    \end{subfigure}
    \hspace{1em}
    \begin{subfigure}[b]{.95\columnwidth}
        \centering
        \includegraphics[height=4.2cm,clip,trim=2cm 4cm 2cm 4cm]{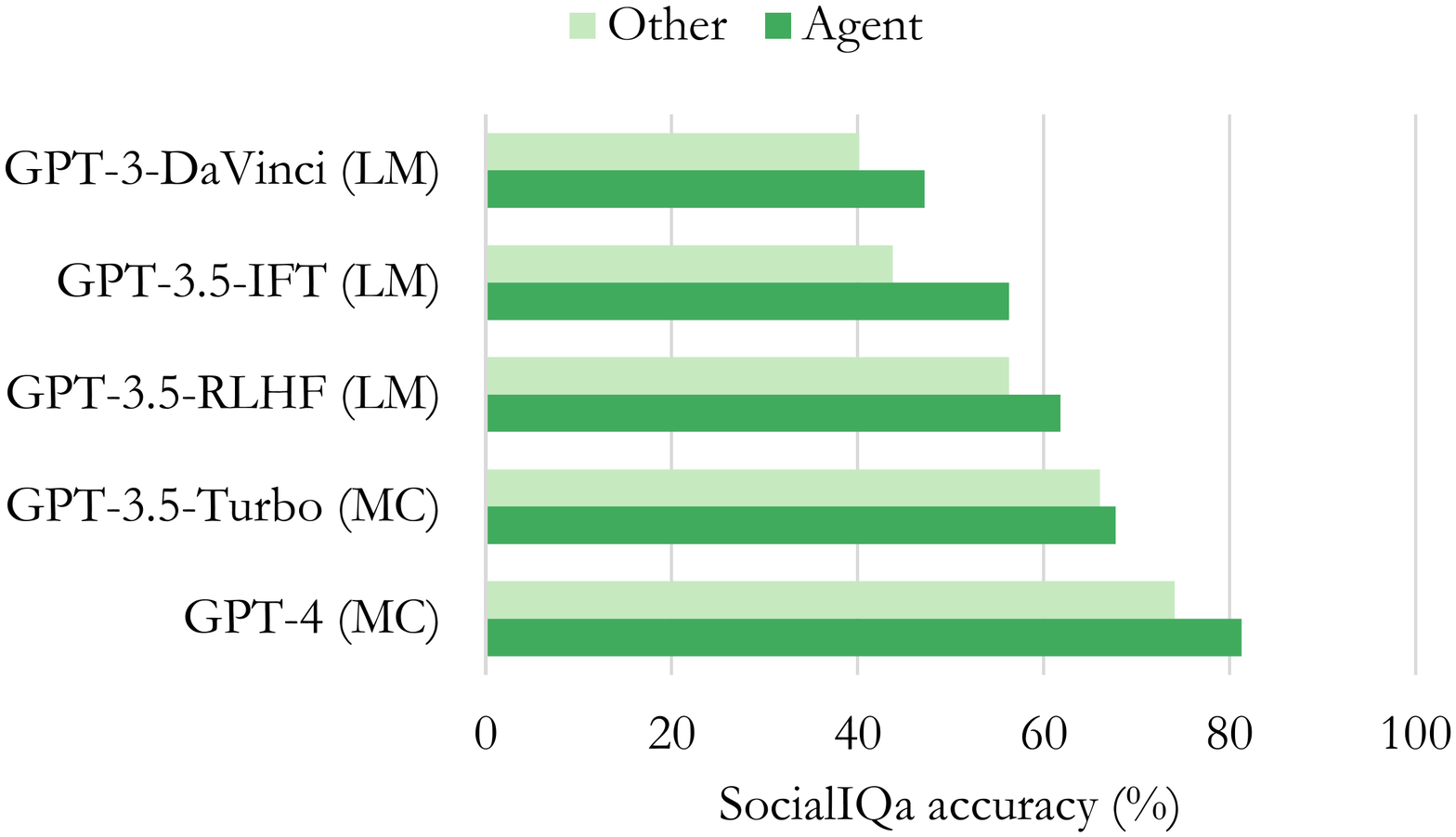}
        \caption{\socialIQa accuracy broken down by character.}
        \label{subfig:gpt4-siqa-char}
    \end{subfigure}
    \caption{\socialIQa results for new instruction-tuned and reinforcement learning \ptlms, broken down by probing type (\ref{subfig:gpt4-siqa-accs}) and by focus character of the question (\ref{subfig:gpt4-siqa-char}), on a random subset of 400 examples.}
    \label{fig:siqa-gpt4-results}
\end{figure*}

Shown in Fig. \ref{fig:siqa-gpt4-results}, our results show that instruction-tuning and RLHF do indeed improve performance on zero-shot social and emotional intelligence question answering in \socialIQa.
With LM probing, supervised instruction-finetuning (\davincitwo: 53\%) improves performance over vanilla language modelling (\gptDavinci: 45\%) slightly less than reinforcement learning (\davincithree: 55\%) .
% Compared to vanilla language modeling (\gptDavinci), supervised instruction-finetuning (\davincitwo) improves performance less than reinforcement learning (\davincithree).
Notably, multiple-choice probing only improves over random chance (33\%) with RLHF models (with 60, 67, and 79\% for \davincithree, \gptturbo, and \gptfour, respectively).

Focusing on the newest models, 
\gptturbo reaches 67.2\% performance, still $>$20\% below human performance reported in \cite{sap2019socialIQa}.
However, surprisingly, \gptfour performance increases to 79.3\%, within 10\% of human performance on \socialIQa.

Interestingly, however, all models and probing setups perform worse on questions pertaining to non-main characters (Fig. \ref{subfig:gpt4-siqa-char}), with \gptfour showing a 7\% decrease in accuracy.
This suggests that reporting biases due to centering theory, as discussed in \S\ref{sec:discussion-conclusion}, may still play a role even for these extreme scale models.

\textbf{Caveat:} it is increasingly likely that models were trained on benchmark data (either due to data collection via the online GUI or API, or by scraping web text without filtering). 
Indeed, the GPT-4 system report notes that some BIG-Bench test data \cite{Srivastava2022BeyondTI}, which contains the \socialIQa dev. dataset, was present in the training data of GPT-4 \cite[footnote 5]{OpenAI2023gpt4}.
Without more transparency on the training data and possible data contamination, conclusions about models achieving near-human performance cannot be drawn.

\subsection{\tomi: Reasoning about Mental States and Realities}
\begin{figure*}[t]
    \centering
    \begin{subfigure}[b]{.95\columnwidth}
        \centering
        \includegraphics[height=4.16cm]{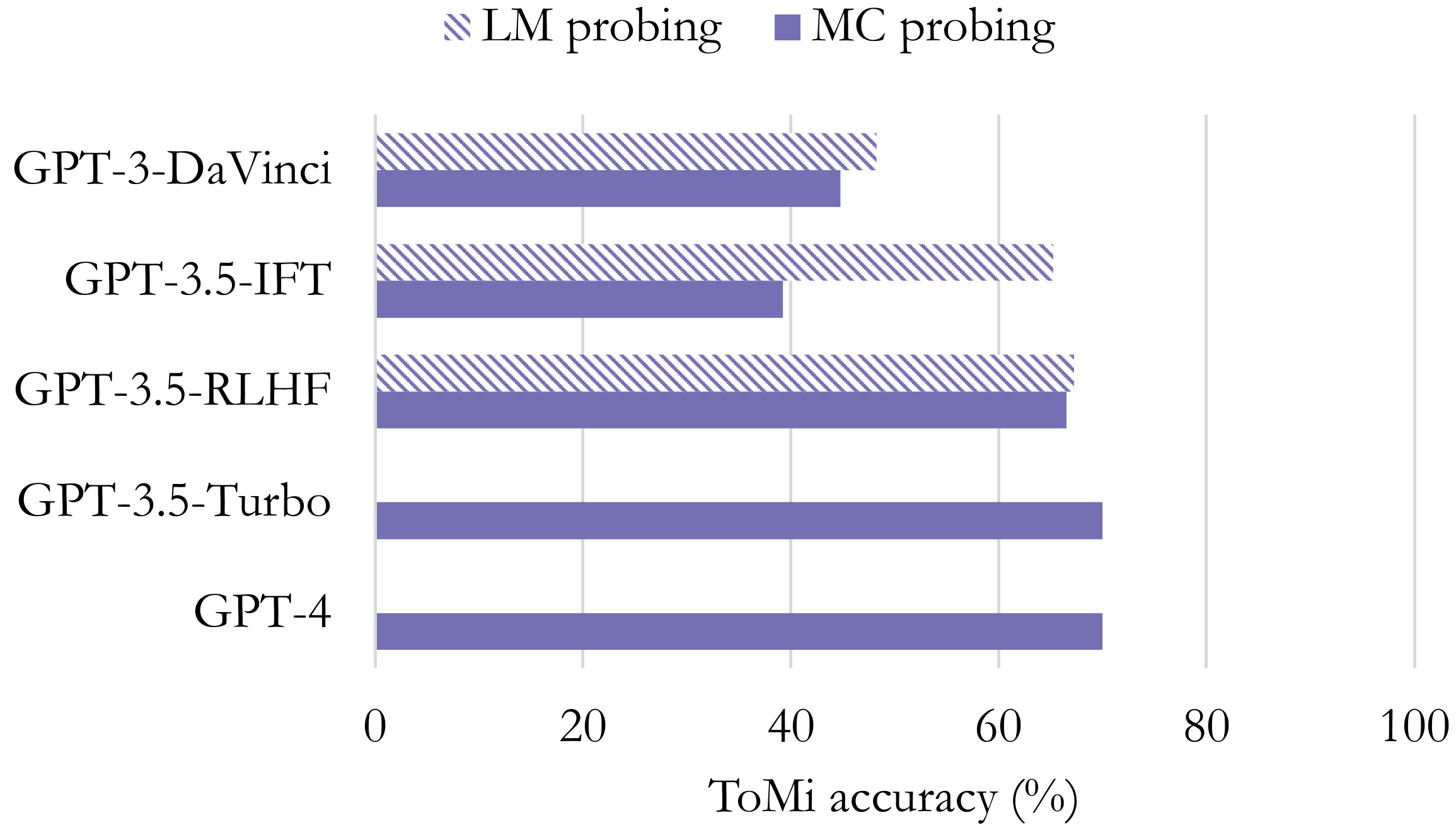}
        \caption{\tomi accuracy with different probing methods.}
        \label{subfig:gpt4-tomi-accs}
    \end{subfigure}
    \hspace{1em}
    \begin{subfigure}[b]{.95\columnwidth}
        \centering
        \includegraphics[height=4.2cm]{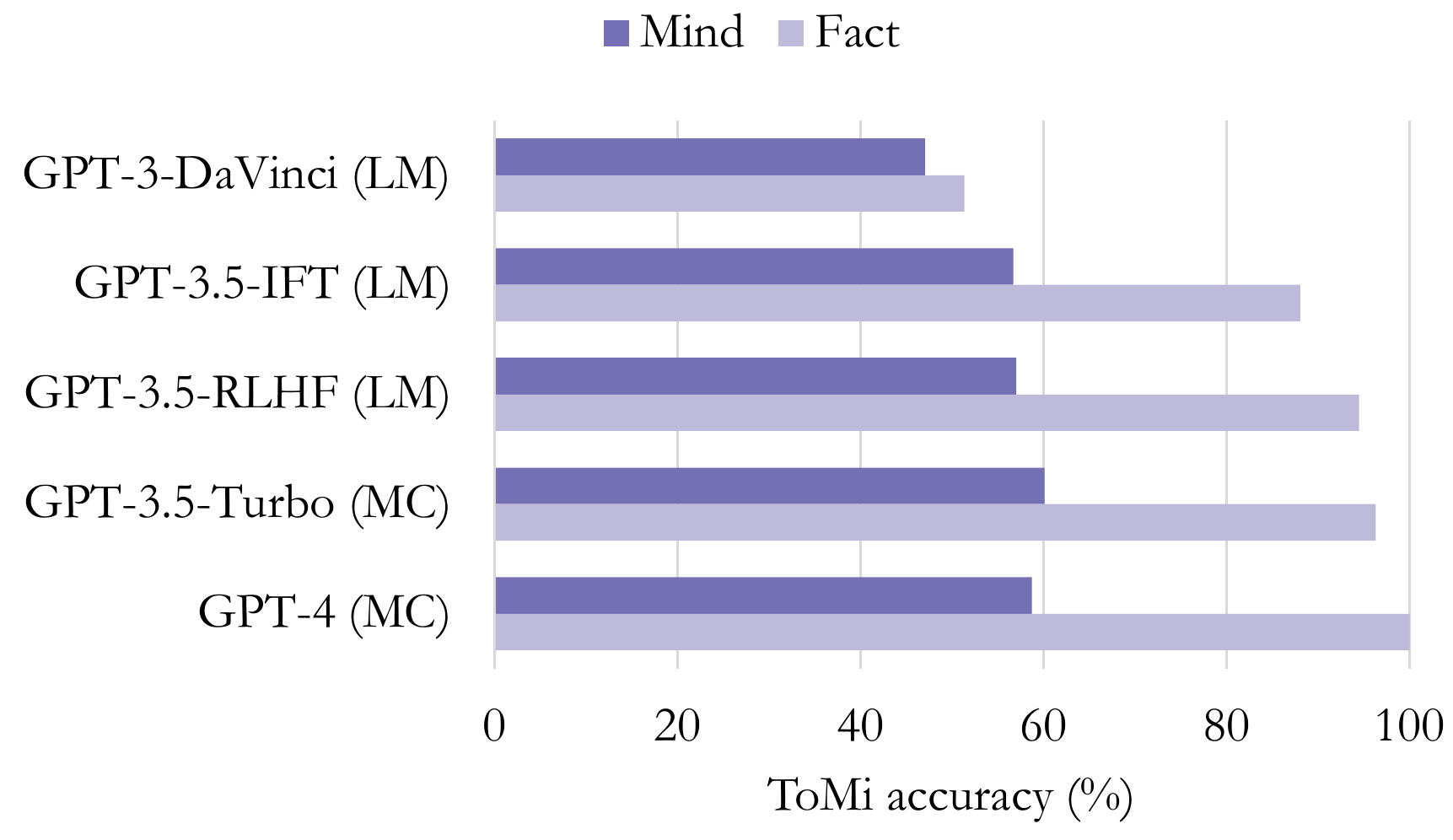}
        \caption{\tomi accuracy broken down by question type.}
        \label{subfig:gpt4-tomi-qType}
    \end{subfigure}
    \caption{\tomi results for new instruction-tuned and reinforcement learning \ptlms, broken down by probing type (\ref{subfig:gpt4-tomi-accs}) and by question type (\ref{subfig:gpt4-tomi-qType}), on a random subset of 400 examples.}
    \label{fig:tomi-gpt4-results}
\end{figure*}
Results on \tomi are plotted in Fig. \ref{fig:tomi-gpt4-results}, showing that instruction-finetuning and RLHF only somewhat improve the models' ability to reason about mental states and realities of others.
Compared to \gptDavinci's performance which is essentially random, both \davincitwo and \davincithree's performance increases by 17\% and 19\%, respectively.
However, in contrast to \socialIQa, performance of newer models (\gptturbo, \gptfour) does not substantially improve over GPT-3.5 models.

When breaking down the accuracy by question type (Fig. \ref{subfig:gpt4-tomi-qType}), we find that the performance of these models improves only on factual questions (fact), but stays low for questions about mental states of participants (mind).
Indeed, \gptturbo reaches the ``best'' performance on the \mind subset of \tomi with 60\% accuracy, surpassing \gptfour's 59\% accuracy.

% Further examining \gptfour's and other models' performance on \mind questions (Fig. \ref{fig:gpt4-tomi-full}), we find that earlier instruction-tuned
% \maarten{Add false vs. true beliefs?}

% \begin{figure*}
%     \centering
%     \includegraphics[width=\textwidth,clip,trim=2cm 6cm 2cm 6cm]{newgraphics/ToMi.gpt4.FactVsMindTbFb.pdf}
%     \caption{\tomi results for new instruction-tuned and reinforcement learning \ptlms, broken down by question type and belief type (\facts: factual questions, \mind: mental state questions, \mindTB: true-belief questions about mental states, \mindFB: false belief questions about mental states). \textbf{Takeaway}: as models become more capable, models improve on factual question answering, but not on reasoning about mental states.}
%     \label{fig:gpt4-tomi-full}
% \end{figure*}

\subsection{Discussion}
Based on our new results, it is not clear that the newer generation of models have achieved neural Theory of Mind, corroborating findings by \citet{ullman2023large}, \cite{Lenci2023-yg}, and \citet{Marcus2023-sq} and debunking claims of ``emergence of \ToM in \ptlms'' by \citet{kosinski2023theory} and \citet{bubeck2023sparks}.
While models may achieve higher accuracy on social intelligence tasks, their ability to model mental states and realities of others is still very far from humans (only 10\% over random chance).

Examining why these instruction-tuned or RLHF models perform somewhat better remains an open question, hindered by the lack of transparency in pretraining and instruction data.
Possibly, instruction-tuned models are better able to learn social intelligence due to the more interactional nature of instruction following or dialogue responding compared to static text.
However, improvements could solely be due to development and test data leakage as acknowledged by OpenAI \cite{OpenAI2023gpt4}, calling for the development of better evaluation and probing methods for these \ToM abilities.
Additionally, approaches such as person-centric inductive biases as well as interactive, experiential, or multimodal grounding could improve their ability to model mental states, as discussed in \S\ref{ssec:future-directions}.

\end{document}